%% file: main.tex
\def\arxivtng{1}

\if\arxivtng0
\documentclass[dvipsnames,11pt]{article}
\usepackage{fullpage}
\usepackage{parskip}
\else
\documentclass[dvipsnames]{article}
\usepackage[nonatbib,final]{neurips_data_2023}
\fi

\usepackage{fancyhdr}
\usepackage[numbers]{natbib}
\setcitestyle{numbers}

\usepackage[utf8]{inputenc} 
\usepackage[T1]{fontenc}    
\usepackage{url}            
\usepackage{booktabs}       
\usepackage{amsfonts}       
\usepackage{nicefrac}       
\usepackage{microtype}      
\usepackage{amsmath}
\usepackage{amssymb}
\usepackage{mathtools}
\usepackage{amsthm}
\usepackage{multirow}
\usepackage{hyperref}
\usepackage[table,dvipsnames]{xcolor}
\usepackage{babel}
\usepackage{subcaption}

\usepackage{siunitx}
\usepackage{caption} 

\hypersetup{
  colorlinks   = true, 
  urlcolor     = RoyalBlue, 
  linkcolor    = RoyalBlue, 
  citecolor   = RoyalBlue 
}

\definecolor{light-light-gray}{gray}{0.92} 

\usepackage{rotating}
\usepackage{placeins}
\usepackage{enumitem}
\usepackage{wrapfig}

\usepackage{colortbl}
\usepackage{cellspace}
\newcolumntype{w}{>{\columncolor{white}}c}
\usepackage{caption}
\usepackage{marginnote}

\usepackage[capitalize,noabbrev]{cleveref}
\newcommand{\customfootnotetext}[2]{{%
  \renewcommand{\thefootnote}{#1}%
  \footnotetext[0]{#2}}}%

\usepackage{xpatch}
\makeatletter
\renewcommand\paragraph{\@startsection{paragraph}{4}{\z@}                                     {1.35ex \@plus1ex \@minus.2ex}                                {-.5em}
{\normalfont\normalsize\bfseries}}
\makeatother

\usepackage{xspace}
\newcommand{\datanet}{\textsc{DataComp}\xspace}
\newcommand{\users}{participants\xspace}

\newcommand{\byod}{\textsc{BYOD}\xspace}
\newcommand{\pool}{\textsc{CommonPool}\xspace}
\newcommand{\ours}{\textsc{DataComp}-1B\xspace}

\usepackage{pifont}
\newcommand{\cmark}{\ding{51}}%
\newcommand{\xmark}{\ding{55}}%
\usepackage[textsize=tiny]{todonotes}

\title{\datanet:\\In search of the next generation of multimodal datasets\vspace{2pt}}

\iftrue
\author{
\vspace{-5mm}
\\
Samir Yitzhak Gadre*$^{2}$, Gabriel Ilharco*$^{1}$, Alex Fang*$^{1}$, Jonathan Hayase$^{1}$, \\
Georgios Smyrnis$^{5}$, Thao Nguyen$^{1}$, Ryan Marten$^{7,9}$, Mitchell Wortsman$^{1}$,\\
Dhruba Ghosh$^{1}$, Jieyu Zhang$^{1}$, Eyal Orgad$^{3}$, Rahim Entezari$^{10}$, Giannis Daras$^{5}$,\\ Sarah Pratt$^{1}$, Vivek Ramanujan$^{1}$, Yonatan Bitton$^{11}$, Kalyani Marathe$^{1}$,\\
Stephen Mussmann$^{1}$, Richard Vencu$^{6}$, Mehdi Cherti$^{6,8}$, Ranjay Krishna$^{1}$,\\
Pang Wei Koh$^{1,12}$, Olga Saukh$^{10}$, Alexander Ratner$^{1,13}$, Shuran Song$^{2}$,\\
Hannaneh Hajishirzi$^{1,7}$, Ali Farhadi$^{1}$, Romain Beaumont$^{6}$,\\
Sewoong Oh$^{1}$, Alex Dimakis$^{5}$, Jenia Jitsev$^{6,8}$,\\
Yair Carmon$^{3}$, Vaishaal Shankar$^{4}$, Ludwig Schmidt$^{1,6,7}$\\
}
\fi

\begin{document}

\date{}
\customfootnotetext{${^*}$}{Equal contribution, randomly ordered. Correspondence to \url{contact@datacomp.ai}.
  $^{1}$University of Washington
  $^{2}$Columbia University
  $^{3}$Tel Aviv University
  $^{4}$Apple
  $^{5}$UT Austin
  $^{6}$LAION
  $^{7}$AI2
  $^{8}$Juelich Supercomputing Center, Research Center Juelich
  $^{9}$University of Illinois Urbana-Champaign
  $^{10}$Graz University of Technology
  $^{11}$Hebrew University
  $^{12}$Google Research
  $^{13}$Snorkel AI
}

\maketitle

\vspace{-10pt}
\begin{abstract}
\vspace{-5pt}
Multimodal datasets are a critical component in recent breakthroughs such as CLIP, Stable Diffusion and GPT-4, yet their design does not receive the same research attention as model architectures or training algorithms.
To address this shortcoming in the machine learning ecosystem, we introduce \datanet, a testbed for dataset experiments centered around a new candidate pool of 12.8 billion image-text pairs from Common Crawl.
Participants in our benchmark design new filtering techniques or curate new data sources and then evaluate their new dataset by running our standardized CLIP training code and testing the resulting model on 38 downstream test sets.
Our benchmark consists of multiple compute scales spanning four orders of magnitude, which enables the study of scaling trends and makes the benchmark accessible to researchers with varying resources.
Our baseline experiments show that the \datanet workflow leads to better training sets.
Our best baseline, \ours, enables training a CLIP ViT-L/14 from scratch to 79.2\% zero-shot accuracy on ImageNet, outperforming OpenAI's CLIP ViT-L/14 by 3.7 percentage points while using the same training procedure and compute.
We release \datanet and all accompanying code at \url{www.datacomp.ai}.
\end{abstract}

\vspace*{-3mm}
\section{Introduction}
\vspace*{-3mm}
\label{sec:introduction}
Recent advances in multimodal learning such as   CLIP~\citep{radford2021learning}, DALL-E~\citep{ramesh2021zero,ramesh2022hierarchical}, Stable Diffusion~\citep{rombach2022high}, Flamingo~\citep{alayrac2022flamingo}, and GPT-4~\citep{gpt4} offer unprecedented generalization capabilities in zero-shot classification, image generation, and in-context learning.
While these advances use different algorithmic techniques, e.g., contrastive learning, diffusion, or auto-regressive modeling, they all rest on a common foundation: large datasets containing paired image-text examples.
For instance, CLIP’s training set contains 400 million image-text pairs, and Stable Diffusion was trained on the two billion examples from LAION-2B~\citep{laion5b}. 
This new generation of image-text datasets is 1,000 times larger than previous datasets such as ImageNet, which contains 1.2M images \citep{deng2009imagenet,ILSVRC15}.

\begin{table}[t]
\rowcolors{2}{light-light-gray}{white}
\setlength\tabcolsep{4.5pt}
\renewcommand{\arraystretch}{1.1}
\centering
\vspace{-5px}
\caption{Zero-shot performance of CLIP models trained on different datasets. \ours, assembled with a simple filtering procedure on image-text pairs from Common Crawl, leads to a model with higher accuracy than previous results while using the same number of multiply-accumulate operations (MACs) or less during training. 
See Section \ref{sec:evaluation} for details on the evaluation datasets.
}
\label{tab:tab1}
\begin{tabular}{lccccc}
\toprule
 & & \# samples  & &  Train compute  &ImageNet \\
\multirow{-2}{*}{Dataset}  & \multirow{-2}{*}{Dataset size}  & seen & \multirow{-2}{*}{Architecture} &  (MACs) & accuracy \\\midrule
OpenAI's WIT \cite{radford2021learning} & 0.4B & 13B & ViT-L/14 & $1.1\times 10^{21}$ & 75.5 \\
LAION-400M \cite{laion400m,cherti2022reproducible}  &  0.4B  & 13B & ViT-L/14 & $1.1\times 10^{21}$ & 72.8 \\
LAION-2B \cite{laion5b,cherti2022reproducible} & 2.3B  & 13B & ViT-L/14 & $1.1\times 10^{21}$& 73.1 \\
LAION-2B \cite{laion5b,cherti2022reproducible} & 2.3B  & 34B  & ViT-H/14 & $6.5\times 10^{21}$& 78.0 \\
LAION-2B \cite{laion5b,cherti2022reproducible} & 2.3B  & 34B  & ViT-g/14 & $9.9\times 10^{21}$& 78.5 \\
\midrule
\ours (ours)  & 1.4B  & 13B & ViT-L/14 & $1.1\times 10^{21}$ & \textbf{79.2} \\
\bottomrule
\end{tabular}
\vspace{-10pt}
\end{table}

Despite the central role of image-text datasets, little is known about them.
Many state-of-the-art datasets are proprietary,
and even for public datasets such as LAION-2B \cite{laion5b}, it is unclear how design choices such as the data source or filtering techniques affect the resulting models.
While there are thousands of ablation studies for algorithmic design choices (loss function, model architecture,
etc.), datasets are often treated as monolithic artifacts without detailed investigation. 
Moreover, datasets currently lack the benchmark-driven development process that has enabled a steady stream of improvements on the model side and isolates data enhancements from changes to the model.
These issues impede further progress in multimodal learning, as evidenced by recent work showing that public datasets currently do not match the scaling behavior of proprietary alternatives \citep{cherti2022reproducible}.

In this paper, we take a step towards a more rigorous dataset development process.
Our first and central contribution is \textbf{\datanet, a new benchmark for multimodal dataset design}.
\datanet flips the traditional benchmarking paradigm in machine learning where the dataset is fixed and researchers propose new training algorithms.
Instead, we hold the entire training code and computational budget constant so that participants innovate by proposing new training sets.
To evaluate the quality of a training set, we score the resulting model with a testbed of 38 classification and retrieval tasks such as ImageNet \cite{deng2009imagenet},
ImageNetV2 \cite{imagenetv2},
DTD \cite{dtd},
EuroSAT \cite{eurosat},
SUN-397 \cite{sun397},
and MSCOCO \cite{mscoco}.

\datanet focuses on two key challenges that arise when assembling large training datasets: what data sources to train on, and how to filter a given data source.
Each challenge corresponds to one track in our benchmark.
To facilitate the \emph{filtering track}, our second contribution is \textbf{\pool, a dataset of 12.8B image-text pairs collected from Common Crawl} and currently the largest public image-text dataset. 
We release CommonPool as an index of image url-text pairs under a CC-BY-4.0 license, and apply content checks in its construction to remove unsafe or unwanted content.
In the \emph{filtering track}, the goal of participants is to find the best subset of \pool to train on.
In the second track, \emph{Bring Your Own Data} (\byod), participants may leverage any data source, as long as it does not overlap with our evaluation testbed.

Our third contribution is an investigation of \textbf{scaling trends for dataset design}. 
In particular, \datanet contains \emph{four} scales, where we  vary the training budget and the candidate pool size from 12.8M to 12.8B samples (see \Cref{tab:scale-hparams}). 
Expressed in GPU hours, the cost of a single training run ranges from 4 to 40,000 GPU hours on the A100 cluster we used for development.
The different scales enable researchers with different resources to participate in our benchmark.
Moreover, our results show that the ranking of filtering approaches is largely consistent across scale.

Our fourth contribution is \textbf{over three hundred baseline experiments}, including techniques such as querying captions for relevant keywords, filtering based on image embeddings, and applying a threshold on CLIP scores.
A key result from our baselines experiments is that smaller, more stringently filtered datasets can lead to models that generalize \emph{better} than larger datasets coming from the same pool.
At the 12.8B scale, our best filtering baseline increases ImageNet zero-shot accuracy by 6.9 percentage points (pp) relative to the unfiltered pool (see \Cref{tab:main}).
For the \byod track, our initial experiments show that 109M additional data points (less than 1\% of the 12.8B pool) improve the CLIP-filtered subsets of \pool by up to 1.2 pp ImageNet accuracy (see \Cref{tab:byod}). 

Finally, our fifth contribution is \textbf{\ours, a new state-of-the-art multimodal dataset}.
We obtain \ours by combining our two most promising filtering baselines.
\ours enables training a CLIP ViT-L/14 model to an ImageNet zero-shot accuracy of 79.2\% (see \Cref{tab:tab1}), corresponding to a $9\times$ computational cost reduction when compared to a larger CLIP ViT-g/14 model trained on LAION-2B for about $3\times$ longer.
Moreover, our model outperforms OpenAI's original CLIP ViT-L/14 by 3.7 percentage points, while using the same compute budget.

To make \datanet a shared environment for controlled dataset experiments,
we publicly release our candidate pool url index, our tooling for assembling these pools, our filtering baselines, and our code for training and evaluating models at \url{www.datacomp.ai}.
We believe that our infrastructure will help put research on dataset design on rigorous empirical foundations, draw attention to this understudied research area, and lead to the next generation of multimodal datasets.

\begin{figure*}[t]
    \centering
    \includegraphics[width=0.95\textwidth]{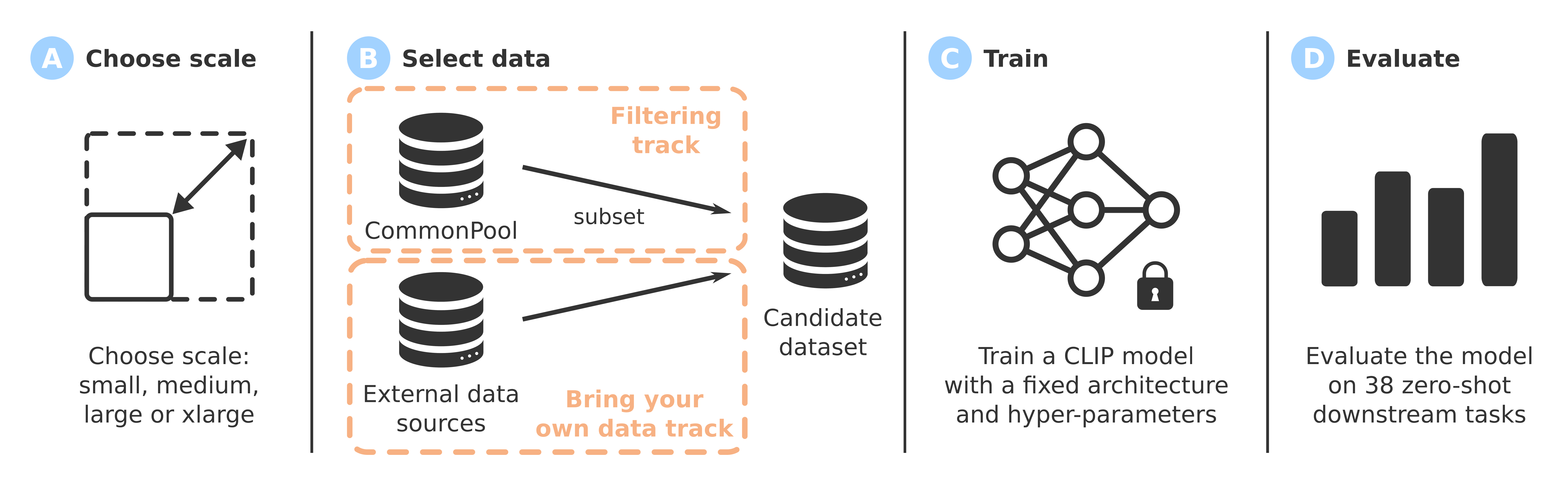}
    \caption{\datanet participant workflow. A) Choose a scale based on resource constraints. B) Design a dataset, in either the filtering or \byod track. C) Train a CLIP model on the designed dataset using a fixed architecture and hyperparameters (Section \ref{sec:training}). D) Evaluate the trained model on a suite of diverse downstream tasks (Section \ref{sec:evaluation}).}
    \label{fig:workflow}
    \vspace*{-3mm}
\end{figure*}

\section{Related Work}
\label{sec:relatedwork}

We review the most closely related work and include additional related work in Appendix \ref{sec:more-relatedwork}.

\paragraph{The effects of data curation.} 
Classical work considers dataset cleaning and outlier removal \cite{jiang2001two,yu2002findout,rousseeuw2011robust,rousseeuw2018anomaly} to discard samples that may lead to undesirable model bias. A related line of work develops coreset selection algorithms \cite{harpeled2004coresets,agarwal2004approximating,Feldman2011Scalable,pmlr-v37-bachem15,lucic2018gaussian,pmlr-v37-wei15,cohen2017input}, which aim to select data subsets that lead to the same performance as training on the entire dataset.
These techniques appear to scale poorly to larger data regimes \cite{guo2022deepcore,abbas2023semdedup}.
More recent efforts in subset selection often operate on already curated datasets \cite{craig,toneva2018empirical,sener2018active,birodkar2019semantic,coleman2020selection,datadiet} (e.g., CIFAR-10, ImageNet) or on smaller data regimes (e.g., YFCC-15M \cite{radford2021learning,yfcc100m}).
These settings often do not reflect newer training paradigms that involve (1) \emph{noisy} image-text pairs instead of category labeled images and (2) large scale datasets (e.g., billions of samples).
While data-centric investigations have led to community competitions like \textsc{dcbench} \cite{dcbench} and \textsc{DataPerf} \cite{dataperf}, existing benchmarks have likewise operated at small data scales \cite{datacentric} compared to datasets like LAION-2B \cite{laion5b}, which contains over two billion images.
\datanet bridges this gap by aligning data-centric investigation with large scale image-text training.

There has also been renewed interest in dataset pruning and deduplication.
\citet{sorscher2022beyond} show that data pruning can improve traditional
scaling trends on ImageNet, but do not consider image-text training or larger datasets.
\citet{raffel2020exploring} remove sentence redundancies when creating the C4 corpus. Subsequent work further demonstrated the benefits of deduplication for better language modeling \cite{Lee2021DeduplicatingTD}.
\citet{rdk+23} introduce CAT filtering for image-text datasets---a rule-based system to retain high quality samples. 
\citet{abbas2023semdedup} propose SemDeDup, which starts with the CAT-filtered LAION-440M subset, further employing clustering to remove semantic duplicates.
\datanet facilitates data-centric investigation at an even larger scale (i.e., 12.8B sample scale) and provides a common experimental setting for fair comparison amongst dataset creation algorithms.

\paragraph{Large-scale multimodal datasets.} Datasets have been instrumental to building multimodal models like CLIP~\cite{radford2021learning}, Flamingo~\cite{alayrac2022flamingo}, Stable Diffusion ~\cite{rombach2022high}, DALL-E~\cite{ramesh2021zero,ramesh2022hierarchical} and GPT-4 \cite{gpt4}.
These methods succeeded by training on large, heterogeneous datasets rather than solely through advanced modelling techniques.
For example, OpenAI's CLIP trains on 400M image-text pairs from the web, roughly $300\times$ the size of ImageNet~\cite{deng2009imagenet}.
Prior work on scaling image-text datasets also provides promising trends with respect to zero-shot model performance~\cite{jia2021scaling,pham2021scaling}. 
Additional large scale datasets like FILIP-300M \cite{filip}, FLD-900M \cite{yuan2021florence}, and PaLI-10B \cite{chen2022pali} were constructed to train multimodal models.
However, many datasets used to train such models (including the dataset for OpenAI's CLIP) are proprietary, making it hard to conduct data-centric investigations.

Even for public image-text datasets like SBU~\cite{sbu}, Flickr30k \cite{flickr30k}, MS-COCO \cite{mscoco}, TaiSu \cite{liu2022taisu}, Conceptual Captions \cite{sharma2018conceptual}, CC12M \cite{changpinyo2021conceptual}, RedCaps \cite{desai2021redcaps}, WIT \cite{srinivasan2021wit}, Shutterstock \cite{nguyen2022quality}, YFCC-100M \cite{yfcc100m}, COYO-700M \cite{coyo700m}, LAION-400M \cite{laion400m}, or LAION-2B \cite{laion5b} little is known about what constitutes a good image-text dataset. Preliminary analysis suggests that different image-text data sources lead to CLIP models with different properties~\cite{nguyen2022quality}. However, previous work is limited to smaller scale data (10-15M examples). 
\citet{Birhane2021MultimodalDM} examine LAION-400M and find NSFW imagery and racial slurs, centering the dangers in web-scale multimodal datasets. To combat toxicity, we preprocess our pool to remove NSFW content and blur human faces detected in images. For more details on our safety preprocessing see Section~\ref{sec:pool}, Appendices~\ref{app:nsfw}~and~\ref{app:face}.

\section{The \datanet benchmark}
\label{sec:thecomp}

\datanet is meant to facilitate data-centric experimentation.
While traditional benchmarks emphasize model design, \datanet is centered around dataset development, where the resulting datasets can be used to train high accuracy models.
We focus on large image-text datasets and quantify a dataset submission by training a CLIP model on it from scratch \cite{radford2021learning} and evaluating on 38 downstream image classification and retrieval tasks. 
We additionally have three secret test sets, which will be released after a year, to guard against overfitting. 
To facilitate such investigations, we provide a candidate pool of uncurated image-text pairs sourced from the public internet.
Our benchmark offers two tracks: one where \users must filter samples from the pools we provide, and another where \users can use external data.
Moreover, \datanet is structured to accommodate \users with diverse levels of computational resources: each track is broken down into four scales with varying compute requirements.
We now discuss high-level design decisions, construction of a 12.8B image-text data pool to facilitate the competition, benchmark tracks, model training, and evaluation.

\subsection{Competition design}
\paragraph{Overview.} In many areas of machine learning, larger datasets lead to better performing models~\cite{krizhevsky2012imagenet,kaplan2020scaling,jia2021scaling,pham2021scaling,chinchilla,cherti2022reproducible,gpt3,radford2021learning,radford2022robust}.
Hence comparing only datasets with the same size is a natural starting point.
However, this approach is flawed as controlling the dataset size ignores critical curation constraints: candidate pool size (i.e., number of image-text pairs to harvest) and training compute.
For instance, assembling a dataset like LAION-2B consists of identifying \emph{data sources} (e.g., Common Crawl or Reddit)
and \emph{filtering} the data source.  
Notably, \emph{the final dataset size is a design choice} and is only upper-bounded by the data sources.
Hence, the true data constraint is the size of the reservoir of samples: \emph{candidate pool} to be filtered.
To make \datanet a realistic benchmark, we therefore fix the candidate pool in the filtering track, but give participants control over the training set size.

Compute cost is another relevant constraint.
To put datasets of different size on equal footing, we specify the total \emph{number of training samples seen}. 
Consider the 12.8B compute scale and filtered datasets $A$ and $B$, with 6.4B and 3.2B image-text pairs respectively.
At this scale, we train by making two passes over $A$, while making four passes over $B$.
A key result from our experiments is that smaller, more stringently filtered datasets can lead to models that generalize \emph{better}. 

\paragraph{Competition tracks.} Two key procedures in assembling a training dataset are filtering a data source \cite{laion400m,laion5b,coyo700m} and aggregating data sources \cite{dave2020tao,deng2009imagenet}.
To reflect this structure, \datanet has two tracks: \textit{filtering}, where \users select a subset of the samples from \pool, and \textit{Bring Your Own Data} (\byod), where \users can use any source of data. Key decisions for each tracks are described in Sections \ref{sec:pool} and \ref{sec:byod}, respectively. For full competition track rules see Appendix~\ref{sec:full-competition-rules}.

\begin{table}
\rowcolors{2}{white}{light-light-gray}
\caption{Experimental configurations, with compute in multiply-accumulate operations (MACs).
}
\setlength\tabcolsep{8.5pt}
\renewcommand{\arraystretch}{1.1}
\small
\centering
\begin{tabular}{llcc}
\toprule
Scale  & Model    & Train compute (MACs) & Pool size and \# samples seen\\\midrule
{\small \texttt{small}}    & ViT-B/32 & $9.5\times 10^{16}$ & 12.8M\\
{\small \texttt{medium}}   & ViT-B/32 & $9.5\times 10^{17}$ & 128M  \\
{\small \texttt{large}}	& ViT-B/16 & $2.6\times 10^{19}$ & 1.28B   \\
{\small \texttt{xlarge}}	& ViT-L/14 & $1.1\times 10^{21}$ & 12.8B \\
\bottomrule
\end{tabular}
\label{tab:scale-hparams}
\end{table}

\paragraph{Competition compute scales.} To facilitate study of scaling trends and accommodate participants with various computational resources, we structure \datanet using four scales of compute: {\small \texttt{small}}, {\small \texttt{medium}}, {\small \texttt{large}} and {\small \texttt{xlarge}}. Each new scale increases the number of samples seen during training by 10$\times$ (from 12.8M to 12.8B samples seen), and the pool we provide by the same factor (from 12.8M samples to 12.8B samples). Table \ref{tab:scale-hparams} gives the experimental configuration used for each scale.
For the {\small \texttt{small}} scale, our runs took 4 hours on an A100 GPU, and for the {\small \texttt{xlarge}} scale 81 hours on 512 GPUs.

\subsection{\pool generation, for the filtering track}
\label{sec:pool}
We construct a large-scale pool of image-text pairs, \pool, from Common Crawl~\cite{commoncrawl}.
CommonPool is distributed as an image url-text pair index under a CC-BY-4.0 license.
Our pool construction pipeline has four steps: url extraction and data download, NSFW detection, evaluation set deduplication, and face blurring. We additionally provide per sample metadata (e.g., CLIP features).
Starting from the {\small \texttt{xlarge}} \pool, we take successive random subsets to create {\small \texttt{large}}, {\small \texttt{medium}}, and {\small \texttt{small}} \pool (e.g., {\small \texttt{medium}} is a subset of {\small \texttt{large}}).

\paragraph{Extracting urls and dowloading data.} We first use \texttt{cc2dataset}~\cite{cc2dataset}, which utilizes Apache Spark~\cite{zaharia2016apache}, to extract pairs of image urls and nonempty alt-text from all Common Crawl snapshots from 2014 to 2022. We then deduplicate the url-text pairs and randomly shuffle. This step results in $\sim$88B possible samples. Not all samples are downloadable; other samples are not suitable due to NSFW content or overlap with our evaluation sets.
We attempt to download $\sim$40B samples using \texttt{img2dataset}~\cite{img2dataset} resulting in $\sim$16.8B image-text pairs. 
For more details, see Appendix~\ref{app:parse-cc}.

\paragraph{Safety preprocessing.}
Since Common Crawl is a snapshot of the internet, we require strict preprocessing to remove unsafe content. We use Detoxify~\cite{Detoxify} to prune samples that contain unsafe text (e.g., obscene, sexually explicit, or threatening language). 
We also discard samples with explicit visual content. To do so, we train a classifier on CLIP ViT-L/14 \cite{radford2021learning} features, using the NSFW dataset used in LAION-5B~\cite{laion5b}. We validate our classifier against the Google commercial image safety API. See Appendix \ref{app:nsfw} for details.
Around 19\% of image-text pairs are considered NSFW, taking the pool of $\sim$16.8B downloads to $\sim$13.6B samples.

\paragraph{Evaluation set deduplication.}
To prevent accidental overfitting to certain test sets in our evaluation suite, we perform a thorough near-duplicate removal between the candidate pool and our evaluation sets, using a state-of-the-art image deduplication model~\cite{Yokoo2021Dedup}. Appendix \ref{app:dedup} contains additional details.
The model flags $\sim$3\% of the 16.8B images as near-duplicates, reducing the $\sim$13.6B pool to $\sim$13.1B samples.
From here we select a random subset to get the {\small \texttt{xlarge}} pool of 12.8B samples.

\paragraph{Face detection \& blurring.} To protect the privacy of individuals, we detect and blur faces from images in our pool using a face detector~\cite{guo2021sample}. As observed by \citet{yang2022study}, obfuscating faces has little impact on model performance, as we also observe in our experiments (Appendix \ref{app:face}).

\paragraph{Pool metadata.}
To bootstrap \users we distribute metadata for each sample in \pool (e.g., image url, alt-text, original image resolution, CLIP features, and CLIP similarity scores).
Following \citet{carlini2023poisoning}, we release SHA256 hashes for each image to guard against data poisoning in subsequent \pool downloads.
For additional details see Appendix \ref{app:metadata}.
We open-source our metadata processing pipeline as \texttt{dataset2metadata}~\cite{dataset2metadata}.

\subsection{The bring your own data (\byod) track}
\label{sec:byod}

While \pool can be used to study different filtering techniques, state-of-the-art models often train on data from different sources.
For instance, the Flamingo model~\cite{alayrac2022flamingo} uses both multimodal massive web (M3W) and ALIGN datasets~\cite{jia2021scaling}.
To facilitate non-proprietary research on curating data from many sources, we instantiate a separate \datanet track to allow \users to combine multiple data streams.
For example, \users could construct a training set from CC12M~\cite{changpinyo2021conceptual}, YFCC100M~\cite{yfcc100m}, and data sources they label themselves. In Section \ref{sec:byod-baselines} and Appendix \ref{app:byod} we describe our exploration using existing public, image-text datasets. These datasets are acquired from their respective sources and are not re-release as part of \datanet.

\subsection{Training}
\label{sec:training}

We create a common experimental setting that enables comparable experiments by fixing the training procedure.
We closely follow the CLIP training recipe proposed by \citet{radford2021learning}: training models from scratch with a contrastive objective over images and captions.
Given a set of image-caption pairs, we train an image encoder and a text encoder such that the similarity between the representations of images and their corresponding text is maximized relative to unaligned pairs.\footnote{More precisely, given a batch of data $\{(x_1,y_1),...,(x_B,y_B)\}$ with images $x$ and captions $y$, we train the image encoder $g$ and text encoder $v$ with the loss $\ell=\frac{1}{2}\sum_{i=1}^B \frac{\sigma_{ii}}{\sum_{j=1}^B\sigma_{ij} } + \frac{1}{2}\sum_{i=1}^B \frac{\sigma_{ii}}{\sum_{j=1}^B\sigma_{ji}}$, where $\sigma_{ij} = \exp{\langle g(x_i), h(y_j) \rangle}$. We also use a learnable temperature parameter as in \citet{radford2021learning}.}
For each scale, we fix the model architecture and hyperparameters (see Table~\ref{tab:scale-hparams}). We pick Vision Transformers (ViTs) \cite{dosovitskiy2021an} as the image encoder, considering the better scaling trends observed by \citet{radford2021learning} compared to ResNets \cite{he2016deep}. 
Models are trained for a fixed number of steps determined by the scale (Table \ref{tab:scale-hparams}), using the OpenCLIP repository \cite{ilharco2021openclip}. 
See Appendix \ref{app:train} for details.

\subsection{Evaluation}
\label{sec:evaluation}

We evaluate on a suite of 38 image classification and retrieval tasks. We also study two additional fairness tasks, detailed in Section~\ref{sec:analysis} and Appendix~\ref{app:fairness}.
As discussed in Section \ref{sec:pool}, we remove test set images from \datanet to avoid contamination.
Image classification datasets range from satellite imagery recognition to classifying metastatic tissues. 
In total we have (with some overlap): 22 of the datasets evaluated in \citet{radford2021learning}, 6 ImageNet distribution shifts (i.e., ImageNet-Sketch~\cite{imagenetsketch}, ImageNet-V2~\cite{imagenetv2}, ImageNet-A~\cite{imageneta_and_imageneto}, ImageNet-O~\cite{imageneta_and_imageneto}, ImageNet-R~\cite{imagenetr}, and ObjectNet~\cite{objectnet}), 13 datasets from
VTAB \cite{vtab}, and 3 datasets from WILDS \cite{wilds2021,sagawa2022extending}.
Retrieval datasets include Flickr30k \cite{flickr30k}, MSCOCO \cite{mscoco}, and the WinoGAViL commonsense association task \cite{bitton2022winogavil}.
To aggregate results over all evaluation tasks, we average the preferred metric for each task.

\datanet adopts a zero-shot evaluation protocol: models are tested without training on the evaluation tasks. This approach is computationally efficient and measures a model's ability to perform well without any additional training.
We find a strong rank correlation (${>}0.99$) between performance in linear probe zero-shot settings (Appendix Figure \ref{fig:linear-probes}).
Additional details are in Appendix~\ref{sec:app-eval}.

\section{Baselines}
\label{sec:baselines}

\subsection{Filtering baselines}
We study six simple filtering methods for the filtering track; see \Cref{sec:app-baselines-pool} for further details.

\paragraph{No filtering.} We simply use the entire pool as the subset, without any filtering. Since each pool size is equal to the sample budget, training consists of one pass over the data.

\paragraph{Random subsets.} To isolate the effects of increasing the compute budget from increasing the dataset size, we form subsets consisting of 1\%, 10\%, 25\%, 50\% and 75\% of the pool chosen at random. 

\paragraph{Basic filtering.} We consider many simple filtering operations inspired by \citet{laion400m} and \citet{coyo700m}: filtering by \emph{language} (English captions, using either fasttext~\citep{joulin2017bag} or cld3~\cite{cld3}); filtering by \emph{caption length} (over two words and five characters); and filtering by \emph{image size} (smaller dimension above 200 pixels and aspect ratio below three). We also experiment with combining language and caption length filtering and combining language, caption length, image size fitering. Unless otherwise specified, ``basic'' refers fasttext English, caption length, and image size filtering.

\paragraph{CLIP score and LAION filtering.} We experiment with CLIP score filtering (also employed by LAION), where we take only examples having cosine similarity scores between CLIP image and text embeddings that exceed a pre-defined threshold. We investigate a range of thresholds and two OpenAI CLIP models for computing the scores: the ViT-B/32 model (as in LAION) and the larger ViT-L/14. We also combine CLIP score thresholds and cld3 English filtering to reproduce the LAION-2B filtering scheme. \Cref{tab:filtering_thresholds} in \Cref{sec:app-baselines-pool} summarizes the different CLIP score configurations. 

\paragraph{Text-based filtering.} We select examples that contain text overlapping with ImageNet class names, which serve as a proxy for relevance to downstream tasks. Specifically, we select English captions (according to fasttext) that contain words from ImageNet-21K or ImageNet-1K~\citep{deng2009imagenet} class synsets.

\paragraph{Image-based filtering.} We select a subset of examples whose visual content overlaps with ImageNet classes. After applying English language (fasttext) and caption length filtering, we cluster the image embeddings extracted by the OpenAI ViT-L/14 model for each image into 100K groups using Faiss~\citep{johnson2019billion}. We then find the nearest neighbor group for every ImageNet training example, and keep examples belonging to these groups. We apply this procedure using either ImageNet-21K (14M images) or ImageNet-1K (1.2M images), forming two subsets.

\subsection{\byod baselines}
\label{sec:byod-baselines}

We experiment with multiple external data sources, including four moderately sized datasets (10 to 58M samples) studied by \citet{nguyen2022quality}---CC12M \cite{changpinyo2021conceptual}, YFCC15M \cite{yfcc100m,radford2021learning}, RedCaps \cite{desai2021redcaps} and Shutterstock \cite{nguyen2022quality}---and the larger LAION-2B~\cite{laion5b}. Additional experiments, along with more details about the data sources are provided in Appendix \ref{app:byod}. We consider these data sources as they are and do not perform additional preprocessing. We also present experiments  combining some of the data sources (using only the external datasets, or in addition to data from our pool).

\begin{table*}
\rowcolors{3}{light-light-gray}{white}
\caption{Zero-shot performance for select baselines in the \textit{filtering} track. On all scales, filtering strategies lead to better performance than using the entire, unfiltered pool. The intersection between imaged-based and CLIP score strategies performs well on most tasks and scales. For all metrics, higher is better (see Appendix \ref{sec:app-eval} for details). $\cap$ denotes the intersection of filtering strategies.
}
\setlength\tabcolsep{5pt}
\renewcommand{\arraystretch}{1.1}
\small
\centering

\resizebox{\textwidth}{!}{
\begin{tabular}{wlccccccc}
\toprule
 &    & Dataset & Samples &  & ImageNet &  &  & Average over \\
\multirow{-2}{*}{Scale} & \multirow{-2}{*}{Filtering strategy} & size & seen &  \multirow{-2}{*}{ImageNet} & dist. shifts & \multirow{-2}{*}{VTAB}  & \multirow{-2}{*}{Retrieval} & 38 datasets \\\midrule
 \cellcolor{white} & No filtering & 12.8M  & 12.8M& 0.025 & 0.033 & 0.145 & 0.114 &  0.132 \\
\cellcolor{white} & Basic filtering & 3M  & 12.8M& 0.038 & 0.043 & 0.150 & 0.118 & 0.142 \\
\cellcolor{white}& Text-based & 3.2M  & 12.8M & 0.046 & 0.052 & 0.169 & \underline{0.125} & 0.157 \\
\cellcolor{white}& Image-based & 3M  & 12.8M & 0.043 & 0.047 & 0.178 & 0.121 & 0.159 \\
\cellcolor{white}& LAION-2B filtering& 1.3M  & 12.8M& 0.031 & 0.040 & 0.136 & 0.092 & 0.133 \\
\cellcolor{white}& CLIP score (L/14 30\%) & 3.8M  & 12.8M& \underline{0.051} & \underline{0.055} & \underline{0.190} & 0.119 & \underline{0.173} \\
\cellcolor{white}\multirow{-7}{*}{{\small \texttt{small} }}& Image-based $\cap$ CLIP score (L/14 30\%) & 1.4M  & 12.8M & 0.039 & 0.045 & 0.162 & 0.094 & 0.144 \\\midrule

\cellcolor{white} & No filtering & 128M  & 128M & 0.176 & 0.152 & 0.259 & 0.219 & 0.258 \\
\cellcolor{white}& Basic filtering & 30M  & 128M & 0.226 & 0.193 & 0.284 & 0.251 & 0.285 \\
\cellcolor{white}& Text-based  & 31M  & 128M & 0.255 & 0.215 & 0.328 & 0.249 & 0.307 \\
\cellcolor{white} & Image-based  & 29M  & 128M & 0.268 & 0.213 & 0.319 & \underline{0.256} & 0.312 \\
\cellcolor{white} & LAION-2B filtering & 13M  & 128M & 0.230 & 0.198 & 0.307 & 0.233 & 0.292 \\
\cellcolor{white} & CLIP score (L/14 30\%) & 38M  & 128M & 0.273 & 0.230 & 0.338 & 0.251 & \underline{0.328} \\
\cellcolor{white} \multirow{-7}{*}{{\small \texttt{medium}}} & Image-based $\cap$ CLIP score (L/14 30\%) & 14M  & 128M & \underline{0.297} & \underline{0.239} & \underline{0.346} & 0.231 & \underline{0.328} \\\midrule

\cellcolor{white}  & No filtering & 1.28B  & 1.28B & 0.459 & 0.378 & 0.426 & 0.419 & 0.437\\
\cellcolor{white} & Basic filtering & 298M  & 1.28B & 0.516 & 0.423 & 0.446 & 0.480 & 0.458 \\
\cellcolor{white} & Text-based & 317M  & 1.28B & 0.561 & 0.465 & 0.465 & 0.352 & 0.466 \\
\cellcolor{white} & Image-based & 293M & 1.28B  & 0.572 & 0.454 & 0.483 & 0.479 & 0.476 \\
\cellcolor{white} & LAION-2B filtering & 130M & 1.28B  & 0.553 & 0.453 & 0.510 & 0.495 & 0.501 \\
\cellcolor{white} & CLIP score (L/14 30\%) & 384M & 1.28B  & 0.578 & 0.474 & 0.538 & 0.466 & 0.529 \\
\cellcolor{white} \multirow{-7}{*}{{\small \texttt{large}}} & Image-based $\cap$ CLIP score (L/14 30\%) & 140M & 1.28B  & \underline{0.631} & \underline{0.508} & \underline{0.546} & \underline{0.498} & \underline{0.537} \\\midrule 

 \cellcolor{white} & No filtering & 12.8B  & 12.8B & 0.723 & 0.612 & 0.611 & 0.569 & 0.621 \\
\cellcolor{white} & LAION-2B filtering & 1.3B & 12.8B  & 0.755 & 0.637 & 0.624 & \underline{0.620} & 0.636 \\
\cellcolor{white} & CLIP score (L/14 30\%) & 3.8B & 12.8B  & 0.764 & 0.655 & 0.643 & 0.588 & 0.650 \\
\cellcolor{white} \multirow{-4}{*}{{\small \texttt{xlarge}}}  & Image-based $\cap$ CLIP score (L/14 30\%) & 1.4B & 12.8B  & \underline{0.792} & \underline{0.679} & \underline{0.652} & 0.608 & \underline{0.663} \\
\bottomrule
    \end{tabular}}
\label{tab:main}
\end{table*}

\section{Results and discussion}\label{sec:analysis}

\subsection{Building better datasets}

\paragraph{Main results.} Our key results are in Table \ref{tab:main}. Most notably, the intersection between image-based filtering and CLIP score filtering excels on most tasks. The exception is at the {\small \texttt{small}} scale and for retrieval datasets.\footnote{\citet{cherti2022reproducible} also observe that models rank differently on classification and retrieval tasks.} Furthermore, other filtering strategies like basic, CLIP score, image-based, text-based filtering show better downstream performance when compared to no filtering.
A much larger suite of experiment results can be found in Appendix~\ref{sec:app-more-plots}.

\paragraph{\datanet leads to better image-text datasets.}
We hope \datanet catalyzes the search for the next generation of multimodal datasets.
We contribute \ours, which is the output of the Image-based $\cap$ CLIP score (L/14 30\%) baseline filter at the {\small \texttt{xlarge}} scale of the filtering track.
Our dataset is comprised of 1.4B samples, which not only is \emph{smaller} than the LAION-2B dataset with 2.3B samples, but also comes from a smaller pool.
Nevertheless, a CLIP L/14 trained on \ours outperforms the LAION-2B competitor by 6.1 percentage points on ImageNet (see Table \ref{tab:tab1}). Moreover, training on \ours improves ImageNet accuracy by 3.7 percentage points over OpenAI's ViT-L/14 trained with the same compute budget. Additionally, even if we restrict ourselves to 400M samples, we can still find a subset of \ours that outperforms OpenAI's ViT-L/14, as seen in Table~\ref{tab:sample400m}. These results demonstrate the impact that \datanet can make and provide a foundation upon which \users can build.

\paragraph{External data sources can improve performance.} Appendix~\ref{app:byod} Table \ref{tab:byod} shows results for several baselines in the \byod track.
We find several instances where adding external data sources improves performance over using just data from \pool. For example, 
at the {\small \texttt{large}} scale, combining CLIP-filtered data from \pool with external data from CC12M \cite{changpinyo2021conceptual}, YFCC15M \cite{yfcc100m,radford2021learning}, RedCaps \cite{desai2021redcaps} and Shutterstock \cite{nguyen2022quality} boosts ImageNet accuracy by 4.3 percentage points.
See Appendix \ref{app:byod} for more experiments and details.

\begin{figure}
    \centering
    \includegraphics[width=\linewidth]{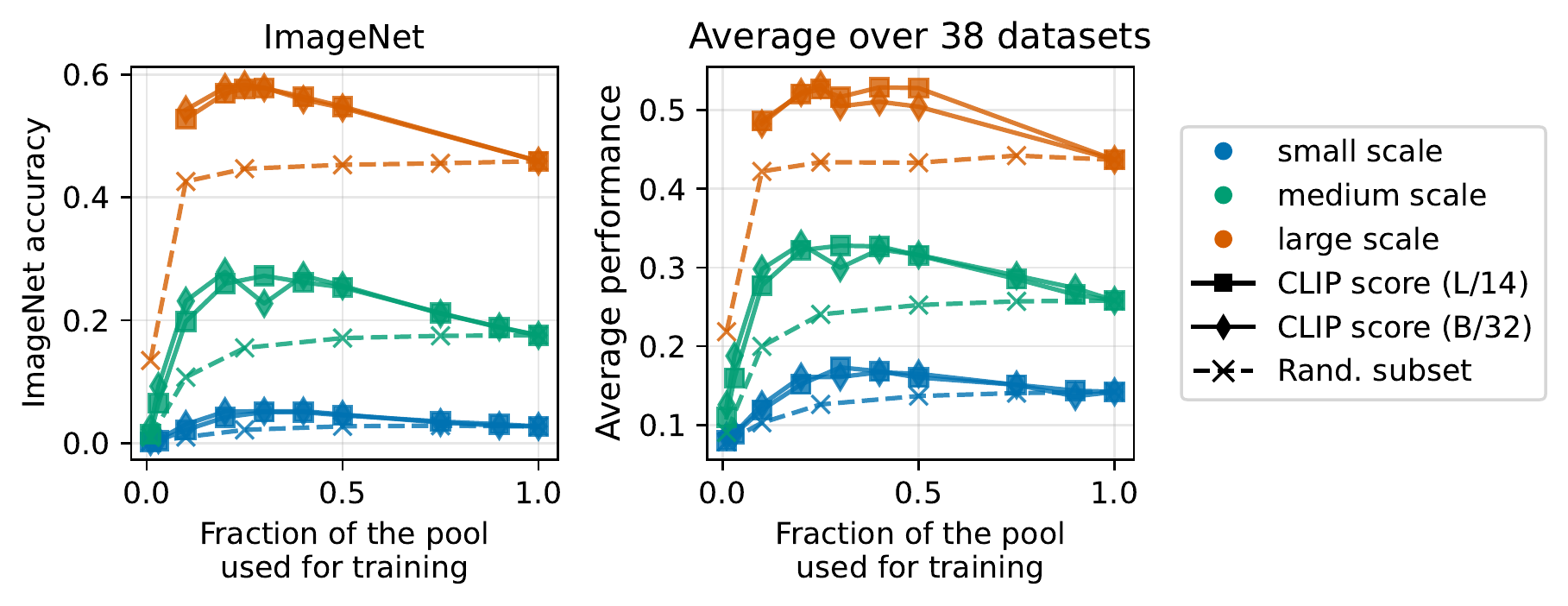}
    \caption{Performance of random subsets (dotted line) and CLIP score filtering (solid line) when varying the subset size. When taking random subsets, larger subsets are always better. For CLIP score filtering, subsets with intermediate size perform best.}
    \label{fig:clip-th}
    \vspace*{-2mm}
\end{figure}

\paragraph{Trade-off between data diversity and repetition.}
In Figure \ref{fig:clip-th}, we see that randomly selecting subsets of the pool has little effect and degrades performance substantially when only small fractions are used. 
When filtering with CLIP scores, the optimal training set comes from selecting $\sim$30\% of the pool with the highest scores.
The difference in performance trends between
random subsets and CLIP score filtering highlights the importance of filtering strategies for selecting samples.

\subsection{\datanet design analyses}
\paragraph{\pool and LAION are comparable with the same filtering.}
To validate our pool construction, we show that we can build datasets comparable to LAION-2B by employing their filtering technique on our pool.
LAION-2B selects all samples where the caption is in English and the cosine similarity score from a trained ViT-B/32 CLIP model is above 0.28. We compare this filtering approach on our pool using the same number samples, 130M samples at the {\small \texttt{large}} scale.
We find that the different data sources perform comparably: 55.3\% vs 55.7\% accuracy on ImageNet, and 0.501 vs 0.489 average performance over our evaluation sets using our pool and LAION-2B, respectively.

\begin{figure}
    \centering
    \includegraphics[width=\linewidth]{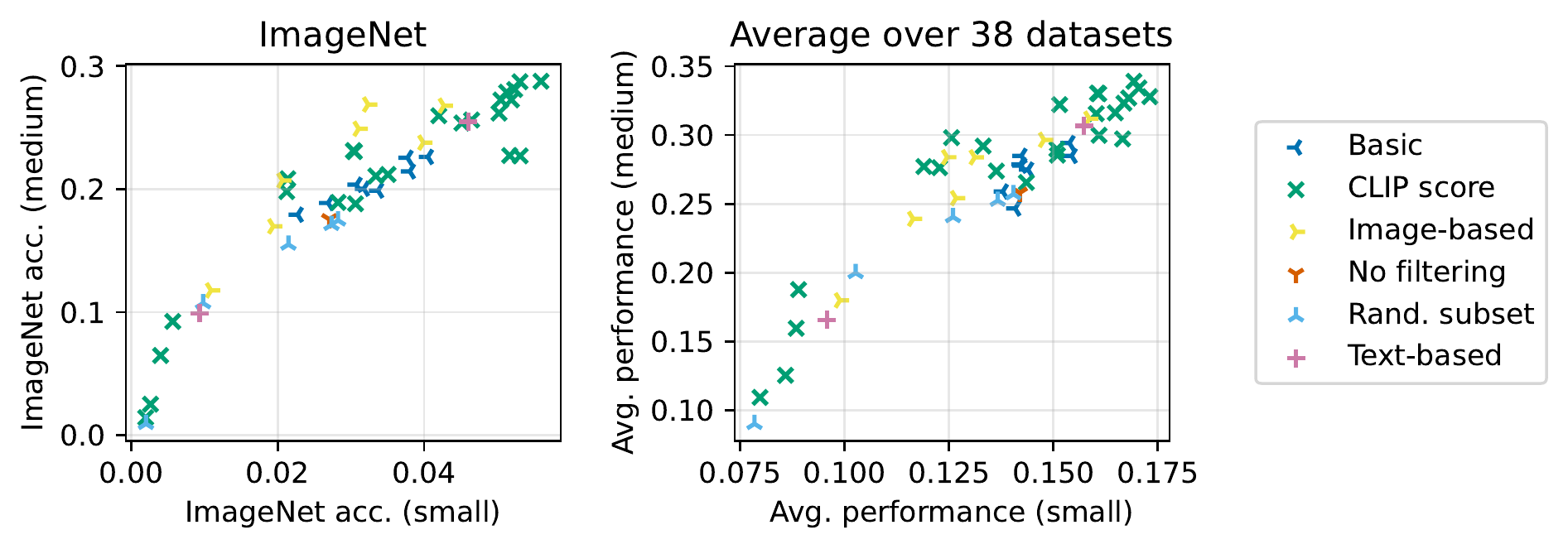}
    \caption{Correlation between {\small \texttt{small}} and {\small \texttt{medium}} scale baselines. Smaller scales can serve as useful guides for larger scales. Results for additional scales are shown in Appendix Figure \ref{fig:scaling-scatter-full}.
    }
    \label{fig:scaling-scatter}
    \vspace*{-2mm}
\end{figure}

\paragraph{Consistency across scales.} We find that the ranking between filtering strategies is typically consistent across different scales. This is illustrated in Figure~\ref{fig:scaling-scatter}, which shows that the baselines at {\small \texttt{small}} and {\small \texttt{medium}} scales are positively correlated. Moreover, as shown in Appendix Table \ref{tab:correlation}, the rank correlations of performance is high, between 0.71 and 0.90 for different scale pairs.

\paragraph{Consistency across training changes.} \datanet fixes the training procedure, so a natural question is whether better datasets from \datanet are better outside of \datanet. While \ours is trained at the {\small \texttt{xlarge}} scale, we show in Appendix Table~\ref{tab:b16} that even when substituting the ViT-L/14 for a ViT-B/16 or ViT-B/32, training on \ours outperforms training on OpenAI's WIT and LAION-2B. Additionally, we found that modifying hyperparameters such as training steps and batch size minimally affects the relative ordering of different data curation methods on downstream performance. Details on hyperparameter ablations are in Appendix \ref{app:hyperparam}.

\subsection{Evaluation trends}
\label{sec:eval-trends}
\paragraph{ImageNet accuracy is indicative, but not the complete picture.} Similarly to \citet{kornblith2019better}, in Appendix Figure~\ref{fig:imagenet-vs-all} we find that ImageNet performance is highly correlated with the average performance across all datasets we study, with an overall correlation of 0.99.
\footnote{Note that unlike \citet{kornblith2019better} we evaluate zero-shot performance rather than transfer learning.} However, ImageNet performance is not representative of all evaluation tasks, as the correlation between ImageNet accuracy and accuracy on other individual datasets varies substantially, in some cases even exhibiting a negative correlation, as discussed in Appendix \ref{sec:app-more-plots}.

\paragraph{Robustness and fairness.} While typical models trained on a target task suffer large performance drops under data distribution shift, zero-shot CLIP models are known to exhibit strong performance across many distributions \cite{radford2021learning}.
In Appendix Figure \ref{fig:robustness}, we show that CLIP models trained with data from our pool are more robust to distribution shift than ImageNet-trained models from
\citet{taori2020measuring}'s testbed.
Examining geographic diversity, we find that our models are better than ImageNet-trained models, but fall short of models fine-tuned on diverse curated datasets (see Appendix Figure~\ref{fig:robustness_diversity}).
We also perform a face classification analysis and identify demographic biases in our models: notably, the BYOD datasets we consider can increase the risk of misclassification.
See Appendix \ref{app:fairness} for more fairness and diversity analyses. 

\section{Limitations and conclusion}

\label{sec:conclusion}

In terms of societal risks, creating an index of image-text pairs from the public internet can be problematic. The internet contains unsafe, toxic, and sensitive content, which ideally should not percolate into machine learning datasets. Though we take steps to remove NSFW content and blur human faces to protect privacy, we hope future work will further explore the biases and risks from \pool and \ours.
We see several additional directions for future work, including
1) Curating more data sources. 
2) Improved data filtering algorithms.
3) Further supervision signals (e.g., image captions coming from captioning models).
4) Additional input modalities (e.g., video, 3D objects).
5) Broader evaluations for vision-and-language and robotics tasks.

Overall, we see \datanet as a first step towards improving training datasets, and hope our new benchmark will foster further research. 
By providing a controlled experimental setting, \datanet enables researchers to iterate on dataset design on rigorous empirical foundations.
We open-source all of our code, data, and infrastructure, and hope these resources will help the community build the next generation of multimodal datasets.

\input{acknowledgements}

\clearpage
\bibliography{main.bib}
\bibliographystyle{main.bst}

\newpage
\appendix
\onecolumn
\input{appendix_neurips}

\end{document}

%% file: acknowledgements.tex
\vspace*{-2mm}
\section*{Acknowledgements}
SYG and JH are supported by NSF Graduate Research Fellowships. GS is supported by the Onassis Foundation - Scholarship ID: F ZS 056-1/2022-2023. GD has been supported by the Onassis Fellowship (Scholarship ID: F ZS 012-1/2022-2023), the Bodossaki Fellowship and the Leventis Fellowship. This research has been supported by NSF Grants AF 1901292, CNS 2148141, DMS 2134012, TRIPODS II-DMS 2023166, Tripods CCF 1934932, IFML CCF 2019844 and research gifts by Western Digital, WNCG IAP, UT Austin Machine Learning Lab (MLL), Cisco, the Len Blavatnik and the Blavatnik Family Foundation, the Stanly P. Finch Centennial Professorship in Engineering, Open Philanthropy, Google, Microsoft, and the Allen Institute for AI.

We would like to thank Amro Abbas, Danny Bickson, Alper Canberk, Jessie Chapman, Brian Cheung, Tim Dettmers, Joshua Gardner, Nancy Garland, Sachin Goyal, Huy Ha, Zaid Harchaoui, Ari Holtzman, Andrew Hundt, Andy Jones, Adam Klivans, Ronak Mehta, Sachit Menon, Ari Morcos, Raviteja Mullapudi, Jonathon Shlens, Brandon McKinzie, Alexander Toshev, David Grangier, Navdeep Jaitly, Kentrell Owens, Marco Tulio Ribeiro, Shiori Sagawa, Christoph Schuhmann, Matthew Wallingford, and Ross Wightman for helpful feedback at various stages of the project. We are particularly grateful to Daniel Levy and Alec Radford for early encouragement to pursue this project and feedback on the experimental design.

We thank Stability AI and the Gauss Centre for Supercomputing e.V.\footnote{\url{https://gauss-centre.eu}} for providing us with compute resources to train models. We are thankful for the compute time provided through the John von Neumann Institute for Computing (NIC) on the GCS Supercomputer JUWELS Booster \citep{JUWELSBooster2020} at Jülich Supercomputing Centre (JSC), and for storage resources on JUST \citep{graf2021just} granted and operated by JSC, as well as computing and storage resources from the Helmholtz Data Federation (HDF).

%% file: appendix_neurips.tex
\if\arxivtng1
\else
\section*{Checklist}


\begin{enumerate}

\item For all authors...
\begin{enumerate}
  \item Do the main claims made in the abstract and introduction accurately reflect the paper's contributions and scope?
    \answerYes{}
  \item Did you describe the limitations of your work?
    \answerYes{See Section~\ref{sec:conclusion} and Section \ref{sec:eval-trends}.}
  \item Did you discuss any potential negative societal impacts of your work?
    \answerYes{See Section \ref{sec:conclusion}.}
  \item Have you read the ethics review guidelines and ensured that your paper conforms to them?
    \answerYes{}{}
\end{enumerate}

\item If you are including theoretical results...
\begin{enumerate}
  \item Did you state the full set of assumptions of all theoretical results?
    \answerNA{}
	\item Did you include complete proofs of all theoretical results?
    \answerNA{}
\end{enumerate}

\item If you ran experiments (e.g. for benchmarks)...
\begin{enumerate}
  \item Did you include the code, data, and instructions needed to reproduce the main experimental results (either in the supplemental material or as a URL)?
    \answerYes{Please see \href{https://www.datacomp.ai/}{\url{datacomp.ai}} for links to code, data, and instructions for reproduction.}
  \item Did you specify all the training details (e.g., data splits, hyperparameters, how they were chosen)?
    \answerYes{See Section \ref{sec:training} and Appendix \ref{app:train}.}
	\item Did you report error bars (e.g., with respect to the random seed after running experiments multiple times)?
    \answerNo{We do not report error bars due to computational constraints.}
	\item Did you include the total amount of compute and the type of resources used (e.g., type of GPUs, internal cluster, or cloud provider)?
    \answerYes{See Section \ref{sec:thecomp}.}
\end{enumerate}

\item If you are using existing assets (e.g., code, data, models) or curating/releasing new assets...
\begin{enumerate}
  \item If your work uses existing assets, did you cite the creators?
    \answerYes{}
  \item Did you mention the license of the assets?
    \answerYes{}
  \item Did you include any new assets either in the supplemental material or as a URL?
    \answerYes{}
  \item Did you discuss whether and how consent was obtained from people whose data you're using/curating?
    \answerYes{See Datasheet~\ref{datasheet:collection} (Q29-Q32).}
  \item Did you discuss whether the data you are using/curating contains personally identifiable information or offensive content?
    \answerYes{See Datasheet~\ref{datasheet:collection} (Q15-Q20).}
\end{enumerate}

\item If you used crowdsourcing or conducted research with human subjects...
\begin{enumerate}
  \item Did you include the full text of instructions given to participants and screenshots, if applicable?
    \answerNA{}
  \item Did you describe any potential participant risks, with links to Institutional Review Board (IRB) approvals, if applicable?
    \answerNA{}
  \item Did you include the estimated hourly wage paid to participants and the total amount spent on participant compensation?
    \answerNA{}
\end{enumerate}

\end{enumerate}
\clearpage
\fi


\part*{Appendix}
\tableofcontents
\newpage
\section{Benchmark rules}
\label{sec:full-competition-rules}

We provide concrete rules below for the two competition tracks that comprise \datanet: filtering and \byod.
Additionally, we provide a checklist, which encourages \users to specify design decisions, which allows for more granular comparison between submissions.

\subsection{Filtering track rules}

\begin{itemize}
    \itemsep0em 
    \item Participants can enter submissions for one or many different scales: \texttt{small}, \texttt{medium}, \texttt{large} or \texttt{xlarge}, which represent the raw number of image-text pairs in CommonPool that should be filtered.
    \item After choosing a scale, \users generate a list of uids, where each uid refers to a \pool sample. The list of uids is used to recover image-text pairs from the pool, which is used for downstream CLIP training.
    \item Duplicate uids are allowed.
    \item Participants are \emph{not} allowed to modify the training procedure. Hence, changing hyperparameters, model architecture, optimizer, compute budget, or number of training steps is not allowed. Changing any other training details is also not allowed.
    \item Participants are strongly encouraged to submit and open-source both the list of uids and the code used to generate this list; however, this is not required.
    \item To avoid overfitting, we do not permit running any code or algorithmic dependence on the test images of the evaluation tasks. However, use of other images associated with these tasks (e.g., supervised training sets) is permitted.
    \item Participants can use templates or class labels from the downstream tasks in their filtering algorithms.
\end{itemize}

For clarity, we include some examples of permitted and forbidden uses:
\begin{itemize}
\itemsep0em 
\item[\checkmark] We \textbf{permit} using the ImageNet class label ``triceratops'' in a filtering algorithm.
\item[$\times$] We \textbf{forbid} examining individual or aggregate predictions on the test sets of the evaluation tasks.
\end{itemize}

\subsection{Bring your own data track: amendments}
To facilitate more open-ended exploration, we provide amendments to the Track 1 competition to allow for more diverse submissions in Track 2.
\begin{itemize}
    \itemsep0em 
    \item Participants are allowed to augment \pool data with existing datasets, so long as these data sources do not contain test images from the evaluation tasks. Participants can use data from any \pool; however, they are not required to do so.
    \item Assembling one's own dataset is allowed; however, test images from the evaluation tasks can neither be contained nor otherwise used to construct said dataset. We encourage releasing the image urls or the images themselves in addition to the text for each image. We also encourage rigorous documentation of face-blurring and other data safety checks (see Section~\ref{sec:pool} for more details). We reserve the right to run our own safety code on participant provided data and disqualify entries that do not meet adequate safety standards.
\end{itemize}

\paragraph{Checklist.}
The following checklist provides the basis for more fine-grained comparison between submissions. 
\begin{itemize}
    \item[$\square$] Images from the evaluation tasks are included in my submission. If yes, please specify which datasets.
    \item[$\square$] I used an existing datasets (e.g., YFCC100M~\cite{yfcc100m}) in my submission. If yes, please specify which datasets. (Note: applies to \byod only)
    \item[$\square$] I curated my own data. If yes, please provide (1) image data or urls, (2) text for each image, (3) list of safety steps taken including but not limited to face blurring, explicit content image and text filtering. (Note: applies to \byod only)
\end{itemize}

\section{Contributions}
For this section, contributors are ordered alphabetically.

\subsection{Candidate pool}

\paragraph{Candidate pool lead.} Vaishaal Shankar

\paragraph{Data collection.} Romain Beaumont, Vaishaal Shankar

\paragraph{Pre-processing and metadata.} Giannis Daras, Alex Fang (content filtering lead), Samir Yitzhak Gadre (metadata lead), Ryan Marten (deduplication lead), Vivek Ramanujan, Vaishaal Shankar, George Smyrnis (face blurring lead)

\subsection{Participant tooling}

\paragraph{Participant tooling lead.} Gabriel Ilharco

\paragraph{Resharder.} Romain Beaumont, Yair Carmon, Alex Fang, Jonathan Hayase (lead), Gabriel Ilharco, Vivek Ramanujan, Vaishaal Shankar, Georgios Smyrnis

\paragraph{Training.} Mehdi Cherti, Gabriel Ilharco, Jenia Jitsev, Vivek Ramanujan, Georgios Smyrnis, Mitchell Wortsman (lead)

\paragraph{Evaluation.} Romain Beaumont, Yonatan Bitton, Mehdi Cherti, Dhruba Ghosh (lead), Gabriel Ilharco

\paragraph{Additional infrastructure.} Stephen Mussmann, Sarah Pratt

\subsection{Baselines}

\paragraph{Baselines lead.} Yair Carmon

\paragraph{Filtering track.} Yair Carmon, Rahim Enterazi, Alex Fang, Samir Yitzhak Gadre, Gabriel Ilharco, Kalyani Marathe, Thao Nguyen, Eyal Orgad (co-lead), Georgios Smyrnis, Mitchell Wortsman, Jieyu Zhang (co-lead)

\paragraph{BYOD track.} Gabriel Ilharco, Thao Nguyen

\paragraph{Experiment babysitting.} Alex Fang, Gabriel Ilharco, Samir Yitzhak Gadre

\subsection{Leadership and Advising}

\paragraph{Advising.} Romain Beaumont, Yair Carmon, Alexandros G.\ Dimakis, Ali Farhadi, Hannaneh Hajishirzi, Jenia Jitsev, Pang Wei Koh, Ranjay Krishna, Stephen Mussmann, Sewoong Oh, Alexander Ratner, Olga Saukh, Ludwig Schmidt, Vaishaal Shankar, Shuran Song, Richard Vencu

\paragraph{Leadership.}
Yair Carmon, Alexandros G.\ Dimakis, Jenia Jitsev, Sewoong Oh, Ludwig Schmidt, Vaishaal Shankar

\paragraph{Overall project lead.} Ludwig Schmidt

\clearpage
\section{Additional related work}
\label{sec:more-relatedwork}

Here we expand on the related work described in Section~\ref{sec:relatedwork}.

Image dataset safety is an active area of research, especially in the context of large-scale dataset construction. In addition to \citet{Birhane2021MultimodalDM}, who study problematic content in LAION-400M, \citet{Yang2019TowardsFD} study the ImageNet dataset and reveal limitations associated with the ImageNet curation strategy---with negative implications for downstream model fairness. \citet{Prabhu2020LargeID} also study the ImageNet dataset and find pornographic content. Both \citet{Birhane2021MultimodalDM} and \citet{Prabhu2020LargeID} survey ethical conundrums and harms that are borne out of improper dataset curation. In an effort to combat dataset toxicity, we conduct NSFW preprocessing (Section~\ref{sec:pool}, Appendix~\ref{app:nsfw}) and blur detected faces (Section~\ref{sec:pool}, Appendix~\ref{app:face}) during pool construction. We also conduct preliminary fairness evaluations (Section~\ref{sec:eval-trends}, Appendix~\ref{app:fairness}) for models trained on our data. We hope \pool will serve as a research artifact for future work examining dataset safety.

Beyond data selection, \citet{chan2022data} investigate the effects of dataset distribution on emergent properties of transformers, while \citet{fang2022data} look at the relationship between data and model robustness to distribution shifts. We hope our extensive evaluation suite comprised of 38 diverse tasks will facilitate similar studies when training multimodal models at large scale.

Others study how to reduce the burdens of training data annotation in the curation process. Classic approaches include distant supervision~\cite{hoffmann2011knowledge}, crowd-sourced labels~\cite{yuen2011survey}, heuristic rules~\cite{Awasthi2020Learning} and feature annotation~\cite{mann2010generalized}, among others. A recent line of work known as data programming or programmatic weak supervision~\cite{Ratner16,ratner2017snorkel,zhang2021wrench,zhang2022survey} attempts to reduce annotation cost and is found in many industry applications~\cite{bach2019snorkel,Overton}. In data programming, developers write programmatic labeling functions to automatically label a large amount of unlabeled data. The labeling functions could produce noisy and conflicting labels, so researchers have developed methods to aggregate noisy votes to produce the final training labels~\cite{Ratner19,fu2020fast,shin2021universalizing}.

Previous literature also studies methods for training data attribution, which seek to link a model's behavior (e.g., its accuracy on a particular task or subset of data) to particular subsets of its training data. 
Such methods include influence functions, a classic technique from robust statistics \citep{hampel1974influence,cook1977detection} that uses a second-order Taylor expansion to approximate the effect of removing a training point on the learned model parameters
\citep{koh2017understanding,koh2019accuracy,han2020explaining,guo2020fastif}, as well as methods that fit attribution functions directly to the dynamics of repeated training runs \citep{ghorbani2019data,pruthi2020estimating,ilyas2022datamodels,guu2023simfluence}.
Training data attribution methods assume that we have already trained a model, though they can be subsequently used to refine the training data (e.g., by identifying potentially mislabeled training points \citep{koh2017understanding}).
Our focus in this paper is instead on data curation methods---that is, methods for selecting a subset of the training data to train a model in the first place.

In the context of natural language processing, \citet{dataset-cartography} proposes a tool for characterizing samples in a dataset based on training dynamics, labelling instances as ambiguous, easy to learn or hard to learn. Previous literature such as work by \citet{le2020adversarial,li2019repair,gururangan-etal-2018-annotation} advocate for removing easy instances from the training data. 
\citet{ethayarajh2022understanding} propose a measure of how difficult a dataset is to learn, $\mathcal{V}$-usable information. Such techniques could be promising directions of further exploration in the context of our benchmark.

Finally, another related line of work is studying scaling trends. In addition to \citet{sorscher2022beyond}, researchers have investigated how model performance changes as a function of compute budget, model size, and number of training samples \cite{kaplan2020scaling,chinchilla,caballero2022broken,cherti2022reproducible}.
However, this line of work does not consider how dataset design may affects scaling trends. 
Beyond dataset size, we measure the effects of different dataset sources and filtering strategies.
While scaling trends are central to our investigations, the purpose of our benchmark is to search for the next generation of large multimodal datasets to facilitate more accurate and reliable models.

\section{Parsing Common Crawl}
\label{app:parse-cc}
Common Crawl releases metadata files for the websites that they index (i.e., WAT files). They release these files approximately once a month. We consider all files available from 2014 through November of 2022. We first parse these files, utilizing Apache Spark~\cite{zaharia2016apache} to extract image urls and corresponding alt-text. We map each url, text pair to a uid hash and remove duplicates. This results in 88 billion url, text pairs, which are randomized via a distributed shuffle. Note, we do not consider image content when running uid deduplication at this step. Hence, two identical images with different urls and the same caption would both be retained.

\section{Not safe for work (NSFW) filtering}
\label{app:nsfw}
Our data is sourced from Common Crawl, which contains snapshots of the web. Therefore, we apply multiple layers of NSFW content filtering to remove problematic images and captions from \pool.

First, we filter our captions with Detoxify \citep{Detoxify}, a language model for toxic comment classification. Specifically, we use the multilingual XLM-RoBERTa~\cite{Conneau2019UnsupervisedCR} variant. The model outputs scores between zero and one for the following categories: toxicity, severe toxicity, obscene, identity attack, insult, threat, and sexually explicit. As we had no ground truth for our data, we manually spot check a 1 million random subset of \pool at varying thresholds. We found that a threshold of 0.1 provided good coverage of filtering out NSFW text.
If any of the detoxify category scores exceeds the threshold, the sample is discarded.
Qualitatively, we found that the model struggled with multilingual content, acronyms, and innuendo.
Even at 0.1, we noticed there are some captions that are NSFW.
However, lowering the threshold further heavily affected false positives. We therefore use a 0.1 threshold for all NSFW categories, which on a random subset of one million captions achieves positive rates shown in Table~\ref{tab:detoxify}. 

\begin{table*}
\renewcommand{\arraystretch}{1.1}
\rowcolors{2}{white}{light-light-gray}
    \caption{Detoxify positive rates by threshold on 1 million caption subset of Common Crawl.}
    {
    \centering
    \resizebox{\textwidth}{!}{

    \begin{tabular}{cccccccc}
    \toprule
    Threshold & Toxicity & Severe Toxicity & Obscene & Identity Attack & Insult & Threat & Sexual Explicit   \\ \midrule
    0.01       & 9.5\%   & 1.0\%     & 33.4\% & 1.8\% & 35.0\% & 1.3\% & 2.0\% \\
    0.1       & 3.6\%   & 0.1\%     & 0.8\% & 0.3\% & 1.4\% & 0.1\% & 1.0\% \\ \bottomrule
    \end{tabular}}\par
    }
    \label{tab:detoxify}
\end{table*}

Second, on the vision side, we use a modified version of LAION-5B's \citep{laion5b} CLIP-based binary classification NSFW model, which takes CLIP ViT-L/14 visual embeddings as input. We remove the initial multi-category encoder from the model, and retrain on the same data with an initial normalization layer followed by a 4-layer multilayer perceptron. Our retrained model matches the performance of the original model on their manually annotated testset. Specifically, we achieve 97.4\% classification accuracy on a held out test set compared to  96.1\% for the original LAION NSFW image filtering model. Additional details about the training data can be found in Appendix C.5 of the LAION-5B paper. In brief, the training data contains 682K images that is roughly balanced with images from safe for work and NSFW categories.

\begin{table*}
\renewcommand{\arraystretch}{1.1}
\rowcolors{3}{light-light-gray}{white}
    \caption{Comparing LAION-2B CLIP based NSFW filtering model to Google Vision API Safe Search adult category on a 40,000 random subset of Common Crawl.}
    {
    \centering
    \resizebox{\textwidth}{!}{
    \begin{tabular}{ccccc}
    \toprule
     & False Positive Rate & True Positives &  &      \\ 
   \multirow{-2}{*}{Threshold} &  (Relative to Google) & (Manual Review) & \multirow{-2}{*}{Model Positive Rate} & \multirow{-2}{*}{Google API Positive Rate} \\\midrule
    
    0.1       & 3.6\%  &      2       & 14.4\%                & 3.5\% \\
    0.2       & 0.6\%  &      2       & 9.1\%                 & 3.5\% \\
    0.3       & 0.3\%  &      3       & 7.2\%                 & 3.5\% \\ \bottomrule
    \end{tabular}}\par
    }
    \label{tab:nsfw}
\end{table*}

To evaluate our model and determine a threshold, we used Google Vision API's SafeSearch explicit content detector to generate labels for an 40,000 random subset of our candidate pool. Specifically, an image is NSFW if SafeSearch classifies it as likely or very likely adult (i.e., sexually explicit). As shown in Table~\ref{tab:nsfw}, we found that by thresholding at 0.1 we achieve high recall relative to SafeSearch and very few true positives after manual review. We also manually reviewed images classified by SafeSearch as likely or very likely racy and found that the images were either benign, subjectively suggestive but not explicit, or already found in the set of images labeled as adult.

\section{Deduplication against evaluation sets}
\label{app:dedup}

\begin{figure*}
    \centering
    \includegraphics[width=\textwidth]{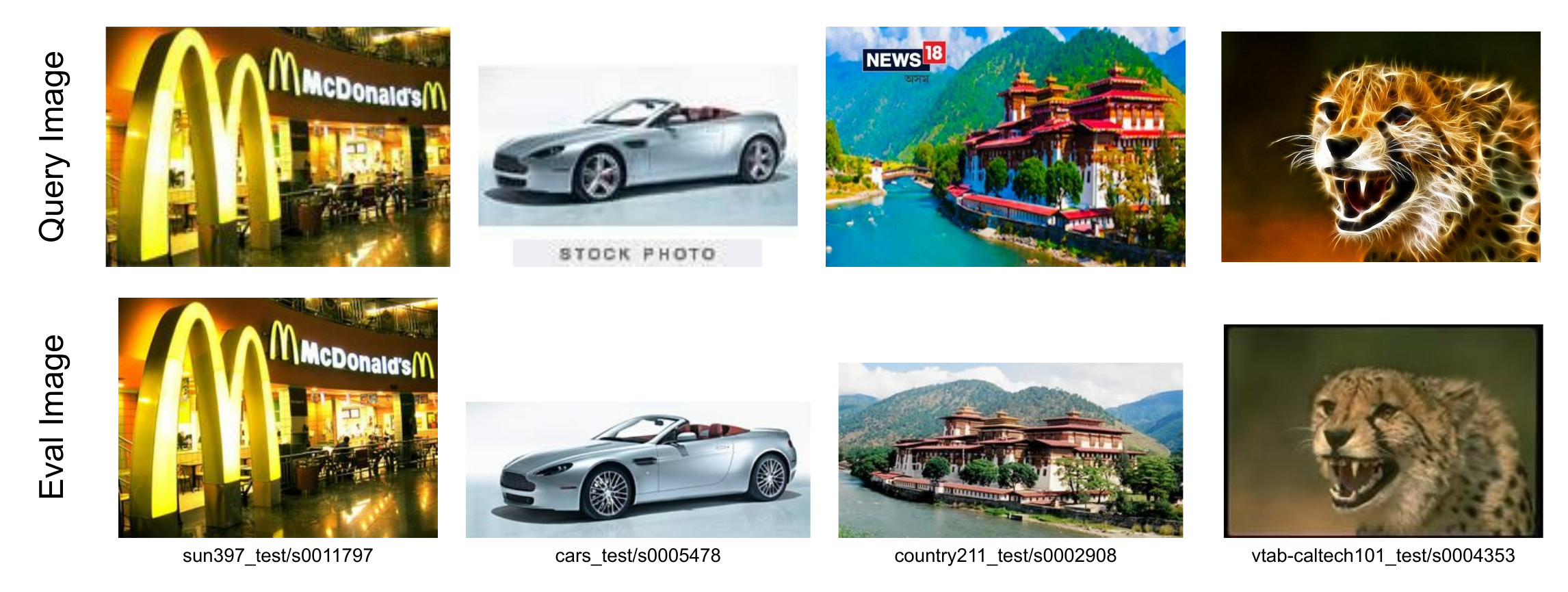}
    \caption{ Candidate images (top) that are detected as duplicates against images in the evaluation sets (bottom) are removed from the pool. In addition to exact duplicate images, near-duplicates with variable aspect ratios, JPEG compression, overlays, color adjustment, and artistic rendering are also detected. }
    \label{fig:dedups}
\end{figure*}

To prevent data leakage, we filter \pool by removing duplicate and near-duplicate matches of evaluation set images. See Figure \ref{fig:dedups} for example query images from Common Crawl and corresponding near-duplicates in our evaluations sets. We consider images as duplicates when the cosine similarity between a query (Common Crawl image) feature and a reference (evaluation image) feature is higher than a fixed threshold. We employ the deduplication model proposed by \citet{Yokoo2021Dedup}, which earned 1st place in the Facebook AI Image Similarity Challenge (ISC) \cite{douze2021isc}. We choose a cosine similarity threshold of 0.604169 to maximize the true duplicates detected, without removing too many false duplicates from the pool. We compare against OpenAI's CLIP ViT-B/32 as a baseline on ISC. We find that for our threshold, the ISC model achieves precision 0.9 and recall 0.8. At a threshold of 0.96, CLIP achieves the same precision 0.9, but a significantly worse recall of 0.02. Approximately 2.8\% of downloaded samples are flagged as evaluation set near-duplicates.

\begin{figure}[t]
    \centering
    \includegraphics[width=0.48\linewidth]{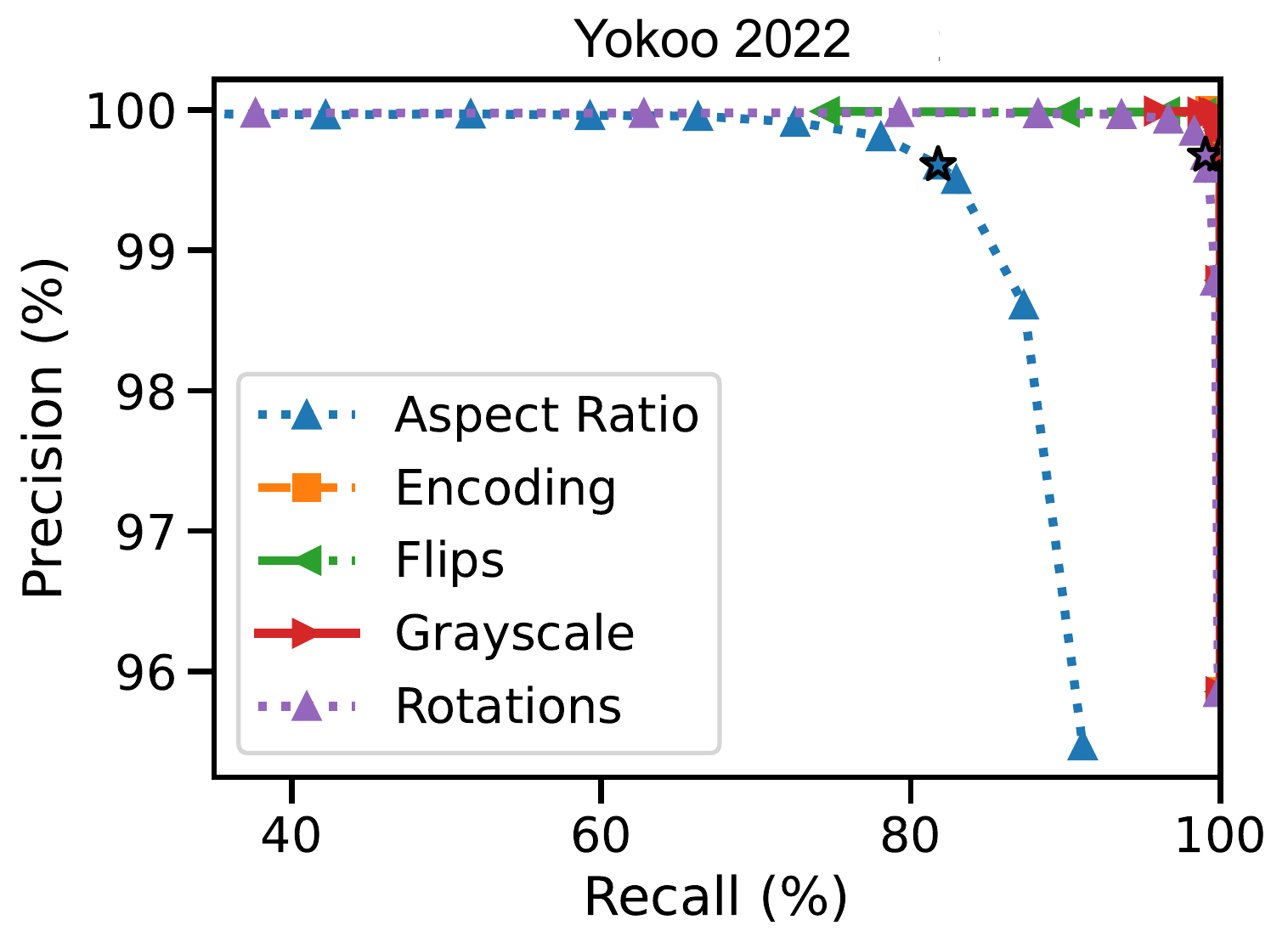}
    \includegraphics[width=0.48\linewidth]{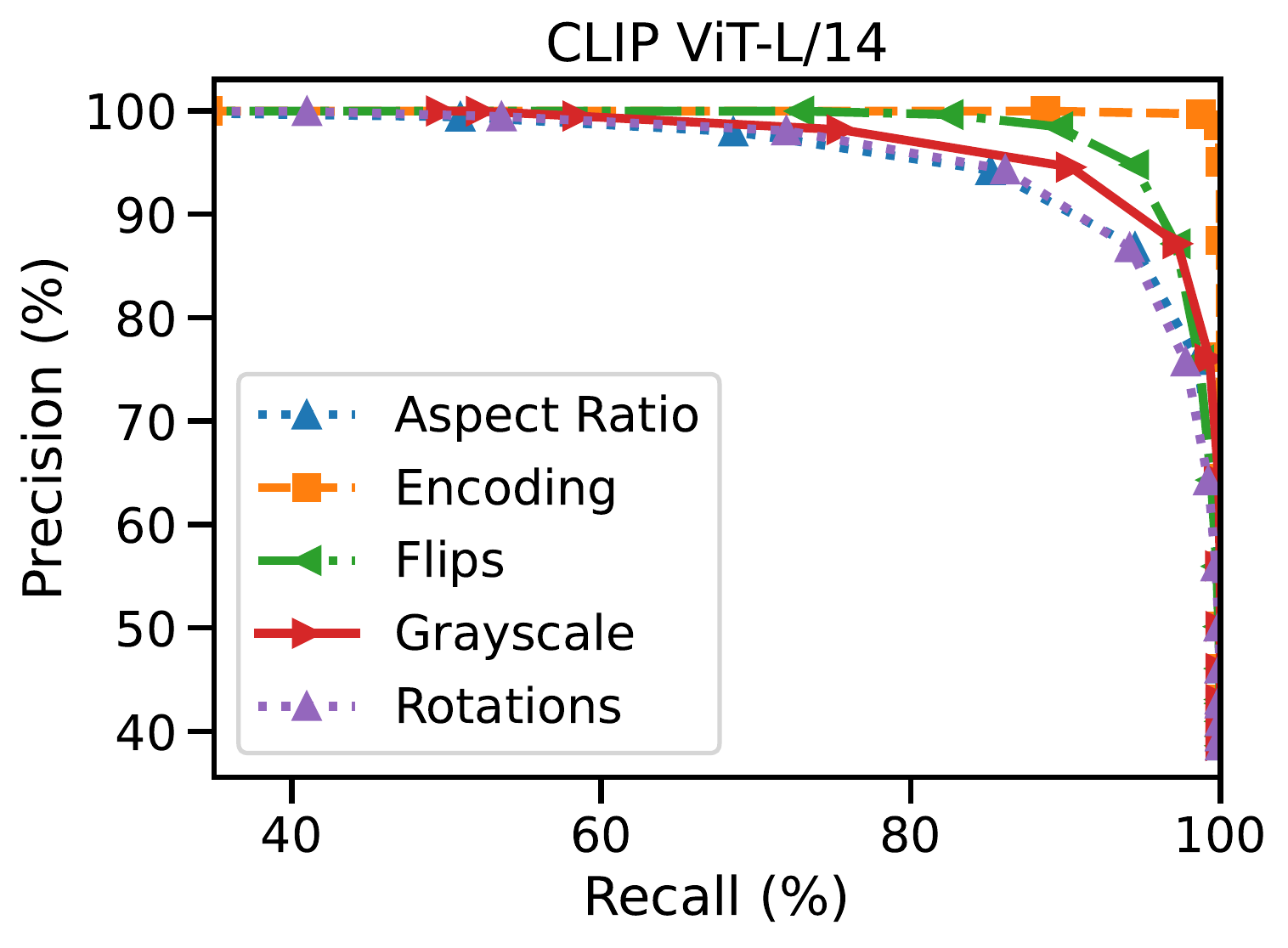}
    \caption{Analysis of different de-duplication strategies across a variety of image transformations. We see that the model introduced by \citet{Yokoo2021Dedup} is better in almost every transformation, with the exception of very aggressive aspect ratio modification.}
    \label{fig:dedup-analysis}
\end{figure}

To verify the performance of our de-duplication models with greater granularity, we modify the evaluation procedure in \citet{douze2021isc} to include transformations which are representative of naturally-occurring duplications on the Internet. Specifically, we study: 1) jpeg compression (encoding), 2) image flips, 3) image rotations, 4) aspect ratio modifications, and 5) grayscaling. To do this, we sample 20\% of the images from each of our evaluation datasets uniformly at random to serve as a reference set of about 140,000 images. Next we sample 560,000 images uniformly at random from LAION-2B to serve as distractors, for a 4-to-1 distractor to reference ratio. Finally, we apply each of the augmentations above and use threshold filtering to determine duplicates. Figure~\ref{fig:dedup-analysis} shows the results from the deduplication model \cite{Yokoo2021Dedup} compared with OpenAI's CLIP ViT-L/14. At high recall values, we see that CLIP filtering results in removing over 2$\times$ the data as that of the deduplication model from \citet{Yokoo2021Dedup}.

\section{Face blurring}
\label{app:face}

As an extra step to safeguard against issues of privacy that may arise from the use of data scraped from the web, we include face blurring as part of our pool creation. To create face metadata, we use the SCRFD face detector \cite{guo2021sample} to extract bounding boxes for the faces in our images. These bounding boxes are included as part of the image metadata in our pool. We make use of the pretrained SCRFD-10G model. We use the same preprocessing as the one described in the official repository of the paper, with the exception of providing $224 \times 224$ input images (by padding each image to square and then resizing) to limit computation costs. Invoking this model provides us with bounding boxes along with an associated score, which we then compare against a threshold of $0.3$ to keep or discard this bounding box. This threshold is the default one used in the repository of SCRFD for the visualization of bounding boxes, and we found it to perform well on our data as discussed next.

In Table \ref{tab:face_detection} we can see the result of face detection on a set of 3293 images from \pool. We evaluate the detection on whether the image has visible faces or not (where images such as cartoon drawings of non-real human faces are not considered as positives), and whether the detector has detected these visible faces. We considered an image as a true positive if all the clearly visible faces in the image were detected, based on the above thresholding process.
We did not do extensive box labeling. True positives are instead determined by human inspection.
We compare the quality of these detections with the Amazon Rekognition system, which is the one upon which the face detections on ImageNet were based \cite{yang2022study}. Note that in this scenario, the recall of the detectors is more important than precision (as detecting a few more bounding boxes across our pool does not affect privacy).

\begin{table*}
\renewcommand{\arraystretch}{1.1}
\rowcolors{2}{white}{light-light-gray}
    \centering
    \caption{Face detection performance on a set of 3293 random images from \pool.}
    \label{tab:face_detection}
    \begin{tabular}{lrr}
    \toprule
    ~ & SCRFD-10G & Amazon Rekognition \\ \midrule
    Accuracy & 93.87 & 96.57 \\
    Precision & 75.87 & 86.09 \\
    Recall & 90.53 & 93.75 \\ \bottomrule
    \end{tabular}
\end{table*}

\begin{figure*}
    \centering
    \includegraphics[width=0.6\linewidth]{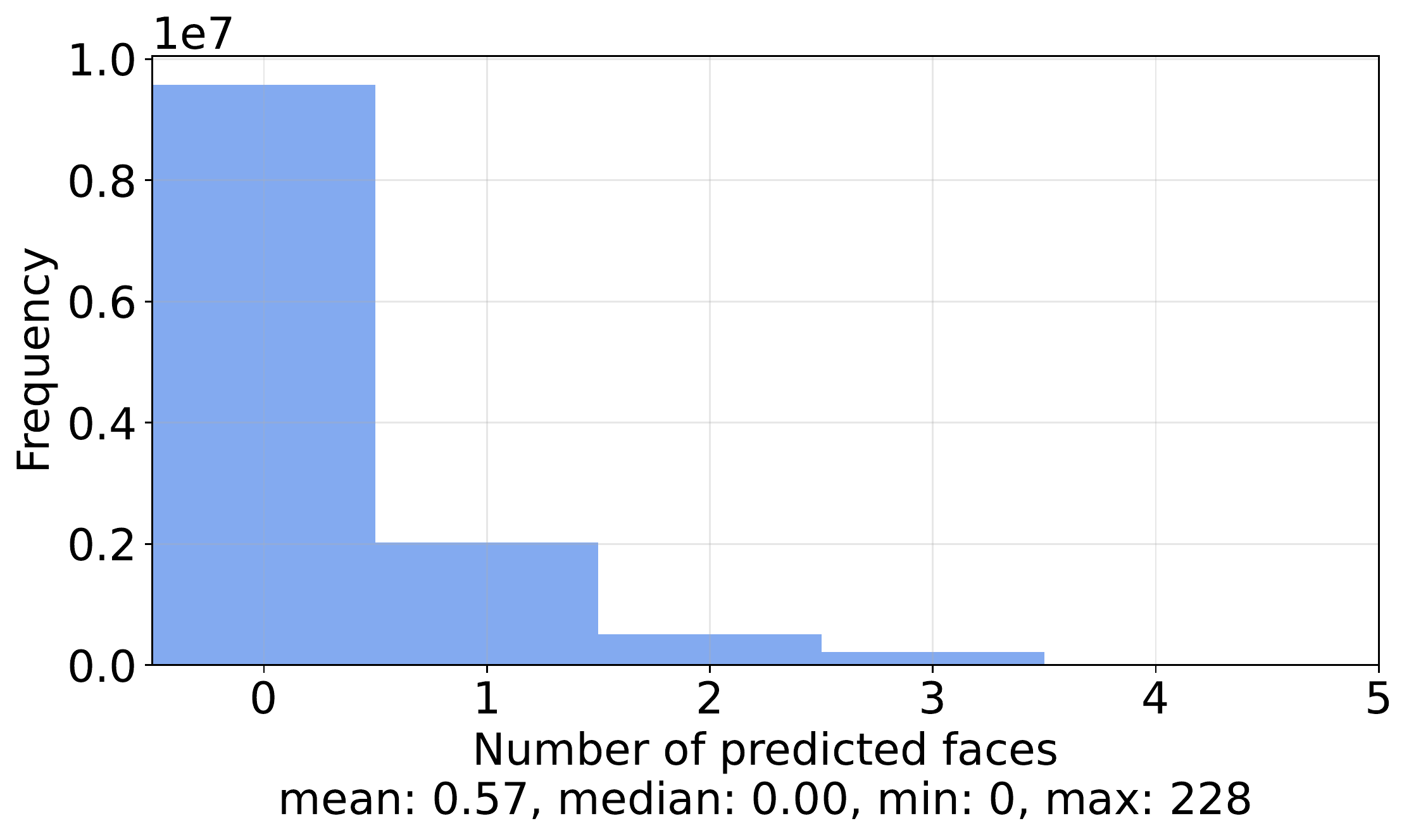}
    \caption{Frequency of predicted number of faces in the {\small \texttt{small}} \pool.}
    \label{fig:faces}
\end{figure*}

To utilize these bounding boxes on our data, we apply a standard blurring pipeline, as proposed by \citet{yang2022study}. The result of this process is an image where the faces is blurred and there is a smooth transition from blurred to clean parts of the image.
In Figure \ref{fig:faces} we see the distribution of faces for the {\small \texttt{small}} \pool. Note that the majority of images do not contain faces.

As part of our competition pipeline, images are by default blurred during the download process. In Table \ref{tab:face_blur_test} we can see the results of training on a set of images with the size of our {\small \texttt{medium}} scale after filtering with each method, with and without the application of face blurring as provided by our detector. We can see that the difference in performance is small, which suggests that the application of face blurring does not significantly affect the performance on our downstream tasks.
However, we note that this design decision may be more detrimental in generative settings, especially when a generative model needs to output faces. Our competition is primarily focused on discriminative tasks, and as such when designing our dataset, we wished to prioritize the safety and privacy of individuals through  blurring faces in our download tooling by default. 

Finally, we evaluated the detector we used for potential biases. More specifically, we used the detector on the validation set of the FairFace dataset \cite{karkkainen2021fairface}. We found that the central face of the image was detected in all the images of the validation set, regardless of subgroup annotate in the dataset.

\begin{table*}
    \centering
    \renewcommand{\arraystretch}{1.1}
    \rowcolors{2}{white}{light-light-gray}
    \caption{Effect of face blurring on zero-shot performance. Face blurring improves the privacy preservation of our dataset, while affecting model performance negligibly. Results shown for training on a set of images with the size of our {\small \texttt{medium}} scale, after filtering with each method.}
    \label{tab:face_blur_test}
    \resizebox{\textwidth}{!}{

    \begin{tabular}{lccc}
    \toprule
    Filtering & Face blurring & ImageNet acc. & Avg. performance \\ \midrule
    \cellcolor{white} & $\times$ & 0.209 & 0.246 \\
    \cellcolor{white}\multirow{-2}{*}{CLIP score (B/32, thresh. 0.3) + English filtering} & \checkmark & 0.196 & 0.243 \\ \midrule
   \cellcolor{white} & $\times$ & 0.287 & 0.301 \\
   \cellcolor{white} \multirow{-2}{*}{CLIP score (B/32, 30\%)}  & \checkmark & 0.282 & 0.298 \\ \bottomrule
    \end{tabular}}
\end{table*}

\newpage
\section{\datanet \pool creation pipeline}
\label{app:metadata}

\begin{figure*}[h]
    \centering
    \includegraphics[width=0.9\linewidth]{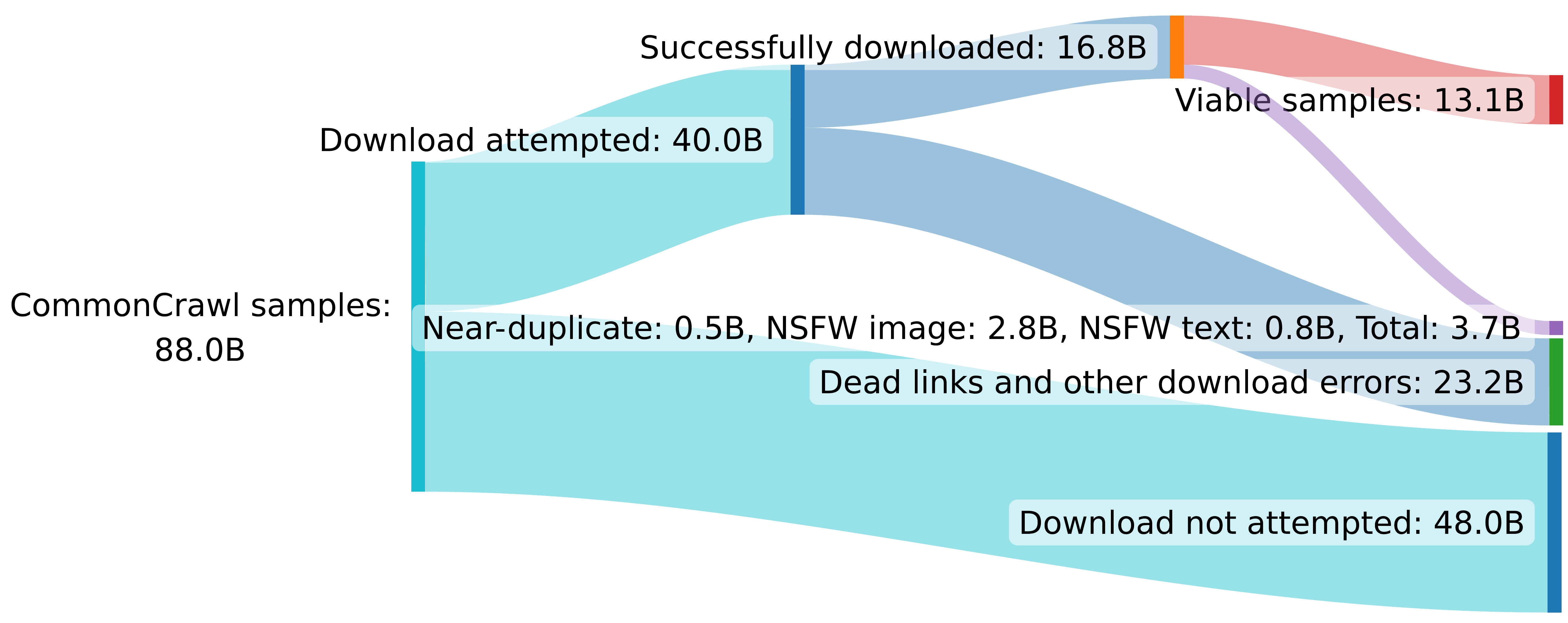}
    \caption{Data funnel from potential samples in Common Crawl to 13.1B image-text pairs that were suitable for \pool. We sampled uniformly 12.8B datapoints for the \small{\texttt{xlarge}} \pool.}
    \label{fig:data_pipeline}
\end{figure*}

\begin{table*}[h]
\small
    \centering
    \renewcommand{\arraystretch}{1.1}
        \rowcolors{2}{white}{light-light-gray}
    \caption{Provided metadata for \pool.}
    {
    \centering
    \begin{tabular}{llr}
    \toprule
    Generation Time & Label & Additional notes  \\ \midrule
     \cellcolor{white}& uid & ~ \\
    \cellcolor{white} & url & Link to the image. \\
    \cellcolor{white} & text & Image caption. \\
    \cellcolor{white}& original\_width & ~ \\
    \cellcolor{white} & original\_height & ~ \\
    \multirow{-6}{*}{Step 2} & sha256 & Safeguard for data poisoning. \\ \midrule
    \cellcolor{white} & clip\_b32\_similarity\_score & ~ \\
    \cellcolor{white} & clip\_b32\_image\_features & In separate file.\\
    \cellcolor{white} & clip\_b32\_text\_features & In separate file. \\~ & clip\_l14\_similarity\_score & ~ \\
    \cellcolor{white} & clip\_l14\_image\_features & In separate file. \\~ & clip\_l14\_text\_features & In separate file.\\
    \cellcolor{white}\multirow{-7}{*}{Step 1} & face\_bboxes & ~ \\ \midrule
    \cellcolor{white} & nsfw\_image\_score & ~ \\
    \cellcolor{white} & nsfw\_text\_score & ~ \\
    \cellcolor{white}\multirow{-3}{*}{Step 2, dropped during Step 3} & dedup\_score & ~ \\ \bottomrule
    \end{tabular}
    }
    \label{tab:app_metadata}
\end{table*}
\FloatBarrier

Creating \pool was a multistep process, which involved (1) parsing image urls and alt-text from Common Crawl dumps and downloading these images, (2) tagging images with metadata and (3) conducting safety content filtering and evaluation set duplication. In this section we provide an overview of the data pipeline used to create \pool. For an overview of our ``data funnel'' see Figure \ref{fig:data_pipeline}.

\begin{enumerate}
    \item For the first step, we use parse Common Crawl metadata files to harvest image-text pairs (Section \ref{app:parse-cc}). We use \texttt{img2dataset}~\cite{img2dataset} to obtain $\sim$16.8B downloaded samples. This is the first, unfiltered version of \pool, and contains only basic information for our images (i.e., the original image height, width, and alt-text caption). During this step we also resize images such that their largest dimension does not exceed 512 pixels.
    This eases storage requirements for large images, but is still larger than the 224 pixel resolution used for later training stages.
    
    \item For the second step, we process our unfiltered pool and create richer metadata for each image-text pair. We generate the following for each sample:
    \begin{itemize}
        \item CLIP ViT-B/32 and CLIP ViT-L/14 image and text features, with their associated similarities.
        \item NSFW scores for the image and the text, using the analysis described in Appendix \ref{app:nsfw}.
        \item Deduplication score for the image, as described in Appendix \ref{app:dedup}.
        \item Bounding boxes for faces detected in the image, using the method described in Appendix \ref{app:face}.
    \end{itemize}

    \item For the third and final step, we filter our image-text pairs based on the metadata generated during the second stage. We filter out image-text pairs where the NSFW and deduplication scores exceed the respective thresholds (Section \ref{app:nsfw}). From the images that pass through this filtering, we keep only the desired amount (e.g., 12.8B images from the {\small \texttt{xlarge}} \pool).
    Smaller pools are telescoping subsets of larger pools.
    We package the metadata and image urls, which is made publicly available to the \users.
    Note, we do not release raw image data but rather image urls pointing to images.
\end{enumerate}

A summary of the metadata for each sample is found in Table \ref{tab:app_metadata}. 
To validate our pipeline for duplication and CLIP feature correctness, we also take ImageNet train though metadata generation as a unit test. Using the deduplication features, we detect that 100\% of the images are in fact duplicates. Additionally using the CLIP ViT-B/32 and CLIP ViT-L/14 image features and corresponding text features from OpenAI's 80-prompt ensemble, we achieve 63.36\% and 75.54\% top-1 accuracies, which match the performance reported in the CLIP paper~\cite{radford2021learning}.

When creating pools of different scale (i.e., number of samples), we ensure that smaller pools are subsets of larger pools. For instance, the {\small \texttt{small}} \pool is a subset of the {\small \texttt{xlarge}} \pool.

After \pool is created, the \users can then download the final image-text pairs using the provided files via \texttt{img2dataset}. To further ease the computational burden on \users, we additionally provide metadata for each sample in \pool. Note that when downloading, our \texttt{img2dataset} configuration automatically blurs faces. Hence this is an automatic step on not something \users must do ad hoc.

\section{\pool statistics}
\label{app:pool-stats} 

To provide more information about the kinds of samples in our \pool, we conduct additional analysis on the {\small \texttt{small}} pool, which is an i.i.d. sample of downloaded data and a subset of the larger pools.

\begin{figure*}
    \centering
    \includegraphics[width=\linewidth]{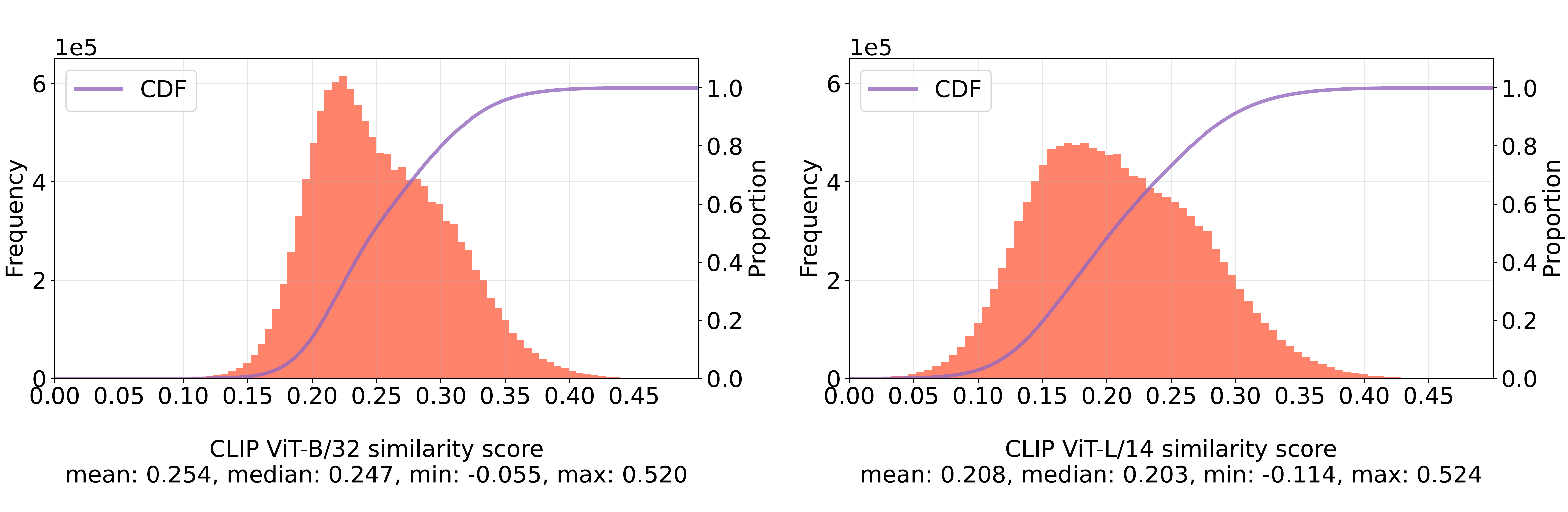}
    \caption{Image-text similarity score distributions using CLIP ViT-B/32 \emph{(left)} and ViT-L/14 \emph{(right)} models. We plot samples from the \texttt{small} \pool, which are an i.i.d. sample of the \texttt{xlarge} \pool.}
    \label{fig:clip_sim}
\end{figure*}

In Figure \ref{fig:clip_sim} we show CLIP similarity similarity scores between images and their corresponding text. We notice a flatter distribution of CLIP ViT-L/14 scores than corresponding B/32 scores.

\begin{figure*}
    \centering
    \includegraphics[width=\linewidth]{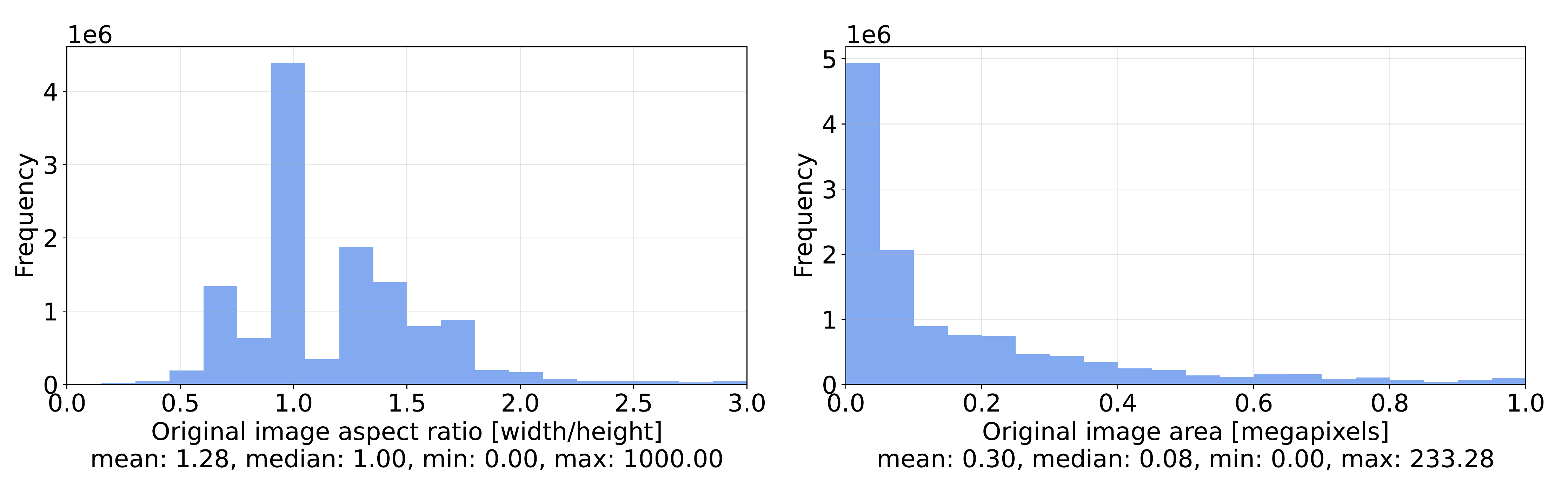}
    \caption{Statistics for images in the \texttt{small} \pool, before applying resizing.}
    \label{fig:image_stats}
\end{figure*}

\begin{figure*}
    \centering
    \includegraphics[width=0.4\linewidth]{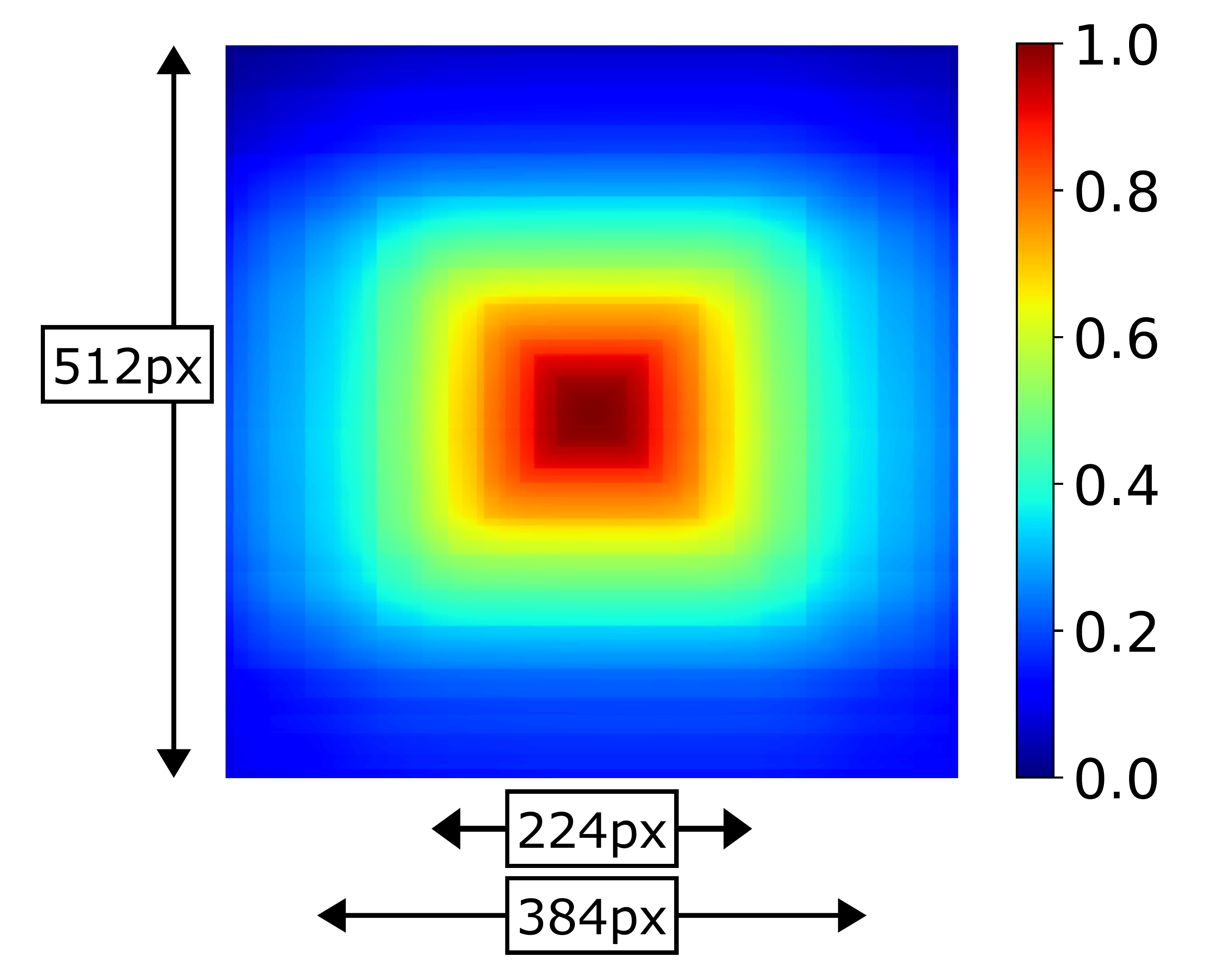}
    \caption{\textbf{Image pixel heatmap.} Each entry in the above heatmap represents the estimated probability that a pixel is occupied. The center entry has a value of 1.0 as every image has a center pixel. We compute the heatmap over the {\small \texttt{small}} \pool. Note that image sizes are bounded as we resize all images such that their max dimension does not exceed 512 pixels during dataset download.}
    \label{fig:resize_distrib}
\end{figure*}

Turning our attention to images in \pool, in Figure \ref{fig:image_stats}, we visualize the aspect ratios and sizes of original images (i.e., before they are downloaded and resized).
In Figure \ref{fig:resize_distrib}, we display a distribution of image height and width after \emph{download} resizing. Notice that the majority of images are around $224 \times 224$ pixels, which is the final resized resolution used for training.

\begin{figure*}
    \centering
    \includegraphics[width=0.7\linewidth]{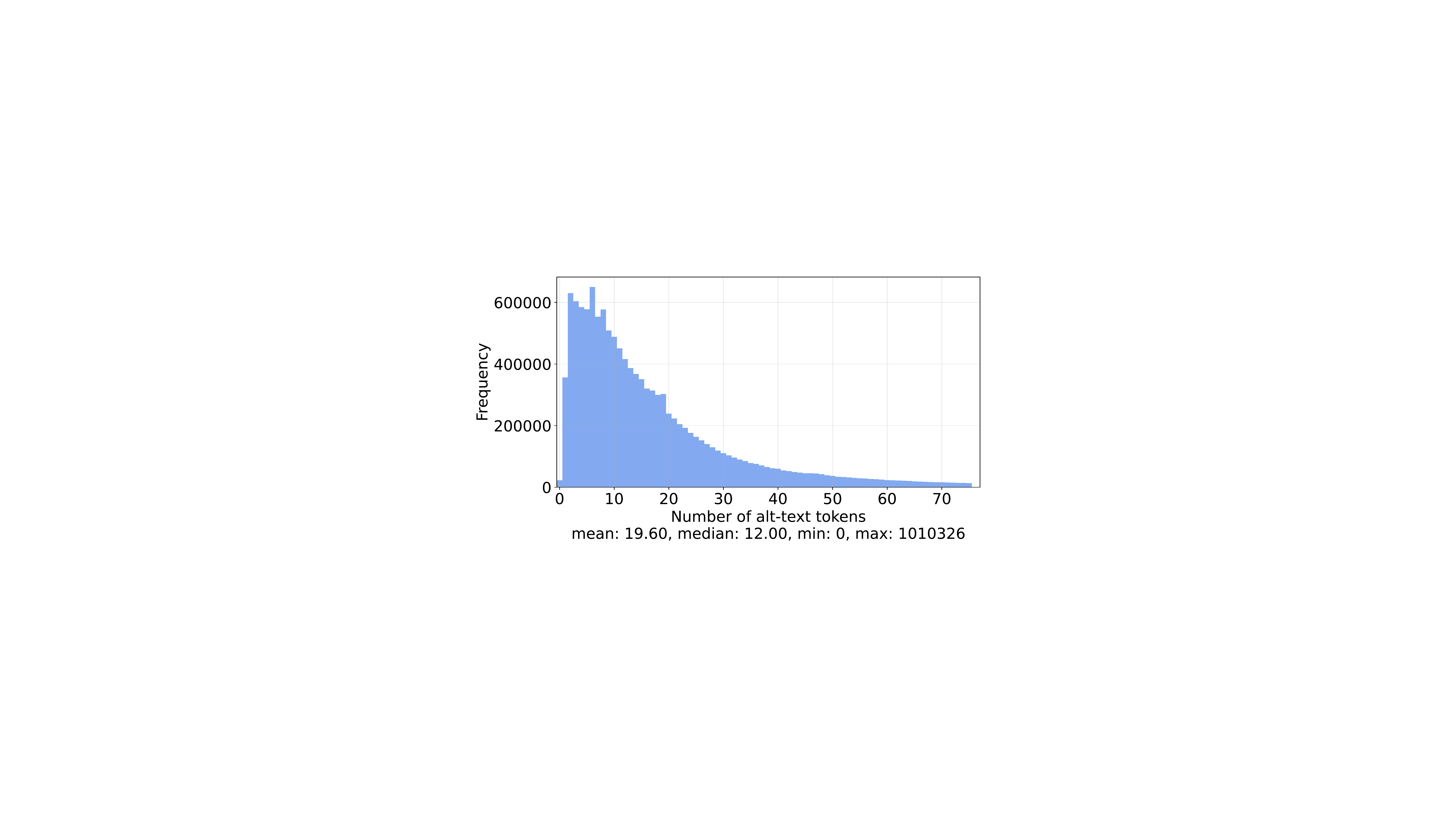}
    \caption{Distribution of token length for alt-text in the {\small \texttt{small}} \pool. The CLIP BPE tokenizer is used for tokenization.}
    \label{fig:tokens}
\end{figure*}

\begin{figure*}
    \centering
    \includegraphics[width=0.9\linewidth]{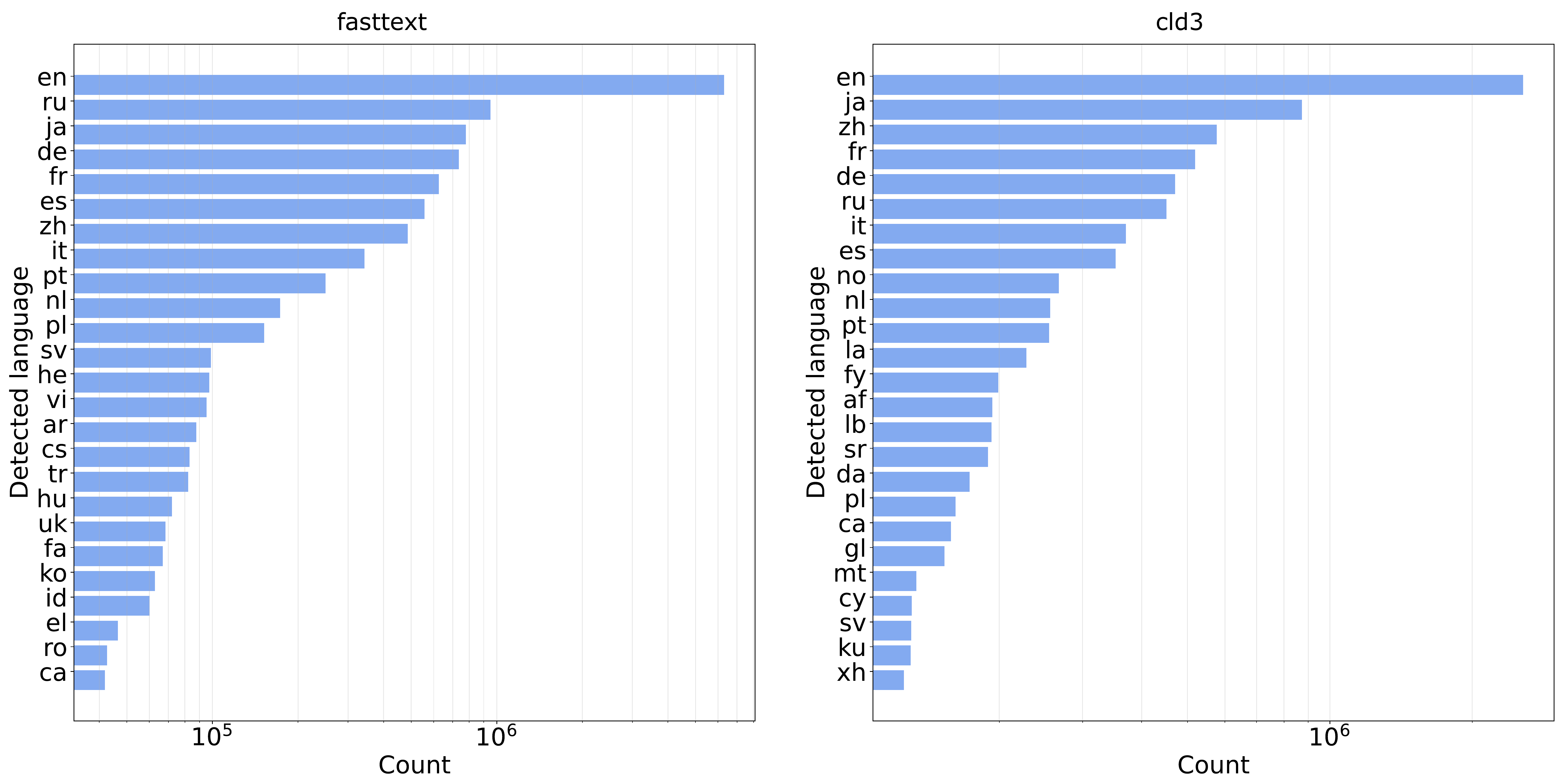}
    \caption{Counts for the top 25 most frequent languages in the {\small \texttt{small}} \pool, as predicted by fasttext \emph{(left)} and cld3 (\emph{right}).}
    \label{fig:langdet_language}
\end{figure*}

Analysing the textual component of each sample, we visualize frequency of the number of CLIP BPE tokens in the captions (Figure \ref{fig:tokens}) and most common languages (Figure \ref{fig:langdet_language}).
Token counts follow a long-tailed distribution with much more mass in the short sequence range, while English is the predominant language in \pool according to fasttext and cld3.

\begin{figure*}
    \centering
    \includegraphics[width=0.9\linewidth]{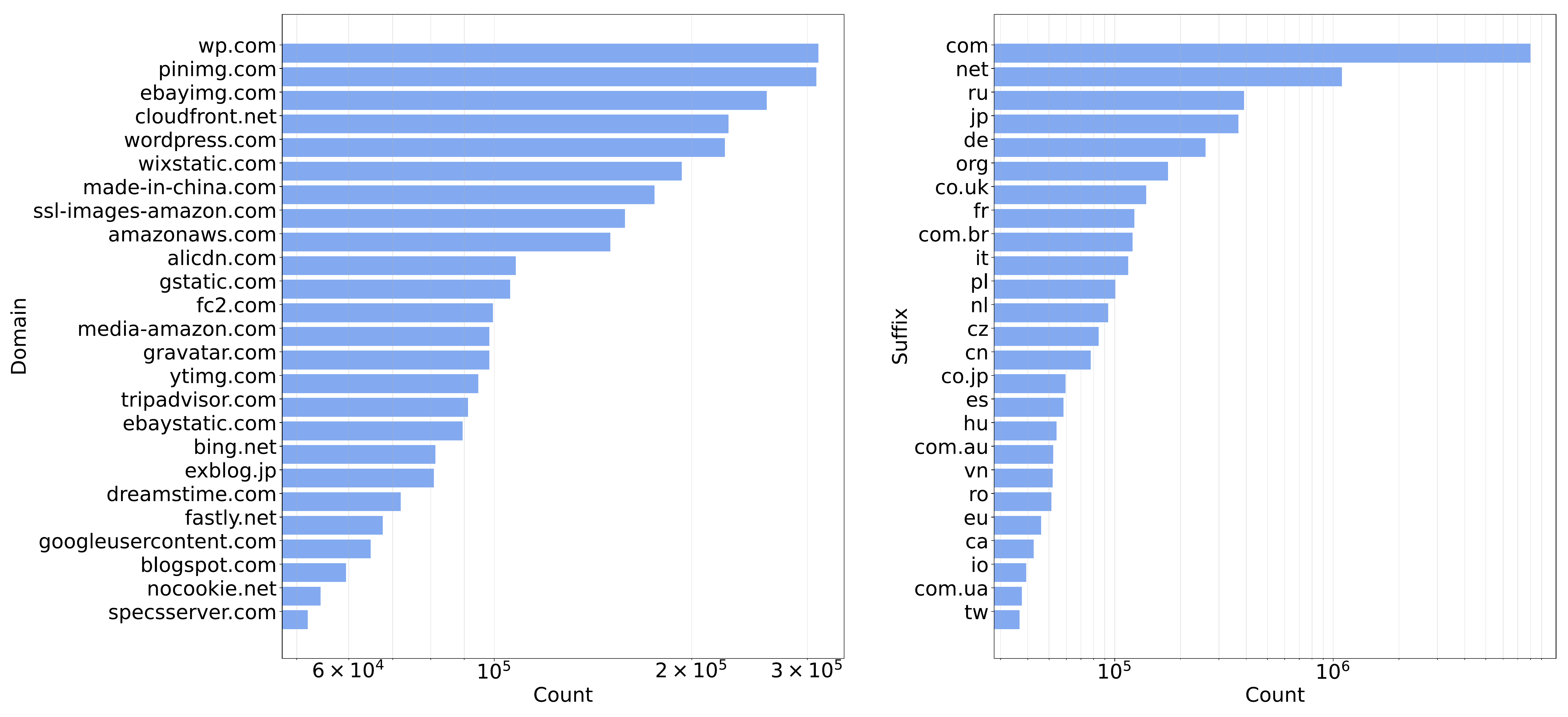}
    \caption{Counts for the top 25 most frequent domains \emph{(left)} and suffixes (\emph{right}) in the {\small \texttt{small}} \pool.}
    \label{fig:domain}
\end{figure*}

We also look at url statistics. In Figure \ref{fig:domain} we see common domain names in \pool (e.g., wordpress domains) and common suffixes (e.g., .com or .net).

\FloatBarrier

\section{Efficient training on data subsets}
\label{app:resharder}

When training at large scale, it is important to use efficient access patterns to load training data.
This typically means that data must be loaded using large sequential reads instead of random reads in order to maximize throughput.
In \datanet{}, this is facilitated by the WebDataset\footnote{\url{https://github.com/webdataset/webdataset}} format which stores the training examples in tar files (called ``shards'') and WebDataLoader which makes it easy to load data stored in this format.

Given an arbitrary subset of a pool, we would like to efficiently train on that subset.
Because WebDataset format does not permit efficient random access (a feature inherited from tar), we must read through the entire pool to select the required images.
There are two ways to implement this filtering:
\begin{enumerate}
    \item \textbf{Filter during training:} we apply a predicate during training data loading that discards data not present in the subset.\label{app:resharder:enum-train}
    \item  \textbf{Filter before training:} we iterate over the pool, selecting the images in the subset, and write them to a new WebDataset.\label{app:resharder:enum-reshard}
\end{enumerate}
After some profiling, we concluded that option~\ref{app:resharder:enum-train} had too much overhead in the case where the subset is much smaller than the pool. To see why, note that if the subset is an \(p\)-fraction of the pool size, then we would end up reading a \(1/p\) factor more data than needed for training.
Instead, we give an implementation of option~\ref{app:resharder:enum-reshard}, which performs at most twice as many reads as needed for training.\footnote{Since in \datanet{}, the number of examples seen is equal to the pool size.}

Our tool, called the \textit{resharder}, reads a set of uids in NumPy array format, scans through the pool, selecting those examples, and writes them to a new WebDataset.
The resharder uses multiprocessing to make good use of hardware and can be distributed over many computers to further increase throughput.
The resharder also supports streaming data to and from cloud storage such as Amazon S3.
The resharder is provided to participants as part of the competition tooling.

\section{Effect of duplicates in the training data}
\label{app:dedup_training}

Given that \pool was constructed by scraping the web for image and text pairs, there is a likelihood that some of our images are duplicates of each other, even if they originated from different web sources and have different captions. Here we examine the effect of removing such duplicates. We used the technique proposed by \citet{webster2023deduplication}, where CLIP image features are first compressed and then used to do an approximate nearest neighbor search. After this process, two images $x$ and $y$ are considered duplicates if $\frac{|d_{ADC}(x,x) - d_{ADC}(x,y)|}{d_{ADC}(x,x)} < T_{ADC}$, where $T_{ADC}$ is some threshold and $d_{ADC}(x,x)$ is the distance of a vector with its quantized version used for approximate nearest neighbor search. For each image, we search duplicates across its $1000$ nearest neighbors, and keep it if it's the one with the highest CLIP ViT-L/14 similarity score across its duplicates. Results can be seen in Table \ref{tab:app_dedup_training}, both when this technique is used by itself and in conjunction with ViT-B/32 filtering. We can see that results are similar to when only using CLIP filtering.

\begin{table*}
\caption{Effect of deduplication of training set for the medium size \pool. The filtering performed here is CLIP B32 score top 30\% (see Table \ref{tab:full-medium}). Higher threshold values lead to more samples being labeled as duplicates.}
\setlength\tabcolsep{4pt}
    \renewcommand{\arraystretch}{1.1}
        \rowcolors{2}{white}{light-light-gray}
\small
\centering
\begin{tabular}{lccc}
\toprule
Subset & Training dataset size & ImageNet accuracy & Average performance \\
\midrule
$T_{ADC} = 0.1$, without filtering & 99.8M & 0.195 & 0.275 \\
$T_{ADC} = 0.2$, without filtering & 85.9M & 0.200 & 0.277\\
$T_{ADC} = 0.5$, without filtering & 29.6M & 0.227 & 0.295\\
$T_{ADC} = 0.1$, with filtering & 33.5M & 0.288 & 0.337 \\
$T_{ADC} = 0.2$, with filtering & 30.6M & 0.289 & 0.337 \\
$T_{ADC} = 0.5$, with filtering & 15.5M & 0.252 & 0.311 \\
\bottomrule
\end{tabular}
\label{tab:app_dedup_training}
\end{table*}

\section{Hyperparameter ablations}
\label{app:hyperparam}

Recall that in \datanet, we freeze the training procedure and hyperparameters to focus the competition on dataset curation. However, this leads to the natural question: do ``better'' datasets (i.e., datasets that lead to higher accuracy models on zero-shot downstream tasks) remain consistent when training is modified. Hence we ablate key experimental choices: batch size, model architecture, and number of training steps.

\subsection{Batch size}
\label{app:batch_size}
We ablate over the batch size hyperparameter, doubling the batch size at the {\small \texttt{medium}} scale, but holding all other hyperparameters constant.
As see in Table~\ref{tab:bs}, we find that the delta rankings are largely consistent, for both ImageNet and Average performance, with rankings changing by at most plus or minus one position.
More specifically, rank correlation before and after doubling batch size is 0.96 for ImageNet and 0.98 for the Average over 38 datasets metric. 

\subsection{Model architecture}
\label{app:arch}
We choose to use the ViT architecture~\cite{dosovitskiy2021an} because of favorable CLIP scaling trends over vanilla ResNets~\cite{he2016deep} as reported by \citet{radford2021learning}. However, we still hope that better datasets for downstream ViT performance will lead to better datasets to train convolutional architectures. We look at the {\small \texttt{medium}} scale, swapping the ViT-B/32 architecture with a ConvNeXt model~\cite{liu2022convnet} with matched giga multiplier–accumulate operations (GMACs). Looking at Table~\ref{tab:conv}, we see that ranking of different filtering methods is again relatively consistent (i.e., 1.0 rank correlation for ImageNet and 0.87 rank correlation for the average metric). We conclude that improvements in dataset filtering have potential to improve more than just CLIP ViT model performance. 

\subsection{Number of training steps}
\label{app:steps}
\begin{figure*}
    \centering
    \includegraphics[width=.9\textwidth]{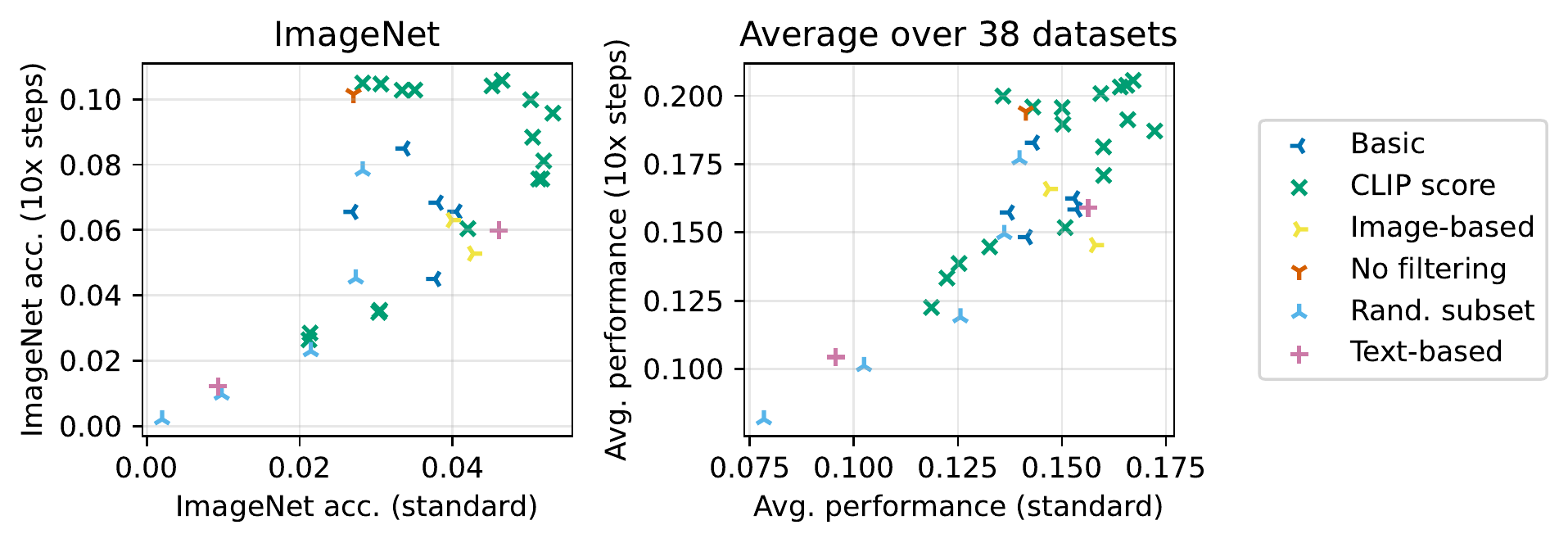}
    \caption{\emph{(left)} The effect of training for 10$\times$ steps for for {\small \texttt{small}} filtering track baselines on ImageNet. \emph{(right)} Similar plot but for Avg. performance. While the ordering of some methods changes quite drastically, we, in general, see a positive correlation.}
    \label{fig:steps}
\end{figure*}

Recall that one of our major design decisions for \datanet is to fix the hyperparameters associated with model training, following closely hyperparameters from prior work \cite{radford2021learning}. We choose to fix hyperparameters to place emphasis on data curation and remove confounders arising from hyperparameter differences between \users. Here we ablate our hyperparameter configuration by training {\small \texttt{small}} baselines for 10$\times$ more steps. In Figure \ref{fig:steps} we see positive correlation for ImageNet accuracy for the ablated and original hyperparameter configurations. We see similar correlation for average performance. See Table \ref{tab:app_steps} for specific values.

\begin{table*}
\rowcolors{3}{light-light-gray}{white}
\caption{Batch size ablation at the {\small{\texttt{medium}}} scale. We compare the standard \datanet{} {\small{\texttt{medium}}} configuration, with batch size 4096 against an ablated configuration with batch size 8192 ({\small{\texttt{medium: batch size 2x}}}). We find that the rankings of the baseline filtering strategies are relatively consistent. More precisely, the rank correlation is 0.96 on ImageNet and 0.98 for the Average over 38 datasets.}
\setlength\tabcolsep{5pt}
\renewcommand{\arraystretch}{1.1}
\small
\centering

\resizebox{\textwidth}{!}{
\begin{tabular}{wlccccccccc}
\toprule
 &    & Dataset & Samples & ImageNet & Average over & Delta ranking & Delta ranking\\
\multirow{-2}{*}{Scale} & \multirow{-2}{*}{Filtering strategy} & size & seen & & 38 datasets & ImageNet & Average\\\midrule

\cellcolor{white} & No filtering & 128M  & 128M & 0.176 & 0.258 & - & -\\
\cellcolor{white}& Basic filtering & 30M  & 128M & 0.226 & 0.285 & - & -\\
\cellcolor{white}& Text-based  & 31M  & 128M & 0.255 & 0.307 & - & -\\
\cellcolor{white} & Image-based  & 29M  & 128M & 0.268 & 0.312 & - & -\\
\cellcolor{white} & LAION-2B filtering & 13M  & 128M & 0.230 & 0.292 & - & -\\
\cellcolor{white} & CLIP score (L/14 30\%) & 38M  & 128M & 0.273 & \underline{0.328} & - & -\\
\cellcolor{white} \multirow{-7}{*}{{\small \texttt{medium}}} & Image-based $\cap$ CLIP score (L/14 30\%) & 14M  & 128M & \underline{0.297} & \underline{0.328} & - & -\\\midrule

\cellcolor{white} & No filtering & 128M  & 128M & 0.171 & 0.258 & 0 & 0\\
\cellcolor{white}& Basic filtering & 30M  & 128M & 0.219 & 0.277 & +1 (worse) & 0\\
\cellcolor{white}& Text-based  & 31M  & 128M & 0.251 & 0.299 & 0 & -1 (better) \\
\cellcolor{white} & Image-based  & 29M  & 128M & 0.260 & 0.299 & 0 & 0\\
\cellcolor{white} & LAION-2B filtering & 13M  & 128M & 0.215 & 0.288 & -1 (better) & 0\\
\cellcolor{white} & CLIP score (L/14 30\%) & 38M  & 128M & 0.271 & \underline{0.324} & 0 & 0\\
\cellcolor{white} \multirow{-7}{*}{{\small \texttt{medium: batch size 2x}}} & Image-based $\cap$ CLIP score (L/14 30\%) & 14M  & 128M & \underline{0.276} & 0.311 & 0 & +1 (worse)\\\midrule

\bottomrule
\end{tabular}}
\label{tab:bs}
\end{table*}

\begin{table*}
\rowcolors{3}{light-light-gray}{white}
\caption{Architure ablation at the {\small{\texttt{medium}}} scale. We compare the standard \datanet{} {\small{\texttt{medium}}} configuration, with a ViT-B/32 model against an ablated configuration ({\small{\texttt{medium: ConvNeXt}}}), which uses a ConvNeXt model with the same number of multiply-accumulate operations as the ViT. We find that the rankings of the baseline filtering strategies are relatively consistent. More precisely, the rank correlation is 1.0 on ImageNet and 0.87 for the Average over 38 datasets.}
\setlength\tabcolsep{5pt}
\renewcommand{\arraystretch}{1.1}
\small
\centering

\resizebox{\textwidth}{!}{
\begin{tabular}{wlccccccccc}
\toprule
 &    & Dataset & Samples & ImageNet & Average over & Delta ranking & Delta ranking\\
\multirow{-2}{*}{Scale} & \multirow{-2}{*}{Filtering strategy} & size & seen & & 38 datasets & ImageNet & Average\\\midrule

\cellcolor{white} & No filtering & 128M  & 128M & 0.176 & 0.254 & - & -\\
\cellcolor{white}& Basic filtering & 30M  & 128M & 0.226 & 0.280 & - & -\\
\cellcolor{white}& Text-based  & 31M  & 128M & 0.255 & 0.301 & - & -\\
\cellcolor{white} & Image-based  & 29M  & 128M & 0.268 & 0.307 & - & -\\
\cellcolor{white} & LAION-2B filtering & 13M  & 128M & 0.230 & 0.287 & - & -\\
\cellcolor{white} & CLIP score (L/14 30\%) & 38M  & 128M & 0.273 & \underline{0.323} & - & -\\
\cellcolor{white} \multirow{-7}{*}{{\small \texttt{medium}}} & Image-based $\cap$ CLIP score (L/14 30\%) & 14M  & 128M & \underline{0.297} & \underline{0.323} & - & -\\\midrule

\cellcolor{white} & No filtering & 128M  & 128M &  0.178 & 0.255 & 0 & 0\\
\cellcolor{white}& Basic filtering & 30M  & 128M & 0.232 & 0.272 & 0 & 0\\
\cellcolor{white}& Text-based  & 31M  & 128M & 0.255 & 0.298 & 0 & 0 \\
\cellcolor{white} & Image-based  & 29M  & 128M & 0.270 & 0.298 & 0 & +1 (better)\\
\cellcolor{white} & LAION-2B filtering & 13M  & 128M & 0.253 & 0.300 & 0 & -2 (better)\\
\cellcolor{white} & CLIP score (L/14 30\%) & 38M  & 128M & 0.279 & 0.326 & 0 & +1 (worse)\\
\cellcolor{white} \multirow{-7}{*}{{\small \texttt{medium: ConvNeXt}}} & Image-based $\cap$ CLIP score (L/14 30\%) & 14M  & 128M & \underline{0.323} & \underline{0.331} & 0 & 0\\\midrule

\bottomrule
\end{tabular}}
\label{tab:conv}
\end{table*}

\begin{table*}

\setlength\tabcolsep{4pt}
\renewcommand{\arraystretch}{1.1}
        \rowcolors{3}{light-light-gray}{white}
\small
\centering
\caption{Experiment details when extending the number of steps by 10 times the standard amount for that scale.}
\resizebox{\textwidth}{!}{
\begin{tabular}{llccccc}
\toprule

\multirow{2}{*}{Scale} & \multirow{2}{*}{Filtering} & \multirow{2}{*}{ImageNet} & ImageNet & \multirow{2}{*}{VTAB} & \multirow{2}{*}{Retrieval} & Average over \\
& &  & dist. shifts &  & & 38 datasets \\\midrule
\cellcolor{white} & No filtering	 & 0.102 & 0.093 & 0.204 & 0.147 & 0.196 \\
\cellcolor{white} & Random subset(75\%)	 & 0.078 & 0.072 & 0.182 & 0.129 & 0.178 \\
\cellcolor{white} & Random subset(50\%)	 & 0.045 & 0.049 & 0.161 & 0.104 & 0.150 \\
\cellcolor{white} & Random subset(25\%)	 & 0.023 & 0.029 & 0.134 & 0.075 & 0.119 \\
\cellcolor{white} & Random subset(10\%)	 & 0.010 & 0.018 & 0.119 & 0.069 & 0.101 \\
\cellcolor{white} & Random subset(1\%)	 & 0.002 & 0.006 & 0.097 & 0.056 & 0.082 \\
\cellcolor{white} & Caption length	 & 0.085 & 0.080 & 0.198 & 0.136 & 0.184 \\
\cellcolor{white} & Image size	 & 0.066 & 0.064 & 0.153 & 0.115 & 0.158  \\
\cellcolor{white} & English (fasttext)	 & 0.068 & 0.068 & 0.172 & 0.108 & 0.159 \\
\cellcolor{white} & English (fasttext) and caption length	 & 0.066 & 0.065 & 0.182 & 0.106 & 0.163 \\
\cellcolor{white} & English (fasttext), caption length, and image size	& 0.045 & 0.048 & 0.164 & 0.092  & 0.149 \\

\cellcolor{white} & CLIP B32 score top 10\%	 & 0.035 & 0.046 & 0.162 & 0.079 & 0.139 \\
\cellcolor{white} & CLIP B32 score top 20\%	 & 0.076 & 0.076 & 0.182 & 0.099 & 0.172 \\
\cellcolor{white} & CLIP B32 score top 30\%	 & 0.096 & 0.090 & 0.221 & 0.121 & 0.205 \\
\cellcolor{white} & CLIP B32 score top 40\%	 & 0.081 & 0.077 & 0.200 & 0.124 & 0.193 \\
\cellcolor{white} & CLIP B32 score top 50\%	 & 0.106 & 0.097 & 0.211 & 0.134 & 0.205 \\
\cellcolor{white} & CLIP B32 score top 75\%	 & 0.103 & 0.096 & 0.210 & 0.150 & 0.198 \\
\cellcolor{white} & CLIP B32 score top 90\%	 & 0.105 & 0.096 & 0.212 & 0.152 & 0.202  \\
\cellcolor{white} & CLIP B32 threshold at 0.3 + English filter	 & 0.029 & 0.036 & 0.152 & 0.078 & 0.134  \\
\cellcolor{white} & CLIP B32 threshold at 0.28 + English filter	 & 0.035 & 0.041 & 0.168 & 0.086 & 0.145 \\
\cellcolor{white} & CLIP B32 threshold at 0.3	 & 0.076 & 0.078 & 0.199 & 0.102 & 0.182 \\

\cellcolor{white} & CLIP L14 score top 10\%	 & 0.026 & 0.037 & 0.130 & 0.073 & 0.123 \\
\cellcolor{white} & CLIP L14 score top 20\%	 & 0.060 & 0.064 & 0.161 & 0.096 & 0.153 \\
\cellcolor{white} & CLIP L14 score top 30\%	 & 0.088 & 0.087 & 0.199 & 0.115 & 0.188  \\
\cellcolor{white} & CLIP L14 score top 40\%	 & 0.100 & 0.096 & 0.217 & 0.122 & 0.207 \\
\cellcolor{white} & CLIP L14 score top 50\%	 & 0.104 & 0.098 & 0.212 & 0.136 & 0.203 \\
\cellcolor{white} & CLIP L14 score top 75\%	 & 0.103 & 0.095 & 0.189 & 0.146 & 0.191 \\
\cellcolor{white} & CLIP L14 score top 90\%	 & 0.105 & 0.095 & 0.203 & 0.145 & 0.198 \\

\cellcolor{white} & Image-based clustering (ImageNet1k)	 & 0.053 & 0.053 & 0.162 & 0.091 & 0.146 \\
\cellcolor{white} & Image-based clustering (ImageNet21k)	 & 0.063 & 0.059 & 0.173 & 0.108 & 0.167 \\
\cellcolor{white} & Text-based clustering (ImageNet1k)	 & 0.012 & 0.018 & 0.120 & 0.062 & 0.104 \\
\cellcolor{white} & Text-based clustering (ImageNet21k)	 & 0.262 & 0.216 & 0.305 & 0.246 & 0.300 \\

\cellcolor{white} & Intersect IN1k image clustering and CLIP B32 score top 30\%	 & 0.058 & 0.059 & 0.179 & 0.098 & 0.161  \\
\cellcolor{white} & Intersect IN1k image clustering and CLIP L14 score top 30\%	 & 0.049 & 0.051 & 0.171 & 0.090 & 0.150 \\
\cellcolor{white} & Intersect IN21k image clustering and CLIP B32 score top 30\%	 & 0.071 & 0.070 & 0.192 & 0.107 & 0.175 \\
\cellcolor{white} \multirow{-36}{*}{{\small \texttt{small}}}& Intersect IN21k image clustering and CLIP L14 score top 30\%	 & 0.064 & 0.065 & 0.200 & 0.096 & 0.173 \\
\midrule

\cellcolor{white} & No filtering	 & 0.370 & 0.304 & 0.387 & 0.355 & 0.383 \\
\cellcolor{white} & English (fasttext), caption length, and image size	 & 0.317 & 0.269 & 0.324 & 0.271 & 0.334 \\
\cellcolor{white} & CLIP B32 score top 30\%	 & 0.436 & 0.351 & 0.433 & 0.345 & 0.430 \\
\cellcolor{white} & CLIP B32 score top 40\%	 & 0.434 & 0.353 & 0.448 & 0.365 & 0.442  \\
\cellcolor{white} & CLIP B32 score top 50\%	 & 0.426 & 0.352 & 0.439 & 0.377 & 0.433 \\
\cellcolor{white} & CLIP B32 score top 75\%	 & 0.398 & 0.325 & 0.396 & 0.374 & 0.411 \\
\cellcolor{white} & Image-based clustering (ImageNet1k)	 & 0.363 & 0.294 & 0.347 & 0.279 & 0.347 \\
\cellcolor{white} & Image-based clustering (ImageNet21k)	 & 0.374 & 0.303 & 0.372 & 0.318 & 0.372 \\
\cellcolor{white} & Intersect IN1k image clustering and CLIP B32 score top 30\%	 & 0.415 & 0.330 & 0.413 & 0.310 & 0.403 \\
\cellcolor{white} \multirow{-10}{*}{{\small \texttt{medium}}}& Intersect IN1k image clustering and CLIP L14 score top 30\%	 & 0.405 & 0.325 & 0.399 & 0.295 & 0.387 \\
\bottomrule
\end{tabular}}
\label{tab:app_steps}
\vspace{3pt}
\end{table*}

\FloatBarrier

\section{Detector-based baselines}
\label{app:detector}
While controlling for factors such as class balance is common in the supervised settings, experimenting with analogous strategies in the context of multimodal datasets and CLIP training is a pertinent direction. Towards this end, we use the Detic detector~\cite{zhou2022detecting} to annotate the {\small\texttt{medium}
} pool (128M samples) by extracting bounding boxes and class labels for the 1203 LVIS~\cite{gupta2019lvis} objects categories. Following the original Detic paper, we retain predictions whose confidence score exceeds 0.5. Based on these annotations, we construct the following five strategies:
\begin{itemize}
    \item \emph{Object exists}: Subset for which there exists at least one detection from the 1203 LVIS categories.
    \item \emph{Object centered}: Subset for which there exists at least one detection from the 1203 LVIS categories with a bounding box center falling in the center grid cell of a 3x3 grid superimposed on the image.
    \item Balancing by class: We define 1204 buckets—1203 buckets corresponding to the LVIS classes and an additional bucket for images that do not have any detections. For each image in the medium pool, we assign the image to the bucket(s) corresponding to the detected classes. We then construct a dataset such that there are an equal number of samples from each bucket and the total number of samples specified by a particular scale (e.g., 128M samples for medium scale). Note, for rare classes there can be many repeated samples and for common classes only a subset of the total samples will be in the dataset.
    \item Balancing by position: We define 26 buckets—0, 1, …, 24 corresponding to 5x5 grid locations in an image. An image is added to bucket(s) when it contains a bounding box whose center falls in the bucket’s grid cell. The 25th bucket contains images for which there are no detections. We again construct a dataset such that there are an equal number of samples from each bucket.
    \item Balancing by count: We define 12 buckets—0, 1, …, 10 corresponding to zero to ten detections in an image and a twelfth bucket corresponding to images with more than ten detections. We yet again construct a dataset such that there are an equal number of samples from each bucket.
\end{itemize}

We employ each of these strategies on the {\small\texttt{medium}
} scale. Since the above strategies can be composed with any starting pool, we additionally apply each of the above Detic-based strategies to our previous best {\small\texttt{medium}
} scale filtered pool: Image-based $\cap$ CLIP score (L/14 30\%). This yields five more datasets for 10 baselines in total.

\begin{table*}
\rowcolors{3}{light-light-gray}{white}
\caption{Detector-baseed baselines at the {\small{\texttt{medium}}} scale. We start with No filtering and Image-based $cap$ CLIP score (L/14 30\%) pools and apply five additional filtering and balancing strategies described in Appendix~\ref{app:detector}. We find that even with these more sophisticated strategies, the No filtering and Image-based $cap$ CLIP score (L/14 30\%) still performs best at medium scale. Properly balancing multimodal data remains an open direction for future work.}
\setlength\tabcolsep{5pt}
\renewcommand{\arraystretch}{1.1}
\small
\centering

\resizebox{0.8\textwidth}{!}{
\begin{tabular}{wlcccccc}
\toprule
 &    & Samples & ImageNet & Average over\\
\multirow{-2}{*}{Scale} & \multirow{-2}{*}{Filtering strategy} & seen & & 38 datasets\\\midrule

\cellcolor{white} & No filtering & 128M & 0.176 & 0.258\\
\cellcolor{white}& \quad $\cap$ Object exists  & 128M & 0.181 & 0.263 \\
\cellcolor{white}& \quad $\cap$ Object centered & 128M & 0.187 & 0.263\\
\cellcolor{white} & \quad $\cap$ Balance by class& 128M & 0.038 & 0.141\\
\cellcolor{white} & \quad $\cap$ Balance by position & 128M &  0.040 & 0.148\\
\cellcolor{white} \multirow{-7}{*}{{\small \texttt{medium}}} & \quad $\cap$ Balance by object count   & 128M & 0.127 & 0.221\\\midrule
\cellcolor{white} & Image-based $\cap$ CLIP score (L/14 30\%) & 128M & \underline{0.297} & \underline{0.328} \\
\cellcolor{white}& \quad $\cap$ Object exists  & 128M & 0.289 & 0.319 \\
\cellcolor{white}& \quad $\cap$ Object centered & 128M & 0.247 & 0.286\\
\cellcolor{white} & \quad $\cap$ Balance by class& 128M & 0.034 & 0.136\\
\cellcolor{white} & \quad $\cap$ Balance by position & 128M & 0.036 & 0.136\\
\cellcolor{white} \multirow{-7}{*}{{\small \texttt{medium}}} & \quad $\cap$ Balance by object count & 128M & 0.068 & 0.169\\

\bottomrule
\end{tabular}}
\label{tab:detector}
\end{table*}

Our results are summarized in the Table \ref{tab:detector}. In summary: 1) The Image-based $\cap$ CLIP score (L/14 30\%) baseline still performs best. 2) Balancing data in the context of multimodal CLIP training remains an open problem. All balancing strategies lead to divergence of the CLIP contrastive loss and result in poor model performance. We hypothesize that this is due to the long-tailed nature of the data distribution, which leads to many repeated samples in our balanced data construction. This in turn, increases the likelihood that samples are contrasted with themselves in the loss computation.

\section{Training details}
\label{app:train}

The full set of hyperparameters used for each scale is shown in Table \ref{tab:train-hparams}.
For choosing hyperparameters, we follow the OpenCLIP library~\cite{ilharco2021openclip}, an open source reproduction of OpenAI's CLIP. For the \texttt{small}, \texttt{medium}, and \texttt{large} tracks, these hyperparameters are equal to those in the CLIP paper, except with reduced batch size so that training runs on reasonable hardware. For the \texttt{xlarge} track, batch size is increased from that in OpenAI's CLIP to accelerate training by allowing the use of many GPUs simultaneously with high utilization. For this run we also double the learning rate following prior work~\cite{cherti2022reproducible}.

\begin{table*}
\caption{Experimental configuration for each scale, including the size of the pool we provide, the model architecture and hyperparameters.}
\setlength\tabcolsep{4.5pt}
    \renewcommand{\arraystretch}{1.1}
        \rowcolors{2}{white}{light-light-gray}
\small
\centering
\resizebox{\textwidth}{!}{
\begin{tabular}{lcccccccc}
\toprule
Scale   & Model    & Train compute (MACs) & Pool size &  \# samples seen & Learning rate & AdamW $\beta_2$ & Warmup & Batch size \\\midrule
{\small \texttt{small}}    & ViT-B/32 & $9.5\times 10^{16}$ & 12.8M & 12.8M & 5e-4 & 0.98 &  500 & 4096 \\
{\small \texttt{medium}}   & ViT-B/32 & $9.5\times 10^{17}$ & 128M  & 128M  & 5e-4 & 0.98 & 500 & 4096 \\
{\small \texttt{large}}	& ViT-B/16 & $2.6\times 10^{19}$ & 1.28B & 1.28B & 5e-4 & 0.98 & 500 & 8192 \\
{\small \texttt{xlarge}}	& ViT-L/14 & $1.1\times 10^{21}$ & 12.8B  & 12.8B & 1e-3 & 0.95 & 10k & 90112 \\
\bottomrule
\end{tabular}
}
\label{tab:train-hparams}
\end{table*}

\section{Evaluation details}
\label{sec:app-eval}

\begin{figure*}
    \centering
    \includegraphics[width=.7\textwidth]{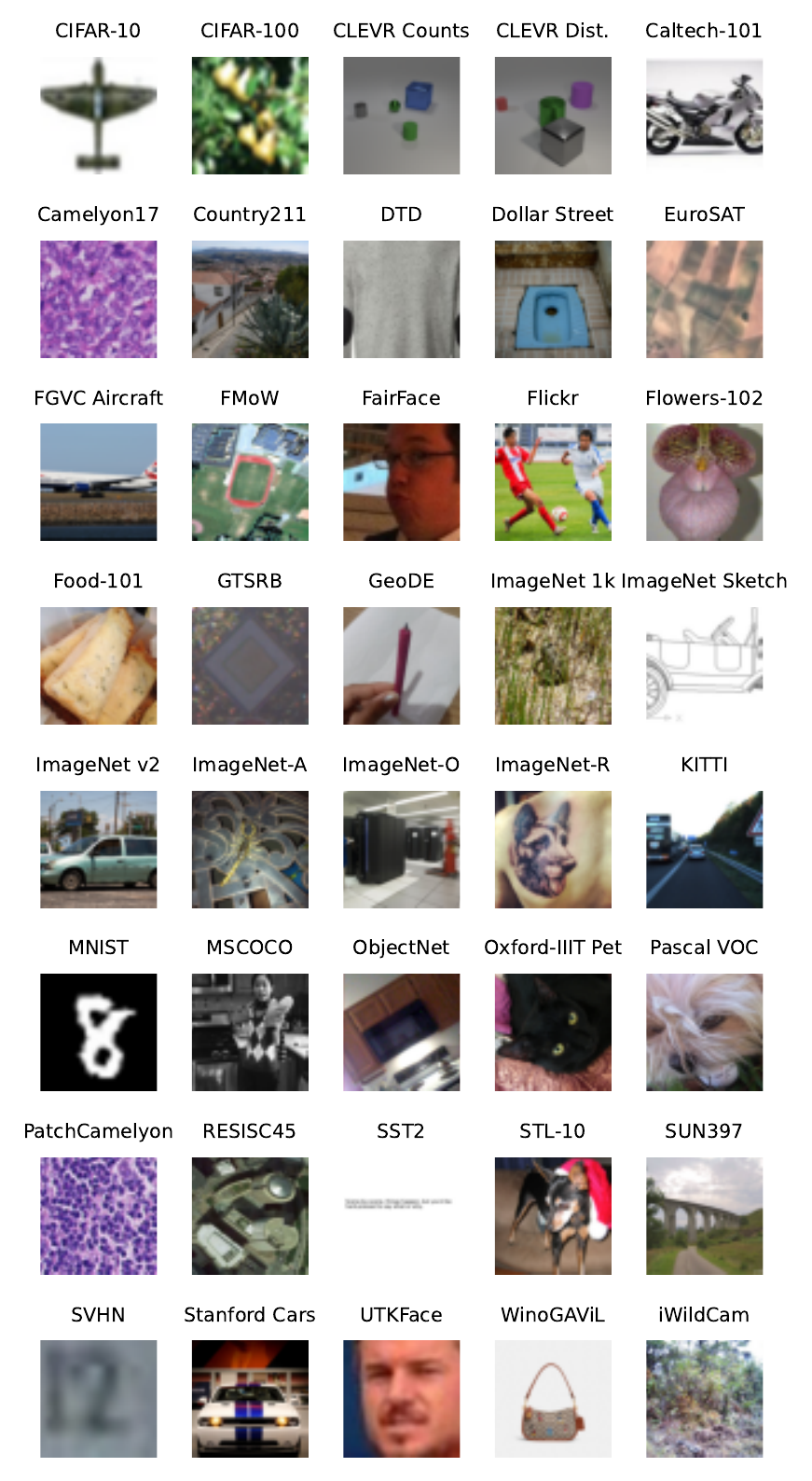}
    \caption{Randomly sampled images from the evaluation datasets we consider.}
    \label{fig:downstream-samples}
\end{figure*}

Models are evaluated over a wide range of 38 tasks to measure proficiency in various domains. We include 22 of the 27 classification tasks in the test suite of \citet{radford2021learning}, excluding the few datasets that have license restrictions, are in video format, or are no longer available in their original form. We include 6 datasets that were designed to test generalization of models trained on ImageNet. We also include a majority of the Visual Task Adaptation Benchmark, excluding 3 datasets that are ill-suited for zero-shot evaluation \cite{vtab}. We include 3 datasets from the WILDS benchmark, which tests robustness to distribution shifts and spurious correlations \cite{wilds2021,sagawa2022extending}. Finally, we include 2 additional datasets, Dollar Street and GeoDE, which test robustness of classification performance across income levels and geographical regions \cite{rojas2022dollar,ramaswamy2022geode}. Furthermore, we evaluate zero-shot image and text retrieval on the Flickr30k and MSCOCO datasets, and image association on the WinoGAViL dataset \cite{flickr30k,mscoco,bitton2022winogavil}.
The complete list of evaluation tasks is given in Table~\ref{tab:eval-sets}. We show a sample from each dataset in Figure \ref{fig:downstream-samples}.

\begin{table*}
\caption{Evaluation tasks.}
\setlength\tabcolsep{4.5pt}
    \renewcommand{\arraystretch}{1.1}
        \rowcolors{2}{white}{light-light-gray}
\resizebox{\textwidth}{!}{
\centering
\begin{tabular}{lllrrrc}
\toprule
Task type & Dataset & Task & Test set size & Number of classes & Main metric & Clean\\ \midrule
\cellcolor{white}  & Caltech-101 \cite{caltech101} & Object recognition & 6,085 & 102 & mean per class & \checkmark\\
\cellcolor{white} & CIFAR-10 \cite{cifar10andcifar100} & Visual recognition & 10,000 & 10  & accuracy & \checkmark\\
\cellcolor{white} & CIFAR-100 \cite{cifar10andcifar100} & Visual recognition & 10,000 & 100 & accuracy & \checkmark\\
\cellcolor{white} & CLEVR Counts \cite{clevr,vtab} & Counting & 15,000 & 8 & accuracy & \\
\cellcolor{white} & CLEVR Distance \cite{clevr,vtab} & Distance prediction & 15,000 & 6 & accuracy & \\
\cellcolor{white} & Country211 \cite{radford2021learning,yfcc100m} & Geolocation & 21,100 & 211 & accuracy & \checkmark\\
\cellcolor{white} & DTD \cite{dtd} & Texture classification & 1,880 & 47 & accuracy & \checkmark\\
\cellcolor{white} & EuroSAT \cite{eurosat,vtab} & Satellite imagery recognition & 5,400 & 10 & accuracy & \checkmark\\
\cellcolor{white} & FGVC Aircraft \cite{fgvc_aicraft} & Aircraft recognition & 3,333 & 100 & mean per class & \checkmark\\
\cellcolor{white} & Food-101 \cite{food101} & Food recognition & 25,250 & 101 & accuracy & \checkmark\\
\cellcolor{white} & GTSRB \cite{gtsrb} & Traffic sign recognition & 12,630 & 43 & accuracy & \checkmark\\
\cellcolor{white} & ImageNet 1k \cite{deng2009imagenet} & Visual recognition & 50,000 & 1,000 & accuracy & \checkmark\\
\cellcolor{white} & ImageNet Sketch \cite{imagenetsketch} & Visual recognition & 50,889 & 1,000  & accuracy & \checkmark\\
\cellcolor{white} & ImageNet V2 \cite{imagenetv2} & Visual recognition & 10,000 & 1,000  & accuracy & \checkmark\\
\cellcolor{white} & ImageNet-A \cite{imageneta_and_imageneto} & Visual recognition & 7,500 & 200 & accuracy  & \checkmark\\
\cellcolor{white} & ImageNet-O \cite{imageneta_and_imageneto} & Visual recognition & 2,000 & 200  & accuracy & \checkmark\\
\cellcolor{white} & ImageNet-R \cite{imagenetr} & Visual recognition & 30,000 & 200  & accuracy & \checkmark\\
\cellcolor{white} & KITTI distance \cite{kitti,vtab} & Distance prediction & 711 & 4  & accuracy\\
\cellcolor{white} & MNIST \cite{lecun1998mnist} & Digit recognition & 10,000 & 10 & accuracy  & \checkmark\\
\cellcolor{white} & ObjectNet \cite{objectnet} & Visual recognition & 18,574 & 113  & accuracy & \checkmark\\
\cellcolor{white} & Oxford Flowers-102 \cite{flowers102} & Flower recognition & 6,149 & 102 & mean per class & \checkmark\\
\cellcolor{white} & Oxford-IIIT Pet \cite{pets,vtab} & Pet classification & 3,669 & 37 & mean per class & \checkmark\\
\cellcolor{white} & Pascal VOC 2007 \cite{pascal-voc-2007} & Object recognition & 14,976 & 20 & accuracy & \checkmark\\
\cellcolor{white} & PatchCamelyon \cite{patchcamelyon,vtab} & Metastatic tissue cls. & 32,768 & 2  & accuracy\\
\cellcolor{white} & Rendered SST2 \cite{vtab} & Sentiment classification & 1,821 & 2  & accuracy & \checkmark\\
\cellcolor{white} & RESISC45 \cite{resisc45,vtab} & Satellite imagery recognition & 6,300 & 45 & accuracy  & \checkmark\\
\cellcolor{white} & Stanford Cars \cite{cars} & Vehicle recognition & 8,041 & 196 & accuracy  & \checkmark\\
\cellcolor{white} & STL-10 \cite{stl10} & Visual recognition & 8,000 & 10  & accuracy & \checkmark\\
\cellcolor{white} & SUN-397 \cite{sun397} & Scene recognition & 108,754 & 397 & accuracy & \checkmark \\
\cellcolor{white} & SVHN \cite{svhn,vtab} & Digit recognition & 26032 & 10 & accuracy & \checkmark \\
\cellcolor{white} & iWildCam \cite{beery2020iwildcam,wilds2021} & Animal recognition & 42,791 & 182 & macro F1 score & \checkmark \\
\cellcolor{white} & Camelyon17 \cite{bandi2018detection,wilds2021} & Metastatic tissue cls. & 85,054 & 2 & accuracy \\
\cellcolor{white} & FMoW \cite{christie2018functional,wilds2021} & Satellite imagery recognition & 22,108 & 62 & worst-region acc.  & \checkmark\\
\cellcolor{white} & Dollar Street \cite{rojas2022dollar} & Object recognition & 3,503 & 58 & worst-income top-5 acc.  & \checkmark\\
\cellcolor{white} \multirow{-35}{*}{Classification}& GeoDE \cite{ramaswamy2022geode} & Object recognition & 12,488 & 40 & worst-region acc. & \checkmark \\
\midrule
 \cellcolor{white} & Flickr30k \cite{flickr30k} & Image and text retrieval & 31,014 & N/A & R@1 & \checkmark \\
\cellcolor{white} & MSCOCO \cite{mscoco} & Image and text retrieval & 5,000 & N/A & R@1  & \checkmark\\
\cellcolor{white} \multirow{-3}{*}{Retrieval}& WinoGAViL \cite{bitton2022winogavil} & Commonsense association & 3,563 & N/A & Jaccard score & \checkmark \\
\bottomrule
\end{tabular}} 
\label{tab:eval-sets}
\end{table*}

\paragraph{Prompt choice.} Since we perform zero-shot evaluation, prompt and class name selection is important, and can have a significant impact on the results. To avoid heavy prompt engineering and overtuning to individual models, we opt to use the prompt templates used in \citet{radford2021learning} whenever possible. Most datasets come with pre-defined class names, but some are overwritten with more descriptive labels, again based on previous literature. For datasets with no precedent in zero-shot evaluation, we reuse prompt templates from other datasets with a similar domain and task (e.g., SVHN is evaluated with MNIST prompts and class names).

\paragraph{Evaluation metrics.} For the majority of classification tasks, the primary evaluation metric is accuracy. For certain datasets with class imbalances, we instead compute mean per-class accuracy, as done in \citet{radford2021learning}. On the WILDS benchmark datasets, we use the primary metric specified for each dataset on their leaderboard. Dollar Street and GeoDE test model generalization across socioeconomic and geographic diversity. Thus, for Dollar Street, we compute worst-group top-5 accuracy, with groups defined by income level, emulating \citet{rojas2022dollar}; for GeoDE, we compute worst-group accuracy, with groups defined by region (Africa, Americas, West Asia, East Asia, Southeast Asia, and Europe), as defined in \citet{ramaswamy2022geode}. For the image-text retrieval tasks, Flickr and MSCOCO, we compute both image and text recall (fraction of text captions for which the correct image was selected and vice versa), and plot their arithmetic mean. On WinoGAViL, we compute the Jaccard score (intersection-over-union) for each example, and show results for the harder samples (10 and 12 candidates). More information on WinoGAViL evaluation can be found in \citet{bitton2022winogavil}.

\paragraph{Clean subset.} For five of our evaluation tasks (the two CLEVR tasks, the two Camelyon tasks, and KITTI) the zero-shot performance of all evaluated models appears to be close to that of random guessing, and lack correlation to the type of filtering method used (see \Cref{fig:imagenet-vs-all-breakdown}). Consequently, we studied performance averaged only on the remaining 33 tasks, but found not substantial qualitative differences in our results.  As a result, we opted to report the average on the full evaluation suite throughout our study. 

\begin{figure*}
    \centering
    \includegraphics[width=.9\textwidth]{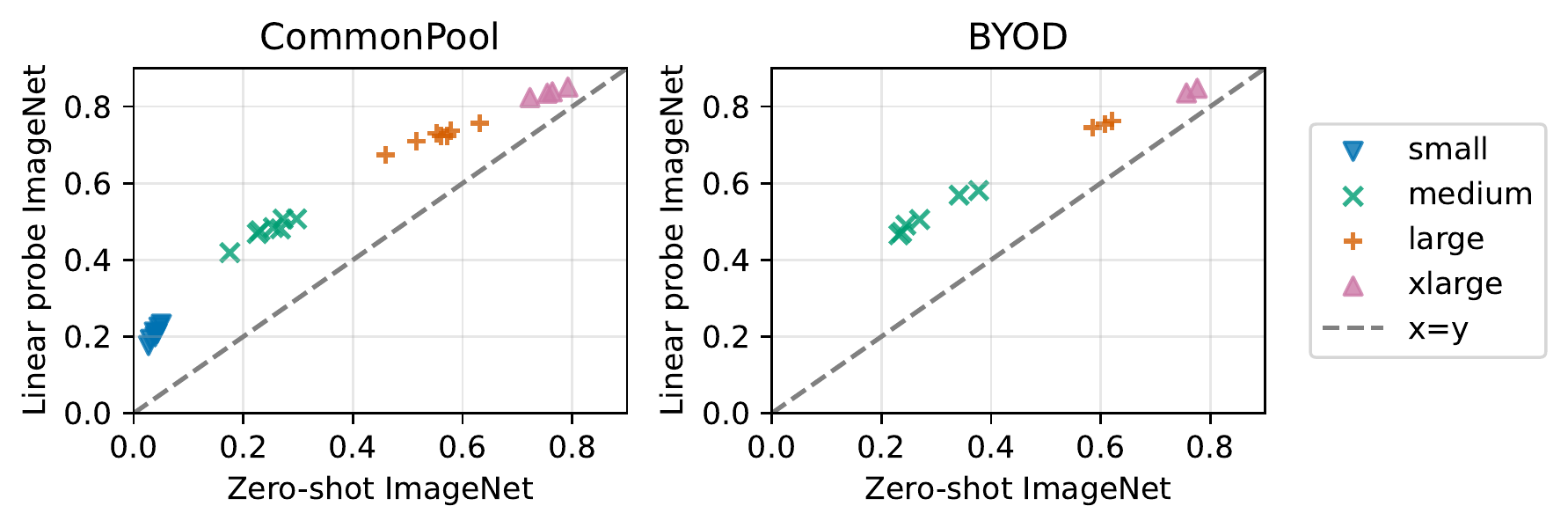}
    \caption{Zero-shot ImageNet and Linear probe ImageNet performance for models from Tables \ref{tab:main} and \ref{tab:byod}. Relative ordering of models demonstrates high rank correlations of 0.99 and 1.0 for \pool and \byod respectively.}
    \label{fig:linear-probes}
\end{figure*}

\paragraph{Zero-shot vs. fine-tuning protocols.}
One critical decision in \datanet is how exactly to evaluate models and whether or not to fine-tune models on evaluation tasks (i.e., supervised fine-tuning directly on task training sets). We opt for zero-shot evaluation, where a models are applied to downstream tasks directly to 1) ease computational burden on \users and 2) measure the out-of-the-box generalization capabilities of our models. To validate this design decision, we conduct linear probes on all models presented in Tables \ref{tab:main} and \ref{tab:byod} on ImageNet. We follow a standard probing protocol and fine-tune the last linear layer from zero-shot initialization for 40 epochs with learning rate 1e-3, batch size 256, AdamW optimizer with default settings with the exception of weight decay (that we set to zero), and a cosine annealing schedule. As seen in Figure \ref{fig:linear-probes}, zero-shot and linear probe performance follow similar trends for both filtering and \byod tracks. Moreover the Spearman rank correlation between the two protocols over the models considered is 0.99 for the filtering track and 1.0 for \byod. This suggests that better zero-shot models on ImageNet are correlated with better representations of linear probe fine-tuning on ImageNet.

\subsection{Visual Question Answering}

In addition to our evaluation suite containing multiple classification and retrieval tasks, we conducted experiments on visual question answering. More specifically, following \citet{shen2021much}, we use the CLIP models to contrast images with prompts formed by the questions and each candidate answer, without fine-tuning (i.e., in a zero-shot setting). Using the VQA v1 dataset [2], for each candidate answer, we construct a text prompt that also includes the question following the template {\small \texttt{Question: [question text] Answer: [answer text]}, as in \citet{ilharco2022patching}. This text is then fed to CLIP’s text encoder. As previously noted by multiple authors, CLIP models struggle on this task, potentially due to the mismatch between the text in the downstream task and the captions seen during pre-training \citet{shen2021much,ilharco2022patching,song2022clip}. Nonetheless, we observe a strong correlation between VQA performance and ImageNet accuracy (0.877) and between VQA performance and average performance on our full evaluation suite. Full results are shown in Figure \ref{fig:vqa}.

\begin{figure}
    \centering
    \includegraphics[width=0.9\linewidth]{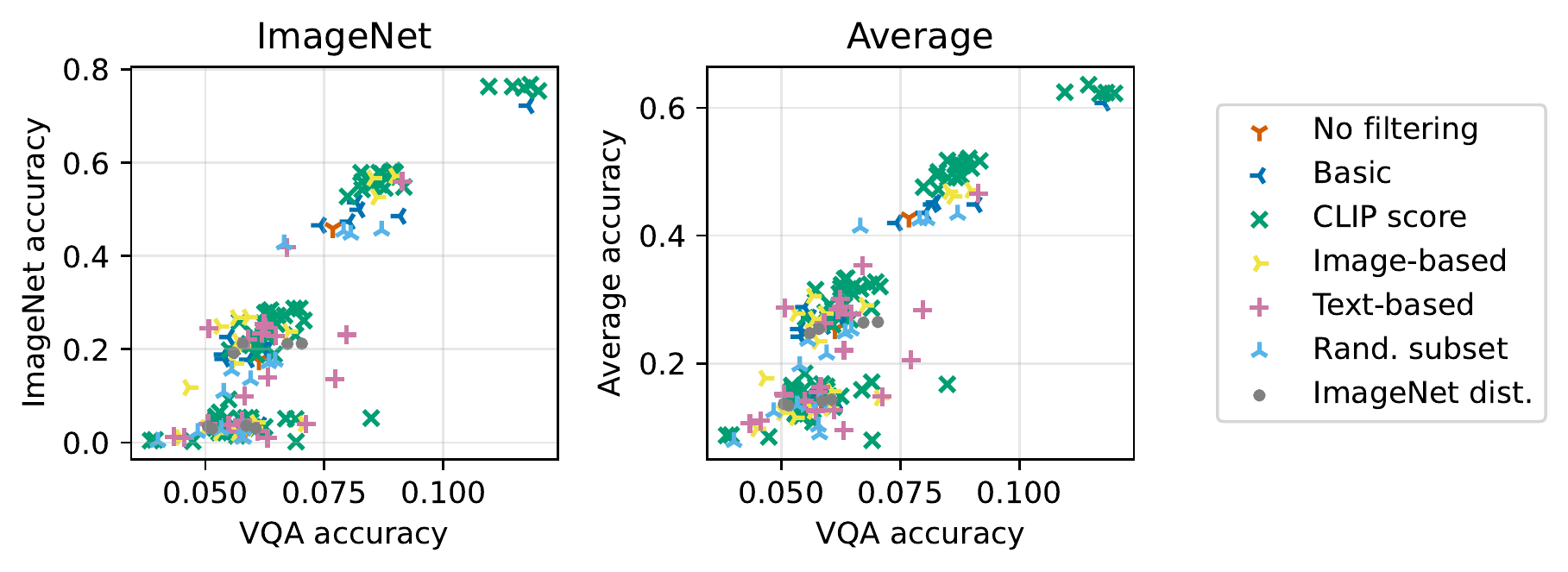}
    \caption{Correlation between zero-shot performance on the VQA v1 dataset and results on ImageNet and our full evaluation suite.}
    \label{fig:vqa}
\end{figure}

\section{Baseline details}
\label{sec:app-baselines}
Here we provide additional details on the creation of our baseline subsets.
To highlight the qualitative differences between the filtering strategies we also provide visualization for \emph{No filtering} (Figure \ref{fig:no_filter}), \emph{Basic filtering} (Figure \ref{fig:basic_filter}), and \emph{CLIP score (L/14 30\%)} (Figure \ref{fig:clip_filter}), which can all be found in Table~\ref{tab:main}. Notice that No filtering gives relatively noisy data (e.g., matching a bicycle with a caption: ``IMG\_2187.jpg''), while CLIP score samples give qualitatively more descriptive cations.

\begin{figure*}
    \centering
    \includegraphics[width=0.9\linewidth]{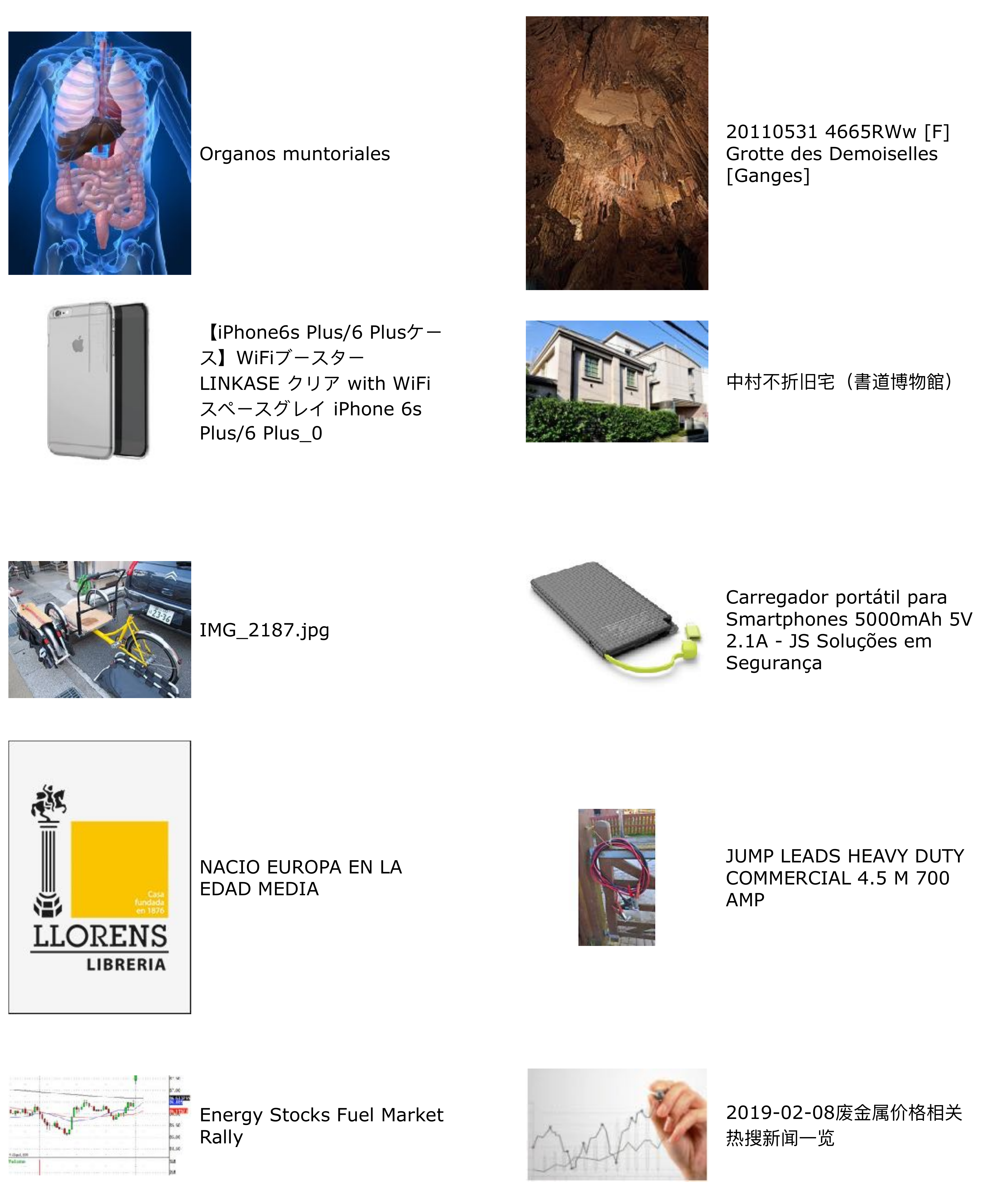}
    \caption{An i.i.d. sample from {\small \texttt{small}} \pool generated after applying the \emph{No filter} strategy. Hence, these samples represent random images from \pool.}
    \label{fig:no_filter}
\end{figure*}

\begin{figure*}
    \centering
    \includegraphics[width=0.9\linewidth]{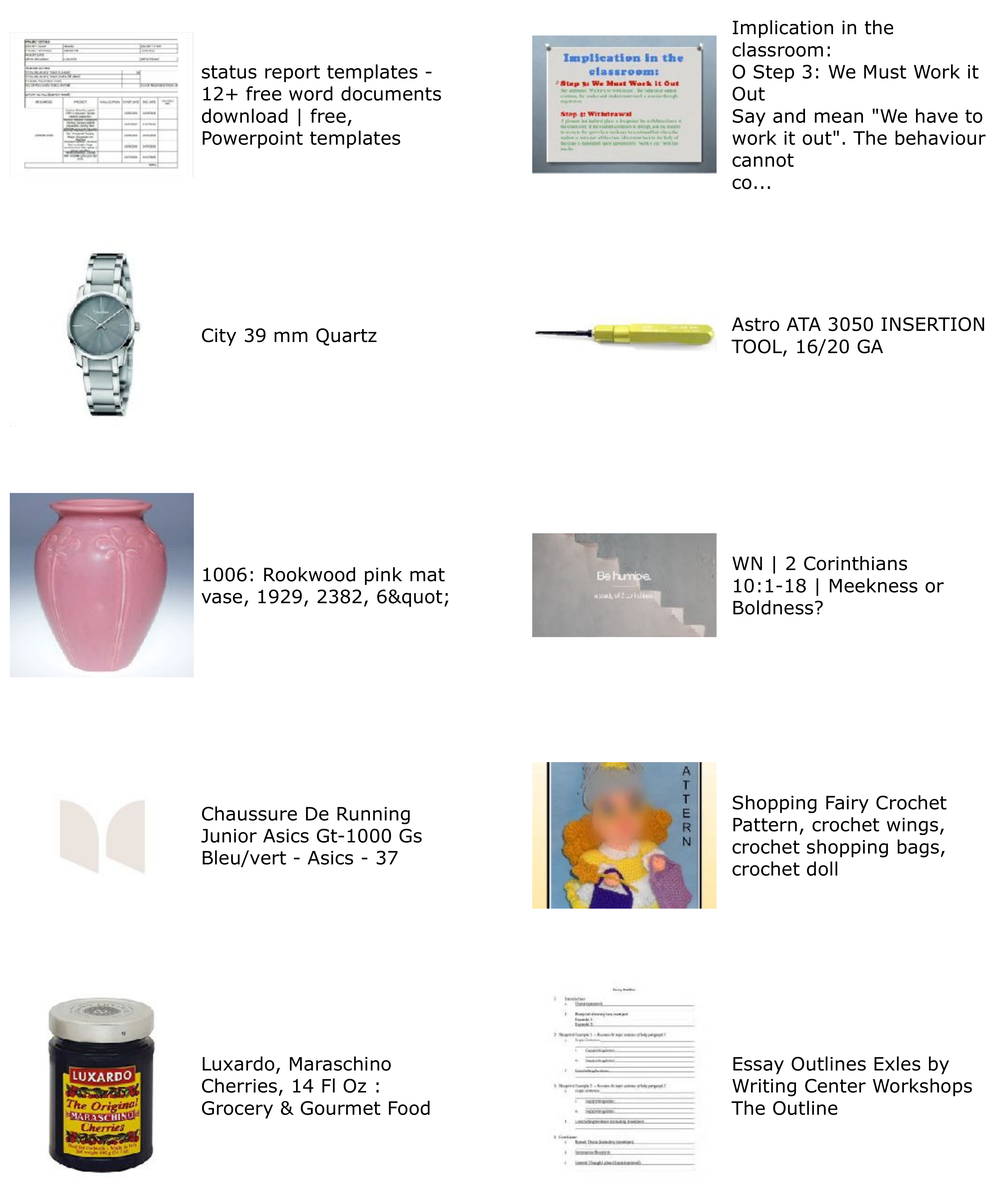}
    \caption{An i.i.d. sample from {\small \texttt{small}} \pool generated after applying the \emph{Basic filter} strategy.}
    \label{fig:basic_filter}
\end{figure*}

\begin{figure*}
    \centering
    \includegraphics[width=0.9\linewidth]{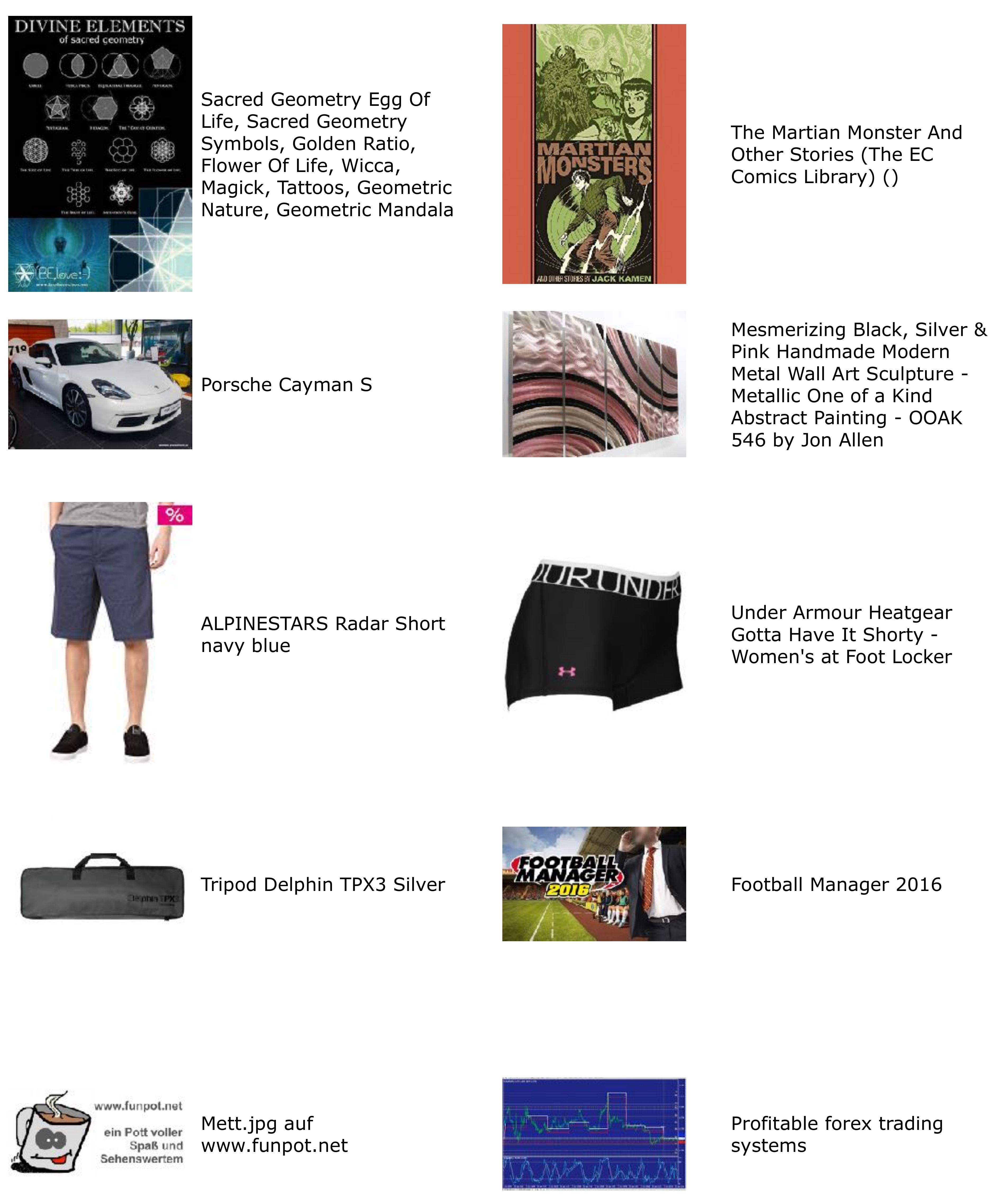}
    \caption{An i.i.d. sample from {\small \texttt{small}} \pool generated after applying the CLIP score (L/14 30\%)} strategy.
    \label{fig:clip_filter}
\end{figure*}

\FloatBarrier

\subsection{Filtering track}\label{sec:app-baselines-pool}

\paragraph{Basic filtering.} 

For language detection, we use Fasttext 0.92, version lid.176, and cld3 - library gcld3 3.0.13.
We count the number of words in each caption by splitting using whitespaces.

\paragraph{CLIP thresholds.}
We use OpenAI pretrained CLIP ViT-B/32 and ViT-L/14 models~\citep{radford2021learning} to compute the cosine similarity text and image tower outputs as the CLIP scores. On the {\small \texttt{small}} and {\small \texttt{medium}} pools, we also experiment with baselines that filter out samples in the top few percentiles of CLIP scores. Specifically, we try baselines that use samples with top \{1,2,5\}-30\% CLIP scores (ViT-B/32 model), and the performance is sightly better on the {\small \texttt{small}} pool (at most 0.5 gain of averaged accuracy) while slightly worse on the {\small \texttt{medium}} pool (0.4-0.8 loss of averaged accuracy). In Table \ref{tab:filtering_thresholds}, we show how the CLIP score thresholds relate to the fraction of the pool retained by the filter.

\begin{table}
    \renewcommand{\arraystretch}{1.1}
        \rowcolors{2}{white}{light-light-gray}
\caption{CLIP threshold filtering configurations. ``Fraction'' denotes the size of the filtered subset relative to the pool.}
\setlength\tabcolsep{2pt}
\footnotesize
\centering
\begin{tabular}{lccc}
\toprule
CLIP model   & En. filtering   & Threshold & Fraction \\
\midrule
ViT-B/32 & \xmark & 0.384 & 1\% \\
ViT-B/32 & \xmark & 0.358 & 3\% \\
ViT-B/32 & \cmark & 0.300 & 10.2\% \\
ViT-B/32 & \xmark & 0.325 & 10\% \\
ViT-B/32 & \cmark & 0.28 & 7.4\% \\
ViT-B/32 & \xmark & 0.300 & 20\% \\
ViT-B/32 & \xmark & 0.281 & 30\% \\
ViT-B/32 & \xmark & 0.263 & 40\% \\
ViT-B/32 & \xmark & 0.247 & 50\% \\
ViT-B/32 & \xmark & 0.215 & 75\% \\
ViT-B/32 & \xmark & 0.193 & 90\% \\
\midrule
ViT-L/14 & \xmark & 0.364 & 1\% \\
ViT-L/14 & \xmark & 0.334 & 3\% \\
ViT-L/14 & \cmark & 0.300 & 5.4\% \\
ViT-L/14 & \xmark & 0.295 & 10\% \\
ViT-L/14 & \cmark & 0.280 & 3.3\% \\
ViT-L/14 & \xmark & 0.266 & 20\% \\
ViT-L/14 & \xmark & 0.243 & 30\% \\
ViT-L/14 & \xmark & 0.222 & 40\% \\
ViT-L/14 & \xmark & 0.203 & 50\% \\
ViT-L/14 & \xmark & 0.160 & 75\% \\
ViT-L/14 & \xmark & 0.129 & 90\% \\
\bottomrule
\end{tabular}
\label{tab:filtering_thresholds}
\end{table}

\paragraph{Text-based filtering.} 
Each synset is represented by a synset offset that can be used to retrieve the synset from WordNet. In order to verify if a caption has a word corresponding to a synset from our set we iterate over every word and retrieve the synsets that this word can describe (using nltk.corpus WordNet). Following that, we retrieve the most likely lemma representing that synset, find its synset offset, and check if the number is part of the IN21K or IN1K sets.\footnote{For the ImageNet 21K synsets, we have used the list in \url{https://storage.googleapis.com/bit_models/imagenet21k_wordnet_ids.txt}} 

\paragraph{Text-based sampling.} 
This baseline uses text only to filter labels which mention concepts (synsets) appearing in IN21K, and applies a temperature parameter to control how equally-represented different concepts are in the dataset.
For synset $j$, let $N_j$ be the number of examples containing words matched to that synset, where as before for each word we only match the most likely synset. Furthermore, for image-text pair $i$ let $T_i$ be the set of synset matched to the caption. 

The probability of sampling example $i$ is proportional to either 
$
\frac{1}{|T_i|} \sum_{j \in T_{i}} N_{j}^{\alpha-1}
$ (average synset score in the data point)
or 
$
\max_{j \in T_{i}} N_{j}^{\alpha-1}
$ (maximum synset score in the data point), where $\alpha$ is a ``temperature'' parameter controlling the flatness of the distribution. We sample examples with replacement but discard any example repeated more than 100 times.

\paragraph{Image-based filtering.}
We now provide a detailed description of the Image-based filtering procedure. First, since the core of the procedure concerns only image content, we begin with basic text-bsaed filtering: we remove from the pool only all examples with non-English captions (as determined by fasttext), and all examples whose captions have less than two words or less than six characters. 

Next, we use clustering of image embeddings to select a subset of examples whose image content is related to a clean training set of interest.
Let $e_1, \ldots, e_M$ denote the CLIP image embeddings of the remaining examples in the pool. We cluster these embeddings into $K=10^5$ clusters using Faiss with 20 iterations, and let $c_1, \ldots, c_K$ denote the resulting cluster centers. Due to memory constraints, for the \texttt{large}  and \texttt{xlarge} pools,
we perform the clustering on a random subset of about 160M examples (that pass the basic text-based filtering). For an embedding vector $v$, let
\[
I(v) = \arg\max_{i\le K} 
{\langle v , c_i \rangle}
\]
 denote the index of the cluster center nearest to $v$ as measured by inner product.
 Let $f_1, \ldots, f_N$ denote the CLIP image embeddings of a clean supervised training set (we experiment with either ImageNet 1K or ImageNet 21K), and let
\[
\mathcal{S} = \{I(f_i) \mid 1 \le i \le N\}
\]
be the set of cluster indices who are nearest neighbors to some clean training set image. We then keep only images in the pool whose nearest cluster center is in $\mathcal{S}$. That is, out of the $M$ examples passing the text-based filtering, the output subset keeps the examples with indices
\[
\{1\le j\le M\mid I(e_j)\in\mathcal{S}\}.
\]

\paragraph{Image-based sampling.}
In addition to filtering methods, we experiment with cluster-based sampling methods. First, we compute the score of $i$-th cluster $s_i$ as the number of ImageNet data assigned to this cluster. Then, for parameter $\alpha>0$ we define a distribution over the pool by sampling cluster $i$ with probability $\frac{s_i^\alpha}{\sum_{j} s_j^\alpha}$ and uniformly sampling an example for the cluster, rejecting any example repeated more than 100 times. We try 5 different $\alpha$, i.e., $\{0, 0.2, 0.5, 1.0, 2.0\}$, and the best average accuracy is obtained when $\alpha=0.2$, while the performance is still worse than the image-based filtering on the \texttt{small} and \texttt{medium} pool. We therefore do not include this line of baselines in the experiments of \texttt{large} pool.

\paragraph{ImageNet distance filtering.}
We rank the samples in the pool by the minimum embedding distance (1 minus cosine similarity) between its image and the ImageNet images; both embeddings are obtained from OpenAI pretrained CLIP ViT-L/14 model~\citep{radford2021learning}. Then we select top images by different fractions as in image-based filtering methods.

\subsection{\byod track}
\label{app:byod}

We experiment with the following data sources:

\begin{itemize}[leftmargin=6pt,topsep=0pt,itemsep=0pt,parsep=4pt]
\item CC12M \cite{changpinyo2021conceptual}: images and HTML alt-text crawled and filtered from web pages.
\item YFCC15M: this is the 15M subset of the YFCC100M dataset \cite{yfcc100m} that \citet{radford2021learning} used for dataset ablation in their CLIP paper.
\item RedCaps \cite{desai2021redcaps}: 12M images and corresponding captions were crawled from 350 manually curated subreddits between 2008 and 2020.
\item Shutterstock: 106M images and captions were obtained from the Shutterstock website in 2021 \cite{nguyen2022quality}. We use the ``photos'' subset of this dataset, with 58M samples, which we found performed best, unless specified otherwise.
\item WIT \cite{srinivasan2021wit}:  Image-text pairs from Wikipedia pages. We use the attribution fields as captions, which we found performed best.
\item COYO \cite{coyo700m}: A collection of 700M image-text pairs from Common Crawl.
\item LAION-2B \cite{laion5b}: A 2.32 billion english subset of LAION-5B.
\item LAION-COCO: A dataset with 600M images from LAION-5B and synthetic captions.\footnote{\url{https://laion.ai/blog/laion-coco/}}
\item LAION-A: According to \href{https://laion.ai/}{laion.ai}, LAION-A is a 900M subset of LAION-2B \cite{laion5b} with the aesthetic filtering procedure used in LAION-aesthetic\footnote{\url{https://github.com/LAION-AI/laion-datasets/blob/main/laion-aesthetic.md}} and pHash deduplication \cite{idealods2019imagededup}.
\end{itemize}

\begin{table*}
\caption{Measuring the quality of external data sources}
    \renewcommand{\arraystretch}{1.1}
        \rowcolors{3}{light-light-gray}{white}
\setlength\tabcolsep{4pt}
\small
\centering
\resizebox{\textwidth}{!}{

\begin{tabular}{lccccc}
\toprule
\multirow{2}{*}{Dataset} & \multirow{2}{*}{Dataset size} & \multirow{2}{*}{ImageNet acc.} & Avg. accuracy & \multirow{2}{*}{Avg. cos. sim. (B/32)} & \multirow{2}{*}{Avg. cos. sim. (L/14)} \\
  &  &  & ImageNet and OOD sets & &  \\
\midrule
CC12M & 10M & 27.8 & 34.0 & 0.306 & 0.268 \\
YFCC15M & 15M & 22.6 & 24.6 & 0.262 & 0.198 \\
RedCaps & 11M & 26.8 & 31.5 & 0.281 & 0.240 \\
Shutterstock & 15M & 21.0 & 28.3 & 0.314 & 0.273 \\
\bottomrule
\end{tabular}
}
\label{tab:app_byod_quality}
\end{table*}

In Table \ref{tab:app_byod_quality}, we use some heuristics to measure the quality of some external data sources. First, following \citet{nguyen2022quality}, we train a CLIP model on a 5M random subset from each source, and evaluate the performance of the resulting models on ImageNet and ImageNet-derived distributions --- ImageNet-V2 \cite{imagenetv2}, ImageNet-R \cite{imagenetr}, ImageNet-Sketch \cite{imagenetsketch} and ObjectNet \cite{objectnet}. Moreover, for each data source, we use OpenAI's pretrained CLIP ViT-B/32 and ViT-L/14 models to compute the cosine similarity between image and text embeddings of a data point, and obtain the average cosine similarity score for the whole dataset.

\begin{table*}

\setlength\tabcolsep{4pt}
\renewcommand{\arraystretch}{1.1}
        \rowcolors{3}{light-light-gray}{white}
\small
\centering
\caption{Zero-shot performance for select baselines in the \byod track. Unless specified otherwise, \pool means our pool filtered with CLIP score (L/14, 30\%). }
\resizebox{\textwidth}{!}{
\begin{tabular}{lclccccc}
\toprule
\multirow{2}{*}{Scale} & \multirow{2}{*}{Data source}   & Training & \multirow{2}{*}{ImageNet} & ImageNet & \multirow{2}{*}{VTAB} & \multirow{2}{*}{Retrieval} & Average over \\
& & dataset size &  & dist. shifts &  & & 38 datasets \\\midrule

 \cellcolor{white} & \#0 & CC12M & 0.099 & 0.080 & 0.223 & 0.197 & 0.205 \\
 \cellcolor{white} & \#1 & LAION15M & 0.083 & 0.076 & 0.210 & 0.144 & 0.189 \\
 \cellcolor{white} & \#2 & RedCaps & 0.076 & 0.066 & 0.177 & 0.141 & 0.168 \\
 \cellcolor{white} & \#3 & Shutterstock 15M & 0.083 & 0.070 & 0.214 & 0.159 & 0.185 \\
 \cellcolor{white} & \#4 & YFCC15M & 0.071 & 0.046 & 0.182 & 0.147 & 0.164 \\
\cellcolor{white}  & \#5 & \#0 + \#1 + \#2 & 0.097 & 0.084 & 0.208 & 0.161 & 0.195 \\
\cellcolor{white}  & \#6 & \#0 + \#1 + \#3 & 0.091	 & 0.081 & 0.222 & 0.138 & 0.202\\
\cellcolor{white}  & \#7 & \#0 + \#2 + \#3 + \#4 & 0.095 & 0.075 & 0.205 & 0.164 & 0.186 \\
\cellcolor{white}  \multirow{-9}{*}{{\small \texttt{small}}}& \#8 & \#0--4 & 0.093	 & 0.076 & 0.205 & 0.162 & 0.193 \\
\midrule

\cellcolor{white}  & \#9 & CC12M & 0.245 & 0.189 & 0.283 & 0.289 & 0.272\\
\cellcolor{white}  & \#10 & LAION15M & 0.270 & 0.215 & 0.317 & 0.255 & 0.306 \\
\cellcolor{white}  & \#11 & RedCaps & 0.237 & 0.166 & 0.271 & 0.178 & 0.263 \\
\cellcolor{white}  & \#12 & Shutterstock 15M & 0.229 & 0.191 & 0.316 & 0.260 & 0.290 \\
\cellcolor{white}  & \#13 & YFCC15M & 0.232 & 0.137 & 0.263 & 0.245 & 0.257 \\
\cellcolor{white}  & \#14 & \#9 + \#10 + \#11 & 0.376 & 0.287 & 0.387 & 0.323 & 0.366 \\
\cellcolor{white}  & \#15 & \#9 + \#10 + \#12 & 0.342 & 0.278 & 0.362 & 0.345 & 0.357 \\
\cellcolor{white}  & \#16 & \#9 + \#11 + \#12 + \#13 & 0.360 & 0.268 & 0.365 & 0.275 & 0.345 \\
\cellcolor{white}  & \#17 & \#9--13 & 0.371 & 0.285 & 0.408 & 0.280 & 0.367 \\
\cellcolor{white}  & \#18 & Shutterstock illustration & 0.053 & 0.094 & 0.205 & 0.125 & 0.180 \\
\cellcolor{white}  & \#19 & Shutterstock photo & 0.342 & 0.209 & 0.364 & 0.350 & 0.331 \\
\cellcolor{white}  & \#20 & Shutterstock vectors & 0.072 & 0.151 & 0.216 & 0.148 & 0.208 \\
\cellcolor{white}  & \#21 & Shutterstock full & 0.313 & 0.254 & 0.353 & 0.331 & 0.342 \\
\cellcolor{white}  & \#22 & WIT full & 0.096 & 0.063 & 0.196 & 0.104 & 0.177 \\
\cellcolor{white}  & \#23 & WIT English & 0.051 & 0.038 & 0.145 & 0.083 & 0.143 \\
\cellcolor{white}  & \#24 & COYO & 0.272 & 0.235 & 0.301 & 0.254 & 0.304 \\
\cellcolor{white} \multirow{-17}{*}{{\small \texttt{medium}}}& \#25 & LAION-COCO & 0.209 & 0.205 & 0.293 & 0.359 & 0.297 \\\midrule

\cellcolor{white}  & \#26 &  Shutterstock illustration & 0.337 & 0.203 & 0.307 & 0.322 & 0.306 \\
\cellcolor{white}  & \#27 & Shutterstock photo & 0.485 & 0.304 & 0.432 & 0.427 & 0.398 \\
\cellcolor{white}  & \#28 & Shutterstock vectors & 0.126 & 0.223 & 0.244 & 0.191 & 0.246 \\
\cellcolor{white} & \#29 & Shutterstock full & 0.500	 & 0.412 & 0.472 & 0.451 & 0.456 \\
\cellcolor{white} & \#30 & COYO & 0.547 & 0.456 & 0.475 & 0.549 & 0.486 \\
\cellcolor{white} & \#31 & LAION-COCO & 0.355 & 0.351 & 0.395 & 0.494 & 0.398 \\
\cellcolor{white} & \#32 & COYO + LAION-COCO & 0.528 & 0.458 & 0.479 & 0.589 & 0.498 \\
\cellcolor{white} & \#33 & LAION-A & 0.611 & 	0.474 & 0.501 & 0.542 & 0.505 \\
\cellcolor{white} & \#34 & LAION-2B & 0.585	& 0.472	& 0.504	& 0.525	& 0.515 \\
\cellcolor{white} & \#35 & \pool +  \#9--13 & 0.602 & 0.498 & 0.541 & 0.416 & 0.537 \\
\cellcolor{white} & \#36 & \pool +  \#9--13 (2x upsampled) & 0.613 & 0.507 & 0.559 & 0.433 & 0.543 \\
\cellcolor{white} & \#37 & \pool +  \#9--13 (4x upsampled) & 0.615 & 0.514 & 0.553 & 0.427  & 0.543 \\
\cellcolor{white} & \#38 & \pool +  \#9--13 (6x upsampled) & 0.620 & 0.519 & 0.558 & 0.437 & 0.549 \\
\cellcolor{white} & \#39 & \pool +  \#9--13 (8x upsampled) & 0.624 & 0.520 & 0.533& 0.443 & 0.537 \\
\cellcolor{white} & \#40 & \pool +  \#9--13 (10x upsampled) & 0.621 & 0.520 & 0.540 & 0.441 & 0.537 \\
\cellcolor{white} & \#41 & \pool +  COYO & 0.561 & 0.472 & 0.504 & 0.508 & 0.513 \\
\cellcolor{white} & \#42 & \pool + LAION-A & 0.607 & 0.480 & 0.531 & 0.514 & 0.527 \\
\cellcolor{white} & \#43 & \pool + LAION-COCO & 0.522 & 0.457 & 0.513 & 0.498 & 0.514 \\
\cellcolor{white} & \#44 & \pool +  \#9+\#11+\#13+\#19 & 0.609 & 0.508 & 0.546 & 0.439 & 0.536 \\
\cellcolor{white} & \#45 & \pool +  \#9+\#11+\#13+\#19 (2x upsampled) & 0.621 & 0.509 & 	0.547 & 0.458 & 0.541 \\
\cellcolor{white} & \#46 & \pool + \#9+\#11+\#13+\#19 (4x upsampled) & 0.632 & 0.515 & 0.533 & 0.452 & 0.532 \\
\cellcolor{white} & \#47 & \pool +  \#9+\#11+\#13+\#19 (6x upsampled) & 0.635 & 0.515 & 0.535 & 0.471 & 0.532 \\
\cellcolor{white} & \#48 & \pool +  \#9+\#11+\#13+\#19 (8x upsampled) & 0.633 & 0.515 & 0.523 & 0.464 & 0.530 \\
\cellcolor{white} \multirow{-24}{*}{{\small \texttt{large}}}& \#49 & \pool +  \#9+\#11+\#13+\#19 (10x upsampled) & 0.630 & 0.513 & 0.523 & 0.356 & 0.521 \\\midrule
\cellcolor{white} & \#50 & LAION-2B & 0.757 & 0.631 & 0.611 & 0.619 & 0.621 \\
\cellcolor{white}  & \#51 & \pool +  \#9+\#11+\#13+\#19 & 0.766	& 0.660 & 0.662 & 0.539 & 0.659 \\
\cellcolor{white} & \#52 & \pool +  \#9+\#11+\#13+\#19 (6x upsampled) & 0.776 & 0.671 & 0.633 & 0.552 & 0.649 \\
\cellcolor{white} \multirow{-4}{*}{{\small \texttt{xlarge}}}& \#53 & \pool +  \#9+\#11+\#13+\#19 (18x upsampled) & 0.771 & 0.667 & 0.629 & 0.554 & 0.643 \\
\bottomrule
\end{tabular}}
\label{tab:byod}
\vspace{3pt}

\end{table*}

\subsubsection{Additional results}

We present a series of results for the \byod track in Table \ref{tab:byod}.

\section{Fairness and biases}
\label{app:fairness}

To study the biases displayed by our models, we include two diversity-related datasets, Dollar Street \cite{rojas2022dollar} and GeoDE \cite{ramaswamy2022geode}, in our evaluation suite, and perform further analysis on the face datasets FairFace \cite{karkkainen2021fairface} and UTKFace \cite{utkface} with demographic labels, following \citet{radford2021learning}.

\subsection{Diversity}

\begin{figure*}
    \centering
    \includegraphics[width=0.7\linewidth]{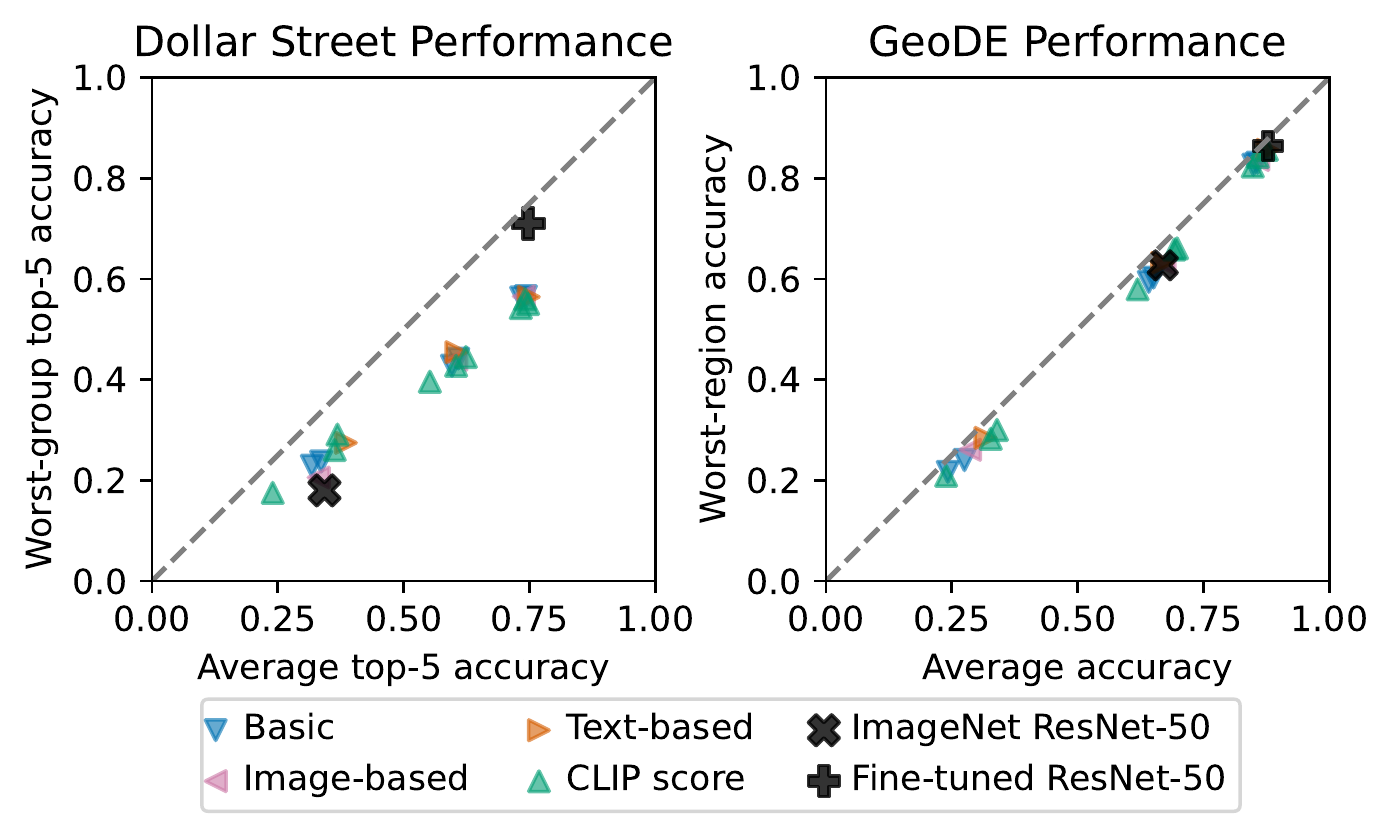}
    \caption{Comparison of average and worst-group scores for Dollar Street and GeoDE diversity datasets. On Dollar Street, our overall higher-performing models display a larger worst-group performance gap (corresponding to lower income households). GeoDE does not show this trend.}
    \label{fig:robustness_diversity}
\end{figure*}

We break down model performance on the Dollar Street and GeoDE datasets in Figure~\ref{fig:robustness_diversity}. Dollar Street consists of images of household items taken in homes around the world, and represents a wide socioeconomic range that includes homes with no Internet access \cite{rojas2022dollar}. The objects belong to ImageNet categories, and the task is image classification. Standard ImageNet-trained models achieve monotonically increasing performance levels with higher household income levels \cite{rojas2022dollar}. Here we use the income-based subgroups defined in \citet{rojas2022dollar}, and find a similar bias as discovered in their paper. While our trained models show a smaller worst-group performance gap than an ImageNet-trained ResNet-50, they underperform a model fine-tuned on Dollar Street. Models with higher average accuracy show a larger worst-group gap, which future work should try to address.

GeoDE consists of images of everyday items and objects, which again fall into ImageNet categories. The dataset represents six world regions equally, and primarily aims to promote geographic diversity of datasets \cite{ramaswamy2022geode}. Both ImageNet models and our models show less bias under this distribution compared to Dollar Street, with a smaller worst-group accuracy gap. The trends show that performance across all regions improves steadily with increased scale, and the performance approaches that of a model fine-tuned on GeoDE. While we know that classifiers trained specifically on ImageNet can display geographic biases \cite{ramaswamy2022geode}, these biases are not apparent in our GeoDE model evaluations. Future work is needed to investigate the extent to which our models have geographic biases not evaluated in GeoDE.

\subsection{Fairness}

Emulating \citet{radford2021learning}, we evaluate our best models from the filtering and \byod tracks on the human face datasets FairFace and UTKFace, using zero-shot classification to predict the race, gender, and age annotated in these datasets. Following \citet{Hanna2019TowardsAC} and \citet{hundt2022robots}, we acknowledge that these evaluations can be problematic as race and gender should not be considered fixed categories, but rather fluid attributes that may change for individuals, based on they way they identify at any given moment---regardless of appearance. We include these evaluations for continuity with prior work and as a probe into model behaviour, but hope future work will consider improved face fairness evaluation. We also note that race, gender, and age classification are not the intended end-goals of the models or benchmark, and we do not condone the use of \pool or models trained on \pool data for \emph{any} decisions involving people.

As described in Appendix~\ref{app:face}, our filleting track models are trained on images with faces blurred. Nevertheless, these models still perform significantly above random chance on face classification. We hypothesize that this is due to a combination of faces bypassing our face blurring filter in the training data, contextual clues outside of the face region, or signal associated with skin color. The BYOD track model performs even better than the filtering track model. We hypothesize that this is because BYOD data is used off-the-shelf and hence contains non-blurred faces.
In Table~\ref{tab:app_fairness_overall}, we present overall accuracy for these three traits. Note that race is treated as a binary variable (white or non-white) to enable comparison to prior results, gender is a binary variable (male or female) according to annotations, and age is binned into 9 ranges according to the annotation precision of FairFace. The \byod model, performs better at distinguishing the annotated gender, but is worse at distinguishing annotated race and age.

We further break down these statistics over the intersection of race and gender, examining gender classification accuracies in Table~\ref{tab:app_fairness_intersection}. We find that there are drastic differences in accuracy across different annotated subgroups, varying by both race and gender. The filtering models shows a tendency to misclassify Black, Southeast Asian, and East Asian males as females at 20.7\%, 17\%, and 19.3\% respectively on FairFace. Furthermore, we find that while the \byod model improves accuracy, on FairFace most of this improvement is on men (ranging from 1.7pp gain to 9.9pp gain), while on women, \byod offers little change (ranging from 0.6pp gain to 6.2pp drop).

Following \citet{radford2021learning}, we also examined associations of particular demographics with potentially harmful language. We replicate their setup with two classification tasks: (1) including race-gender intersection classes (e.g. ``black woman'', ``indian man'', etc.) and several harmful crime-related terms (``thief'', ``criminal'', ``suspicious person''); (2) including the same race-gender intersection classes and non-human terms (``animal'', ``gorilla'', ``chimpanzee'', ``orangutan''). We compute the frequency of misclassification of people into one of the harmful categories and run these experiments on FairFace and UTKFace separately. The results are shown in Table~\ref{tab:app_fairness_harm}. Unlike in \citet{radford2021learning}, we find that our models have a very small probability of classifying human faces as non-human, with a max score across all subgroups of 0.1\%. However, a significant proportion of people are misclassified as criminal. This again highlights the importance of dataset curation and the risks associated with zero-shot classification on models trained on web-scraped datasets.

\begin{table*}

\renewcommand{\arraystretch}{1.1}
        \rowcolors{2}{white}{light-light-gray}
\caption{Overall race, gender, and age classification accuracy of our two best {\small\texttt{xlarge}} baselines, Image-based $\cap$ CLIP score (L/14 30\%) for the filtering track and \pool, CLIP score + 4 external sources (upsampled 6x) for the \byod track. Race classification was binary (white or non-white) as in \citet{karkkainen2021fairface}.}
\setlength\tabcolsep{4pt}
\small
\centering
\begin{tabular}{llccc}
\toprule
Dataset & Track & Race & Gender & Age \\
\midrule
 \cellcolor{white}& Filtering & 86.4 & 91.7 & 34.3 \\
\cellcolor{white}\multirow{-2}{*}{FairFace}& \byod & 76.5 & 93.9 & 33.8 \\
\midrule
\cellcolor{white} & Filtering & 86.2 & 93.8 & 39.5 \\
\cellcolor{white}\multirow{-2}{*}{UTKFace}& \byod & 86.1 & 95.5 & 38.6 \\
\bottomrule
\end{tabular}
\label{tab:app_fairness_overall}
\end{table*}

\begin{table*}

\renewcommand{\arraystretch}{1.1}
        \rowcolors{3}{light-light-gray}{white}
\caption{Gender classification accuracy of our two best {\small\texttt{xlarge}} baselines, Image-based $\cap$ CLIP score (L/14 30\%) for the filtering track and \pool, CLIP score + 4 external sources (upsampled 6x) for the \byod track.}
\setlength\tabcolsep{4pt}

\small
\centering
FairFace
\\
\resizebox{\textwidth}{!}{
\begin{tabular}{llccccccc}
\toprule
\multirow{2}{*}{Track} & \multirow{2}{*}{Gender} & \multicolumn{7}{c}{Race} \\
& & Black & White & Indian & Latino/Hispanic & Middle Eastern & Southeast Asian & East Asian \\
\midrule
\cellcolor{white} & Male & 79.3 & 91.3 & 90.8 & 90.4 & 95.7 & 83.0 & 80.7 \\
\cellcolor{white} \multirow{-2}{*}{Filtering}& Female & 95.4 & 96.6 & 94.2 & 96.6 & 96.5 & 97.2 & 98.2 \\
\midrule
\cellcolor{white} & Male & 89.2 & 94.8 & 93.2 & 93.4 & 97.4 & 90.2 & 90.6 \\
\cellcolor{white} \multirow{-2}{*}{\byod}& Female & 89.2 & 96.0 & 94.2 & 96.0 & 96.2 & 97.1 & 97.0 \\
\bottomrule
\end{tabular}
}
\\
\vspace{10pt}
UTKFace
\\

\renewcommand{\arraystretch}{1.1}
        \rowcolors{3}{light-light-gray}{white}
\begin{tabular}{llccccccc}
\toprule
\multirow{2}{*}{Track} & \multirow{2}{*}{Gender} & \multicolumn{5}{c}{Race} \\
& & Black & White & Indian & Asian & Other \\
\midrule
\cellcolor{white}  & Male & 95.4 & 92.5 & 91.7 & 73.1 & 84.2 \\
\cellcolor{white} \multirow{-2}{*}{Filtering}& Female & 97.3 & 98.7 & 97.4 & 98.3 & 97.4 \\
\midrule
\cellcolor{white}  & Male & 96.8 & 95.9 & 94.7 & 85.7 & 90.4 \\
\cellcolor{white} \multirow{-2}{*}{\byod}& Female & 96.3 & 97.7 & 96.8 & 95.9 & 95.6 \\
\bottomrule
\end{tabular}
\label{tab:app_fairness_intersection}
\end{table*}

\begin{table*}

\renewcommand{\arraystretch}{1.1}
        \rowcolors{3}{light-light-gray}{white}
\caption{Harmful misclassification rates of our two best {\small\texttt{xlarge}} baselines, Image-based $\cap$ CLIP score (L/14 30\%) for the filtering track and \pool, CLIP score + 4 external sources (upsampled 6x) for the \byod track. While very few samples are misclassified as non-human, the filter track model assigns a crime-related label to a significant portion of people, and this is exacerbated by the \byod model in many cases.}
\setlength\tabcolsep{4pt}
\renewcommand{\arraystretch}{0.9}
\small
\centering
FairFace
\\
\resizebox{\textwidth}{!}{

\begin{tabular}{llccccccc}
\toprule
\multirow{2}{*}{Track} & & \multicolumn{7}{c}{Race} \\
& & Black & White & Indian & Latino/Hispanic & Middle Eastern & Southeast Asian & East Asian \\
\midrule
\cellcolor{white}& Crime-related & 4.4 & 24.3 & 8.8 & 14.3 & 23.7 & 7.4 & 8.6 \\
\cellcolor{white}\multirow{-2}{*}{Filtering} & Non-human & 0.0 & 0.0 & 0.0 & 0.0 & 0.0 & 0.0 & 0.0 \\
\midrule
 \cellcolor{white}& Crime-related & 18.4 & 16.8 & 21.5 & 22.9 & 20.9 & 35.3 & 30.9 \\
\cellcolor{white}\multirow{-2}{*}{\byod}& Non-human & 0.0 & 0.1 & 0.0 & 0.1 & 0.0 & 0.1 & 0.1 \\
\bottomrule
\end{tabular}
}
\\
\vspace{10pt}
UTKFace
\\

\renewcommand{\arraystretch}{1.1}
        \rowcolors{3}{light-light-gray}{white}
\begin{tabular}{llccccccc}
\toprule
\multirow{2}{*}{Track} & & \multicolumn{5}{c}{Race} \\
& & Black & White & Indian & Asian & Other \\
\midrule
\cellcolor{white}& Crime-related & 6.8 & 16.1 & 9.1 & 6.9 & 13.9 \\
\cellcolor{white}\multirow{-2}{*}{Filtering} & Non-human & 0.0 & 0.2 & 0.0 & 0.1 & 0.0 \\
\midrule
\cellcolor{white} & Crime-related & 12.8 & 10.8 & 15.2 & 13.2 & 18.6 \\
\cellcolor{white}\multirow{-2}{*}{\byod}& Non-human & 0.0 & 0.2 & 0.0 & 0.0 & 0.0 \\
\bottomrule
\end{tabular}
\label{tab:app_fairness_harm}
\end{table*}

\clearpage

\section{Extra figures and tables}
\label{sec:app-more-plots}

\begin{figure}[h]
    \centering
    \includegraphics[width=\textwidth]{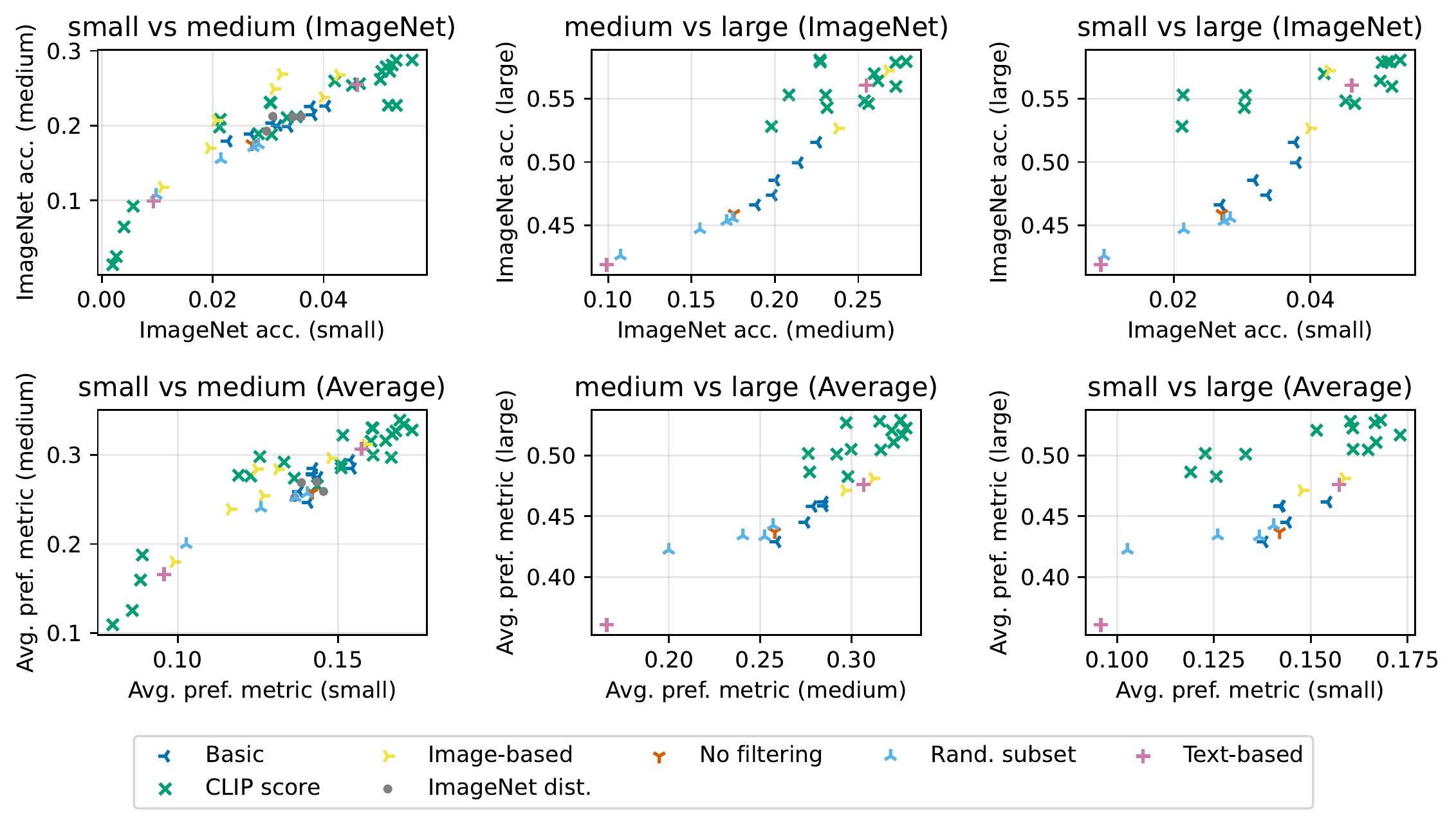}
    \caption{Improving downstream performance at smaller scales correlates positively with performance gains at larger scales. These trends suggests that dataset filtering can be studied effectively at smaller scales, even with less computational resources.}
    \label{fig:scaling-scatter-full}
\end{figure}

\begin{table}[h]
\caption{Rank correlation between the performance obtained with various filtering strategies at two different scales. Our experimental suggest that the ranking is relatively consistent between scales, especially for the adjacent scale pairs.}
\setlength\tabcolsep{6pt}
\renewcommand{\arraystretch}{1.1}
\footnotesize
\centering
\begin{tabular}{lccc}
\toprule
Metric   & {\small \texttt{small}} vs  {\small \texttt{medium}} & {\small \texttt{small}} vs {\small \texttt{large}} & {\small \texttt{medium}} vs {\small \texttt{large}} \\
\midrule
ImageNet acc. & 0.895 & 0.811 & 0.847 \\
Average pref. metric & 0.854 & 0.708 & 0.876 \\
\bottomrule
\end{tabular}
\label{tab:correlation}
\end{table}

\newpage

\begin{figure}
    \centering
    \begin{subfigure}[b]{\textwidth}
    \includegraphics[width=.95\linewidth]{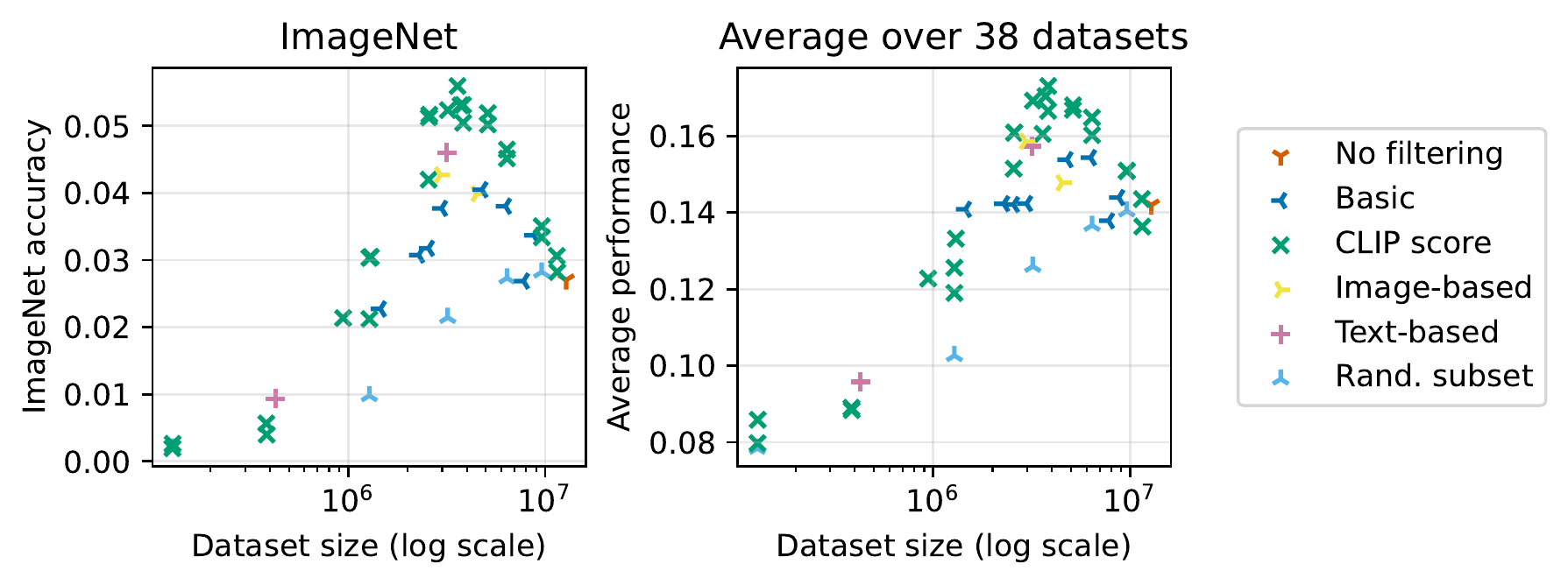}
\end{subfigure}
    \begin{subfigure}[b]{\textwidth}
    \includegraphics[width=.95\linewidth]{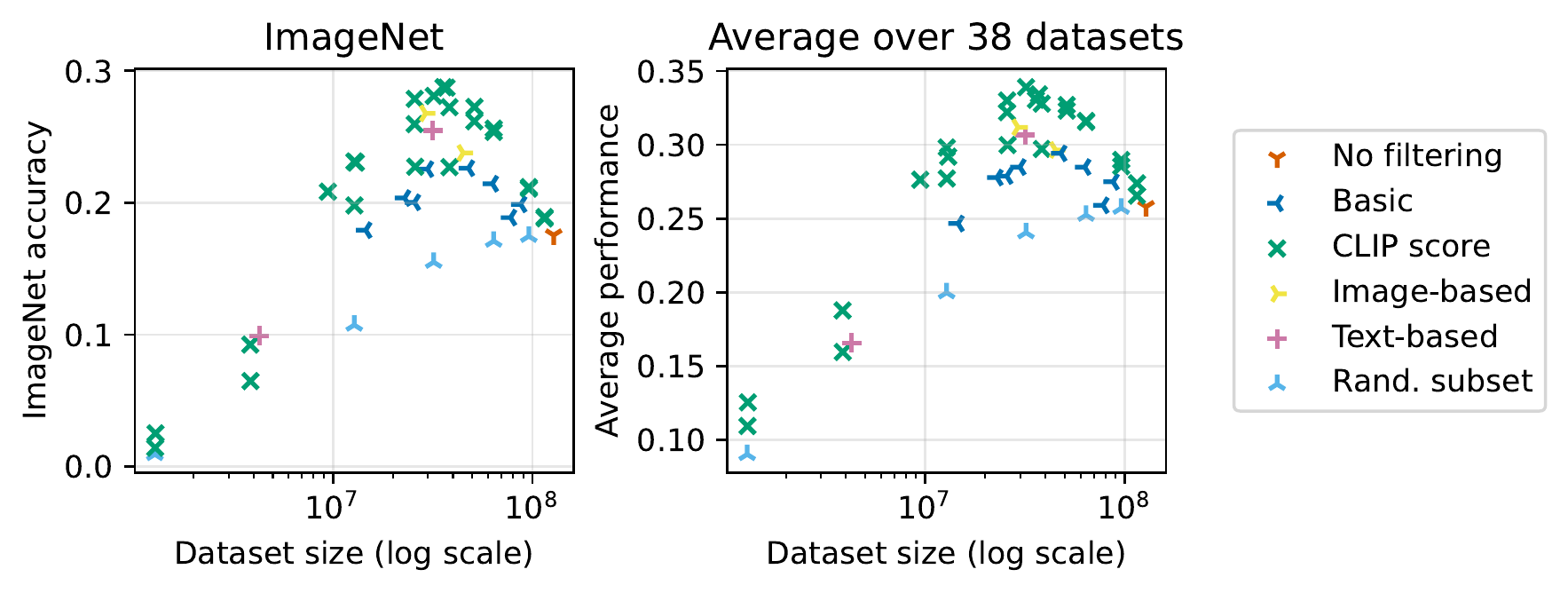}
\end{subfigure}
\begin{subfigure}[b]{\textwidth}
    \includegraphics[width=.95\linewidth]{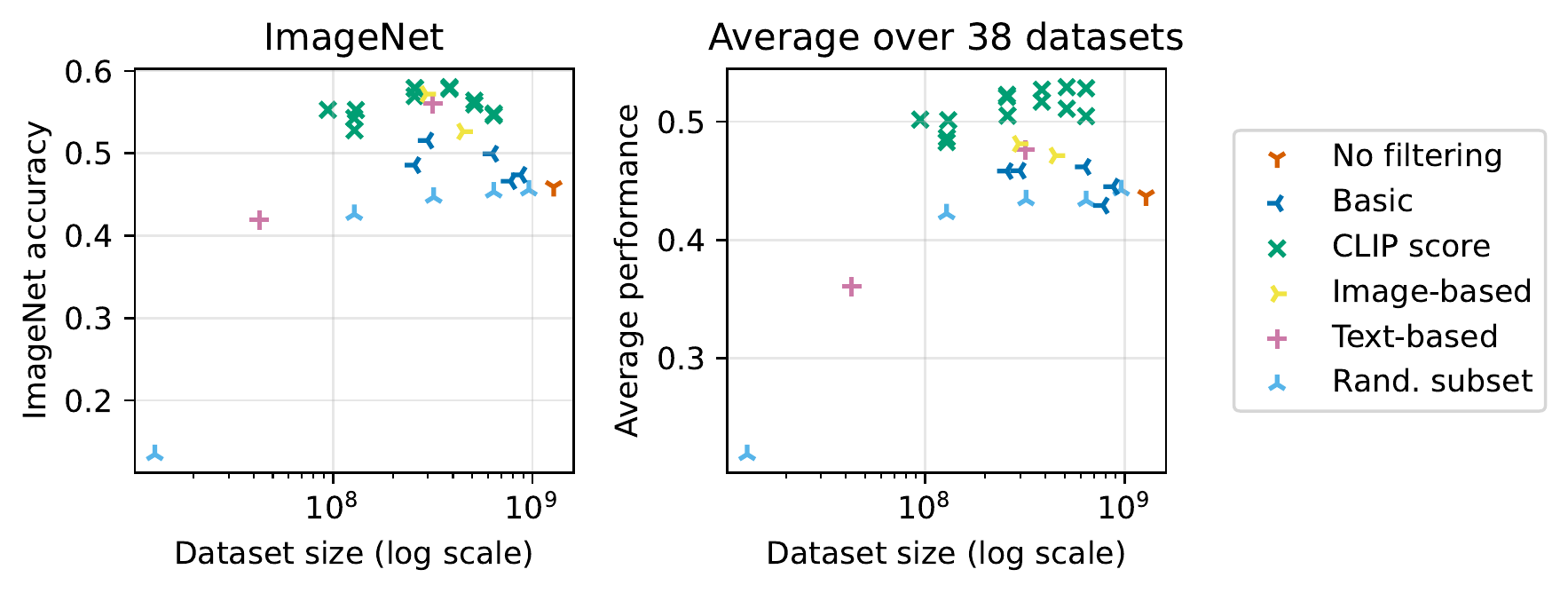}
\end{subfigure}
\caption{Performance as a function of the number of training samples from the {\small\texttt{small}} (top), {\small \texttt{medium}} (middle), and {\small\texttt{large}} (bottom) scales. There is a significant variance in accuracy even when accounting for the size of the training set.}
    \label{fig:training-samples-extra}
\end{figure}

\begin{figure}
    \centering
    \includegraphics[width=0.75\linewidth]{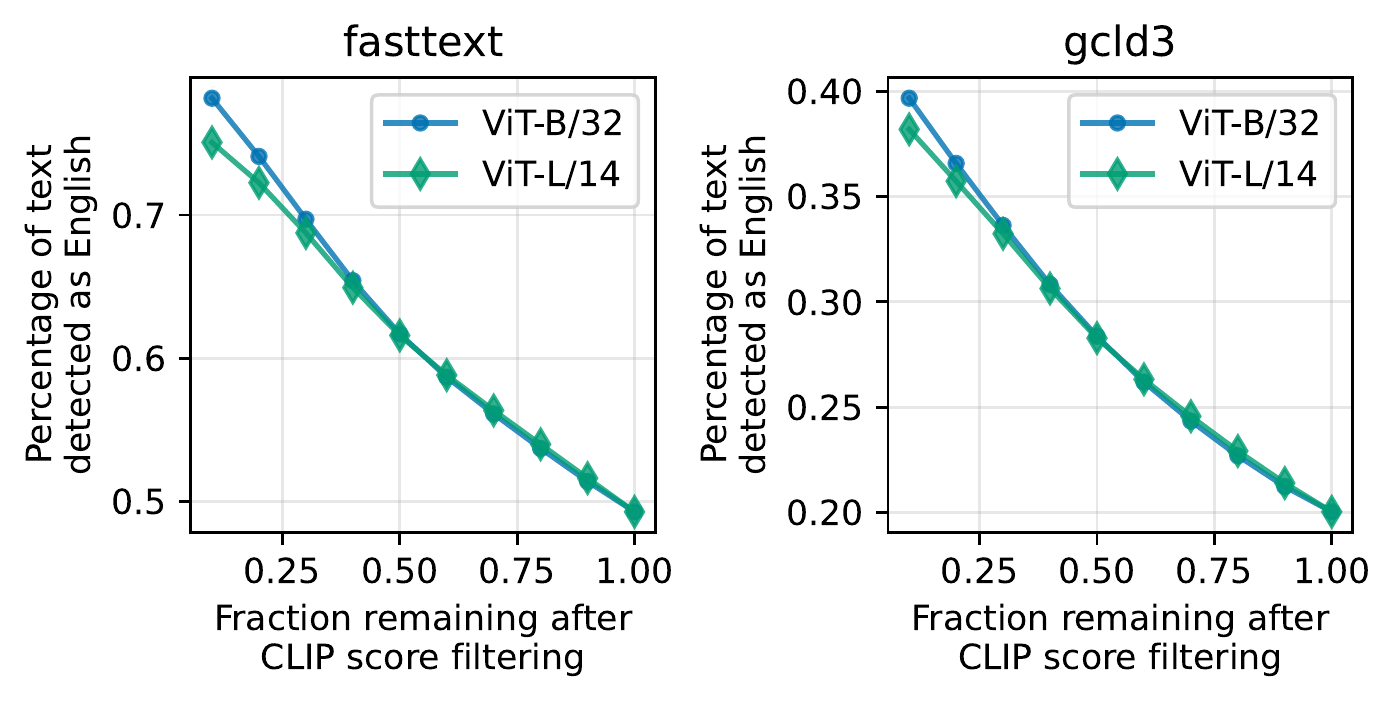}
    \caption{We examine the percentage of texts classified as English after taking the top fraction (on the x-axis) of the {\small \texttt{large}} billion pool as sorted by CLIP similarity score. We see that doing CLIP filtering implicitly does some English filtering, as image-text pairs with a higher CLIP score are more frequently classified as English.}
    \label{fig:clip_english}
\end{figure}

\begin{figure}
    \centering
    \includegraphics[width=\linewidth]{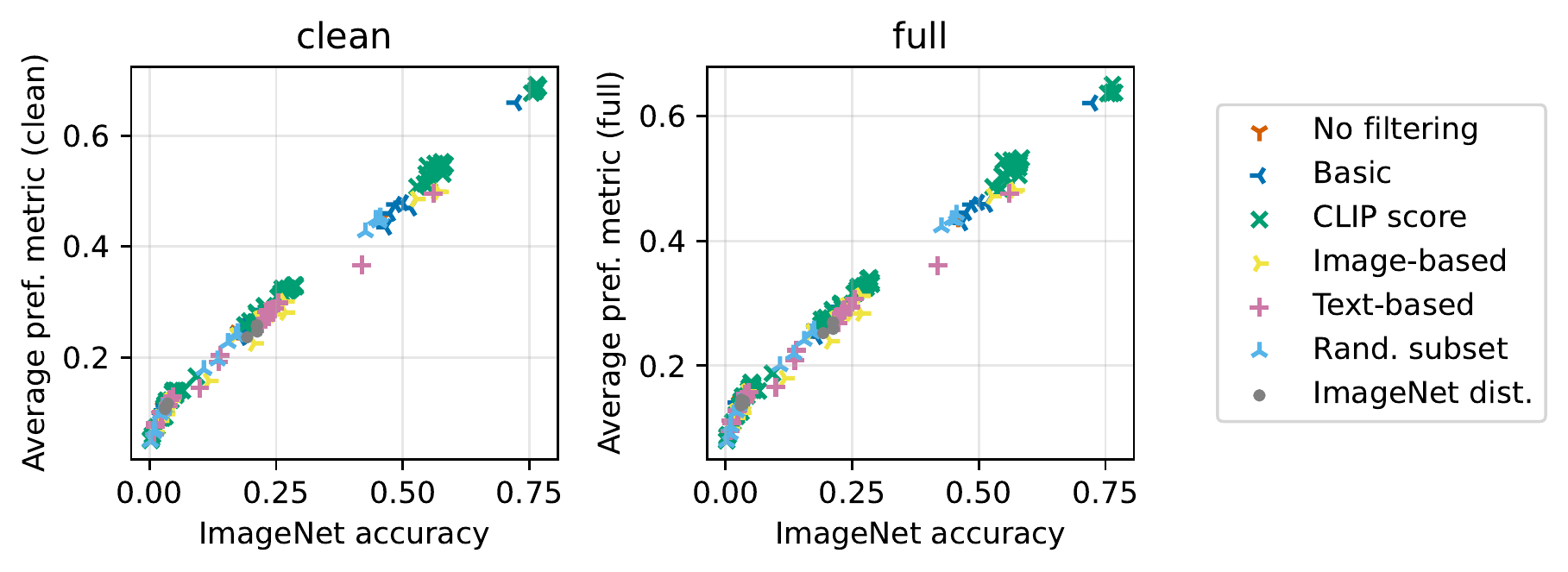}
    \caption{Correlation between ImageNet accuracy and average performance on our suite of evaluation tasks. While ImageNet accuracy strongly correlates with the average performance (both on the clean subset and the full suite), the same is not true for all individual datasets we study, as shown in Appendix \ref{sec:app-more-plots}.}
    \label{fig:imagenet-vs-all}
\end{figure}

\begin{table*}[h!]
 \rowcolors{3}{light-light-gray}{white}
  \rowcolors{3}{light-light-gray}{white}
    \small
    \centering
    \caption{Comparison of ViT-B/32 and ViT-B/16 models across different training datasets.}
\resizebox{\textwidth}{!}{
    \begin{tabular}{lcccccccc}\toprule
         Model & Training Dataset   & Training & Training & \multirow{2}{*}{ImageNet} & ImageNet & \multirow{2}{*}{VTAB} & \multirow{2}{*}{Retrieval} & Average over \\
        & & dataset size & steps & & dist. shifts &  & & 38 datasets\\\midrule
        ViT B/32 & \ours        & 1.4B & 13B & 0.692 & 0.551 & 0.577 & 0.538 & 0.579 \\
        ViT B/32 & OpenAI's WIT & 0.4B & 13B & 0.633 & 0.485 & 0.526 & 0.501 & 0.525 \\
        ViT B/32 & LAION-2B     & 2.3B & 34B & 0.666 & 0.522 & 0.561 & 0.560 & 0.569 \\
        ViT B/16 & \ours        & 1.4B & 13B & 0.735 & 0.608 & 0.621 & 0.578 & 0.615 \\
        ViT B/16 & OpenAI's WIT & 0.4B & 13B & 0.683 & 0.559 & 0.546 & 0.527 & 0.563 \\
        ViT B/16 & LAION-2B     & 2.3B & 34B & 0.702 & 0.566 & 0.572 & 0.583 & 0.587 \\
        \bottomrule 
    \end{tabular}
    }
    \label{tab:b16}
\end{table*}

\begin{table*}[h!]
 \rowcolors{3}{light-light-gray}{white}
  \rowcolors{3}{light-light-gray}{white}
    \small
    \centering
    \caption{Comparison at the xlarge scale between a 400M subset of \pool and OpenAI's WIT which also contains 400M samples. Our 400M subset is created by intersecting IN1k image clustering with English cld3 filtering, then taking the top 400M samples sorted by CLIP L14 score. Our model does better across the various evaluation groupings.}
\resizebox{\textwidth}{!}{
    \begin{tabular}{lcccccccc}\toprule
         Model & Training Dataset   & Training & Training & \multirow{2}{*}{ImageNet} & ImageNet & \multirow{2}{*}{VTAB} & \multirow{2}{*}{Retrieval} & Average over \\
        & & dataset size & steps & & dist. shifts &  & & 38 datasets\\\midrule
        ViT L/14 & top 400M by CLIP L14 of Image-based $\cap$ cld3        & 400M & 13B & 0.763 & 0.657 & 0.641 & 0.595 & 0.638 \\
        ViT L/14 & OpenAI's WIT & 400M & 13B & 0.755 & 0.649 & 0.586 & 0.543 & 0.617 \\
        \bottomrule 
    \end{tabular}
    }
    \label{tab:sample400m}
\end{table*}

\begin{figure}
    \centering
    \includegraphics[width=0.7\linewidth]{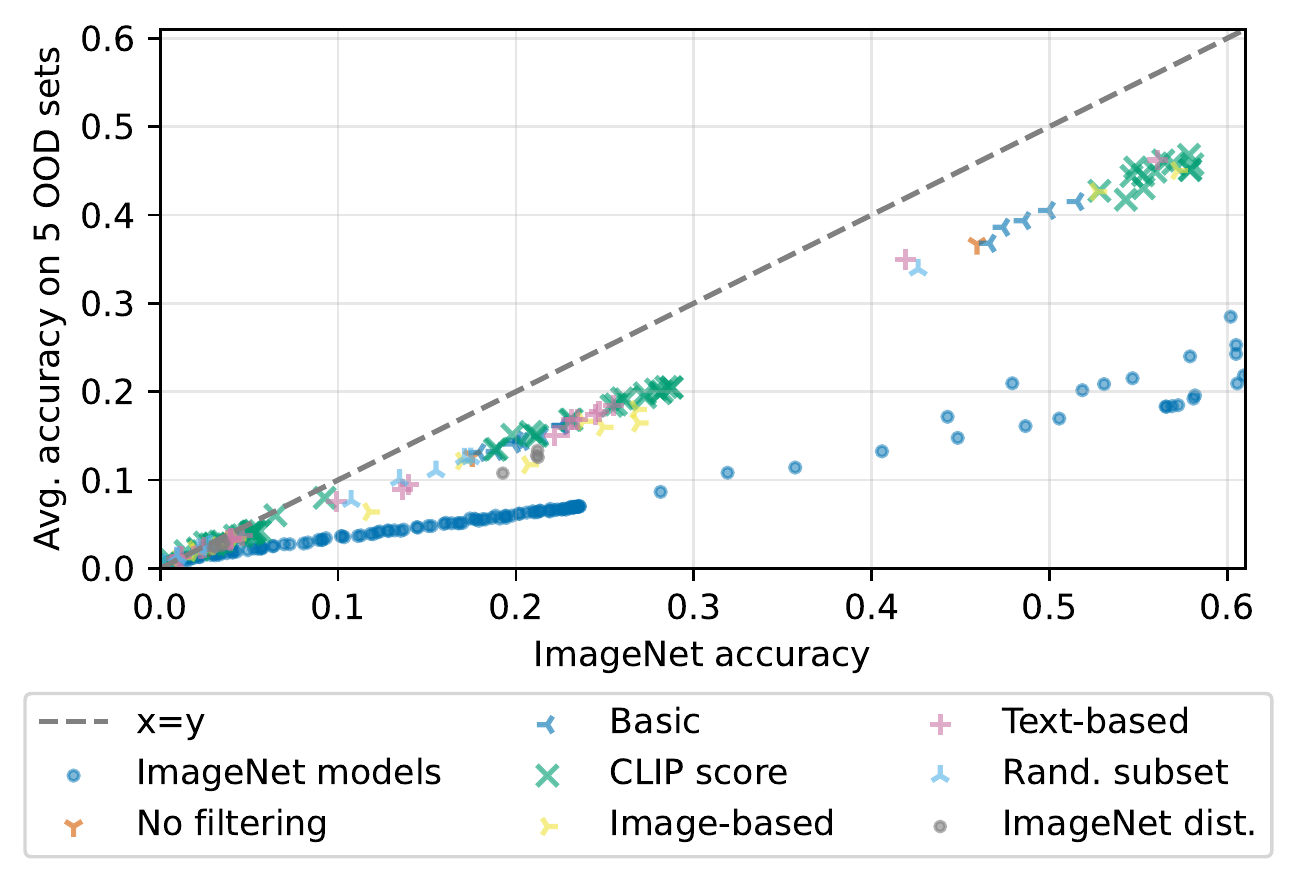}
    \caption{Zero-shot CLIP models trained with various filtering strategies form a reliable trend relating accuracy on ImageNet and related distribution shifts, exhibiting higher effective robustness when compared to ImageNet-trained models from \citet{taori2020measuring}.}
    \label{fig:robustness}
\end{figure}

\FloatBarrier

\begin{figure}
    \centering
    \includegraphics[width=\linewidth]{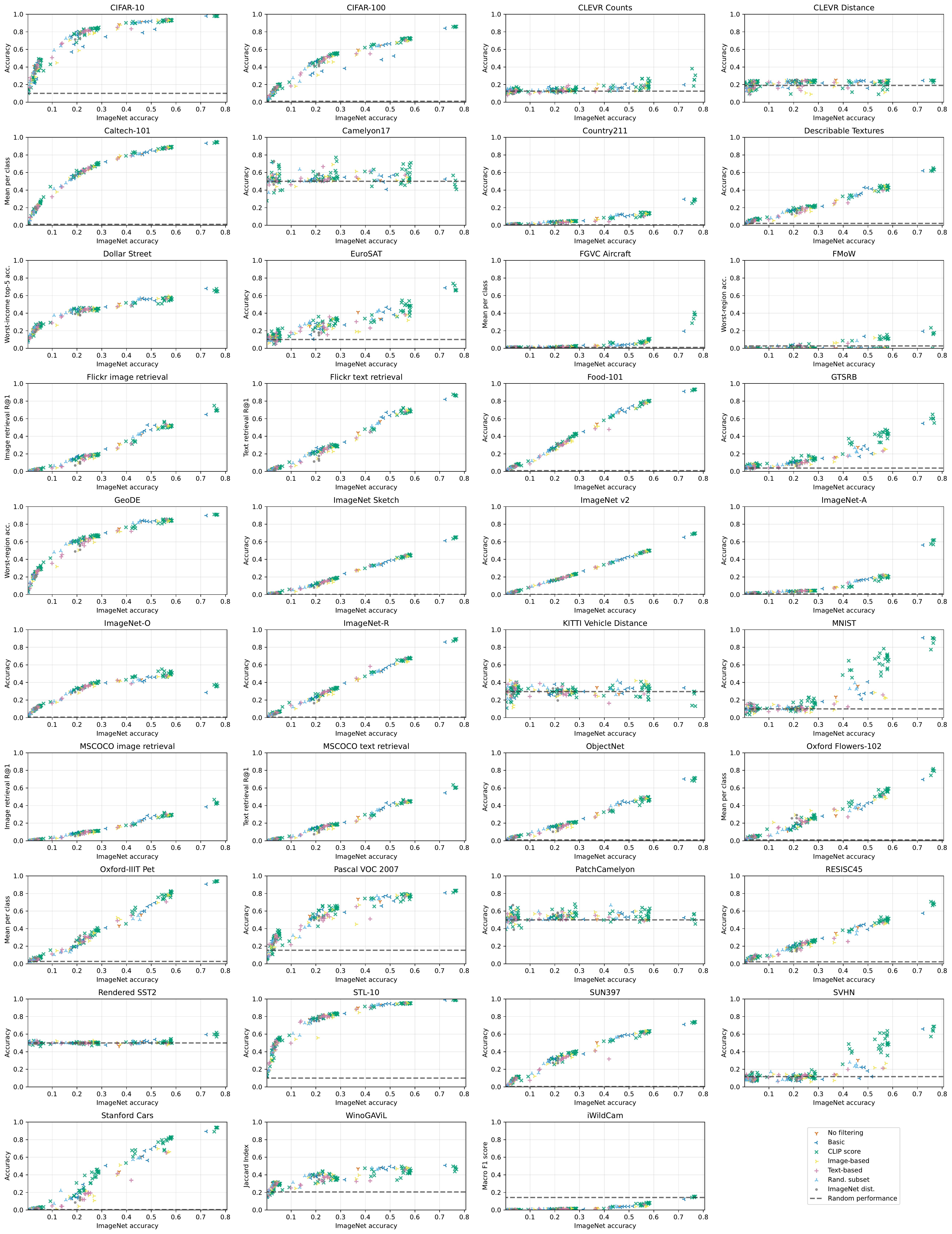}
    \caption{Zero-shot performance on other datasets is often positively correlated with that on ImageNet, but not always. In cases where ImageNet shows close to zero correlation with other datasets, performance on that dataset is often close to random chance.}
    \label{fig:imagenet-vs-all-breakdown}
\end{figure}
\thispagestyle{empty}

\label{app:full_results}
\begin{table*}

        \rowcolors{3}{light-light-gray}{white}
    \small
    \centering
    \caption{Baseline results for the filtering track, {\small \texttt{small}} scale.}
\resizebox{\textwidth}{!}{
    \begin{tabular}{lccccccc}\toprule
         \multirow{2}{*}{Filtering}   & Training & \multirow{2}{*}{ImageNet} & ImageNet & \multirow{2}{*}{VTAB} & \multirow{2}{*}{Retrieval} & Average over\\
        & dataset size &  & dist. shifts &  & & 38 datasets \\\midrule
        No filtering & 12.8M & 0.025 & 0.033 & 0.145 & 0.114 & 0.133  \\
        Random subset (75\%) & 9.6M & 0.028 & 0.037 & 0.153 & 0.110 & 0.140  \\
        Random subset (50\%) & 6.4M & 0.027 & 0.037 & 0.147 & 0.111 & 0.137  \\
        Random subset (25\%) & 3.2M & 0.022 & 0.032 & 0.130 & 0.099 & 0.126  \\
        Random subset (10\%) & 1.3M & 0.010 & 0.018 & 0.116 & 0.077 & 0.103  \\
        Random subset (1\%) & 128K & 0.002 & 0.005 & 0.095 & 0.049 & 0.078  \\
        Caption length & 8.7M & 0.034 & 0.040 & 0.148 & 0.109 & 0.143  \\
        Image size & 7.8M & 0.027 & 0.036 & 0.154 & 0.119 & 0.138  \\
        English (fasttext) & 6.3M & 0.038 & 0.045 & 0.164 & 0.124 & 0.154  \\
        English (fasttext) and caption length & 4.8M & 0.041 & 0.048 & 0.159 & 0.123 & 0.154  \\
        English (fasttext), caption length, and image size & 3.0M & 0.038 & 0.043 & 0.150 & 0.118 & 0.142  \\
        English (cld3) & 2.6M & 0.032 & 0.039 & 0.143 & 0.111 & 0.142  \\
        English (cld3) and caption length & 2.3M & 0.031 & 0.038 & 0.153 & 0.111 & 0.142  \\
        English (cld3), caption length, and image size & 1.5M & 0.023 & 0.030 & 0.154 & 0.092 & 0.141  \\
        
        CLIP B32 score top 1\% & 129K & 0.003 & 0.007 & 0.114 & 0.050 & 0.086  \\
        CLIP B32 score top 3\% & 384K & 0.006 & 0.014 & 0.104 & 0.055 & 0.089  \\
        CLIP B32 score top 10\% & 1.3M & 0.026 & 0.035 & 0.147 & 0.083 & 0.126  \\
        CLIP B32 score top 20\% & 2.6M & 0.051 & 0.056 & 0.173 & 0.114 & 0.161  \\
        CLIP B32 score top 30\% & 3.8M & 0.045 & 0.052 & 0.180 & 0.120 & 0.167  \\
        CLIP B32 score top 40\% & 5.1M & 0.052 & 0.057 & 0.173 & 0.123 & 0.167  \\
        CLIP B32 score top 50\% & 6.4M & 0.047 & 0.053 & 0.174 & 0.124 & 0.165  \\
        CLIP B32 score top 75\% & 9.6M & 0.033 & 0.043 & 0.161 & 0.121 & 0.151  \\
        CLIP B32 score top 90\% & 11.5M & 0.028 & 0.039 & 0.140 & 0.114 & 0.136  \\
        CLIP B32 threshold at 0.3 + English filter & 942K & 0.022 & 0.032 & 0.138 & 0.077 & 0.122  \\
        CLIP B32 threshold at 0.28 + English filter & 1.3M & 0.031 & 0.040 & 0.136 & 0.092 & 0.133  \\
        CLIP B32 threshold at 0.3 & 2.6M & 0.052 & 0.056 & 0.166 & 0.114 & 0.161  \\
        CLIP B32 score 1\% to 30\% & 3.7M & 0.053 & 0.058 & 0.185 & 0.113 & 0.170  \\
        CLIP B32 score 2\% to 30\% & 3.6M & 0.056 & 0.059 & 0.173 & 0.120 & 0.161  \\
        CLIP B32 score 5\% to 30\% & 3.2M & 0.052 & 0.055 & 0.177 & 0.115 & 0.169  \\
        
        CLIP L14 score top 1\% & 128K & 0.002 & 0.007 & 0.111 & 0.050 & 0.080  \\
        CLIP L14 score top 3\% & 386K & 0.004 & 0.009 & 0.110 & 0.052 & 0.088  \\
        CLIP L14 score top 10\% & 1.3M & 0.021 & 0.033 & 0.131 & 0.075 & 0.119  \\
        CLIP L14 score top 20\% & 2.6M & 0.042 & 0.051 & 0.165 & 0.100 & 0.151  \\
        CLIP L14 score top 30\% & 3.8M & 0.051 & 0.055 & 0.190 & 0.119 & 0.173  \\
        CLIP L14 score top 40\% & 5.1M & 0.050 & 0.054 & 0.173 & 0.119 & 0.168  \\
        CLIP L14 score top 50\% & 6.4M & 0.045 & 0.052 & 0.164 & 0.122 & 0.160 \\
        CLIP L14 score top 75\% & 9.6M & 0.035 & 0.043 & 0.164 & 0.120 & 0.151  \\
        CLIP L14 score top 90\% & 11.5M & 0.031 & 0.038 & 0.154 & 0.116 & 0.144  \\
        
        Image-based clustering (ImageNet1k) & 2.9M & 0.043 & 0.047 & 0.178 & 0.121  & 0.159 \\
        Image-based clustering (ImageNet21k) & 4.5M & 0.035 & 0.045 & 0.154 & 0.122 & 0.148  \\
        Image-based sampling, $\alpha$=0 & 12.8M & 0.019 & 0.030 & 0.144 & 0.095 & 0.127  \\
        Image-based sampling, $\alpha$=0.2 & 12.8M & 0.031 & 0.036 & 0.133 & 0.100 & 0.131 \\
        Image-based sampling, $\alpha$=0.5 & 12.8M & 0.032 & 0.038 & 0.129 & 0.096 & 0.125  \\
        Image-based sampling, $\alpha$=1 & 12.8M & 0.021 & 0.028 & 0.128 & 0.078 & 0.116  \\
        Image-based sampling, $\alpha$=2 & 12.8M & 0.011 & 0.017 & 0.116 & 0.065 & 0.099 \\
        
        ImageNet distance (L14, top 30\%) and English & 2.0M & 0.031 & 0.039 & 0.163 & 0.103 & 0.145  \\
        ImageNet distance (L14, top 20\%) & 2.6M & 0.030 & 0.035 & 0.155 & 0.102 & 0.136  \\
        ImageNet distance (L14, top 30\%) & 3.9M & 0.034 & 0.041 & 0.151 & 0.106 & 0.139  \\
        ImageNet distance (L14, top 40\%) & 5.1M & 0.036 & 0.040 & 0.151 & 0.118 & 0.143 \\
        
        Text-based clustering (ImageNet1k) & 427K & 0.009 & 0.016 & 0.120 & 0.056 & 0.096  \\
        Text-based clustering (ImageNet21k) & 3.2M & 0.046 & 0.052 & 0.169 & 0.125 & 0.157  \\
        Text-based sampling with average score, $\alpha$=0 & 12.8M & 0.011 & 0.020 & 0.128 & 0.079 & 0.112  \\
        Text-based sampling with average score, $\alpha$=0.5 & 12.8M & 0.023 & 0.035 & 0.127 & 0.092 & 0.128  \\
        Text-based sampling with average score, $\alpha$=1 & 12.8M & 0.040 & 0.044 & 0.163 & 0.115 & 0.155  \\
        Text-based sampling with average score, $\alpha$=1.2 & 12.8M & 0.038 & 0.045 & 0.150 & 0.112 & 0.143  \\
        Text-based sampling with max score, $\alpha$=0 & 12.8M & 0.012 & 0.020 & 0.126 & 0.074 & 0.107  \\
        Text-based sampling with max score, $\alpha$=0.5 & 12.8M & 0.025 & 0.033 & 0.134 & 0.093 & 0.129  \\
        Text-based sampling with max score, $\alpha$=1 & 12.8M & 0.040 & 0.046 & 0.159 & 0.116 & 0.150  \\
        Text-based sampling with max score, $\alpha$=1.2 & 12.8M & 0.040 & 0.050 & 0.161 & 0.113 & 0.152  \\
        
        Intersect IN1k image clustering and CLIP B32 score top 30\% & 1.4M & 0.049 & 0.053 & 0.150 & 0.103 & 0.148  \\
        Intersect IN1k image clustering and CLIP L14 score top 30\% & 1.4M & 0.039 & 0.045 & 0.162 & 0.094 & 0.145  \\
        Intersect IN21k image clustering and CLIP B32 score top 30\% & 2.1M & 0.052 & 0.057 & 0.179 & 0.112 & 0.167  \\
        Intersect IN21k image clustering and CLIP L14 score top 30\% & 2.1M & 0.047 & 0.053 & 0.176 & 0.110 & 0.163  \\
        \bottomrule
    \end{tabular}
    }
\label{tab:full-small}
\end{table*}
\newpage
\begin{table*}
 \rowcolors{3}{light-light-gray}{white}
    \small
    \centering
    \caption{Baseline results for the filtering track, {\small \texttt{medium}} scale.}
\resizebox{\textwidth}{!}{
    \begin{tabular}{lcccccc}\toprule
         \multirow{2}{*}{Filtering}   & Training & \multirow{2}{*}{ImageNet} & ImageNet & \multirow{2}{*}{VTAB} & \multirow{2}{*}{Retrieval} & Average over \\
        & dataset size &  & dist. shifts &  & & 38 datasets\\\midrule
        No filtering & 128M & 0.176 & 0.152 & 0.259 & 0.219 & 0.258  \\
        Random subset (75\%) & 96.0M & 0.175 & 0.154 & 0.265 & 0.219 & 0.257  \\
        Random subset (50\%) & 64.0M & 0.171 & 0.151 & 0.258 & 0.216 & 0.252  \\
        Random subset (25\%) & 32.0M & 0.155 & 0.136 & 0.246 & 0.203 & 0.240  \\
        Random subset (10\%) & 12.8M & 0.107 & 0.095 & 0.210 & 0.144 & 0.200  \\
        Random subset (1\%) & 1.3M & 0.009 & 0.017 & 0.102 & 0.065 & 0.090  \\
        
        Caption length & 87.5M & 0.199 & 0.172 & 0.275 & 0.236 & 0.275  \\
        Image size & 77.8M & 0.189 & 0.163 & 0.248 & 0.231 & 0.259  \\
        English (fasttext) & 63.0M & 0.214 & 0.182 & 0.290 & 0.246 & 0.285  \\
        English (fasttext) and caption length & 47.8M & 0.226 & 0.193 & 0.284 & 0.251 & 0.285  \\
        English (fasttext), caption length, and image size & 29.8M & 0.226 & 0.193 & 0.297 & 0.253 & 0.294  \\
        English (cld3) & 25.6M & 0.200 & 0.175 & 0.296 & 0.235 & 0.279  \\
        English (cld3) and caption length & 22.9M & 0.204 & 0.175 & 0.287 & 0.243 & 0.278  \\
        English (cld3), caption length, and image size & 14.6M & 0.179 & 0.159 & 0.243 & 0.216 & 0.247  \\
        
        CLIP B32 score top 1\% & 1.3M & 0.025 & 0.037 & 0.140 & 0.076 & 0.126  \\
        CLIP B32 score top 3\% & 3.9M & 0.093 & 0.096 & 0.205 & 0.128 & 0.188  \\
        CLIP B32 score top 10\% & 12.8M & 0.231 & 0.199 & 0.305 & 0.206 & 0.298  \\
        CLIP B32 score top 20\% & 25.7M & 0.279 & 0.234 & 0.337 & 0.241 & 0.330  \\
        CLIP B32 score top 30\% & 38.4M & 0.285 & 0.240 & 0.355 & 0.253 & 0.338  \\
        CLIP B32 score top 40\% & 51.3M & 0.273 & 0.227 & 0.333 & 0.257 & 0.324  \\
        CLIP B32 score top 50\% & 64.0M & 0.256 & 0.219 & 0.322 & 0.259 & 0.316  \\
        CLIP B32 score top 75\% & 96.1M & 0.211 & 0.180 & 0.301 & 0.238 & 0.290  \\
        CLIP B32 score top 90\% & 115M & 0.189 & 0.165 & 0.279 & 0.229 & 0.274  \\
        CLIP B32 threshold at 0.3 + English filter & 9.4M & 0.208 & 0.184 & 0.292 & 0.210 & 0.276  \\
        CLIP B32 threshold at 0.28 + English filter & 13.0M & 0.230 & 0.198 & 0.307 & 0.233 & 0.292  \\
        CLIP B32 threshold at 0.3 & 25.9M & 0.282 & 0.233 & 0.340 & 0.243 & 0.333  \\
        CLIP B32 score 1\% to 30\% & 37.1M & 0.287 & 0.238 & 0.347 & 0.253 & 0.334  \\
        CLIP B32 score 2\% to 30\% & 35.9M & 0.288 & 0.238 & 0.338 & 0.248 & 0.330  \\
        CLIP B32 score 5\% to 30\% & 32.0M & 0.281 & 0.230 & 0.352 & 0.254 & 0.339  \\
        
        CLIP L14 score top 1\% & 1.3M & 0.014 & 0.025 & 0.136 & 0.062 & 0.109  \\
        CLIP L14 score top 3\% & 3.9M & 0.065 & 0.077 & 0.176 & 0.103 & 0.160  \\
        CLIP L14 score top 10\% & 12.8M & 0.198 & 0.183 & 0.283 & 0.188 & 0.277  \\
        CLIP L14 score top 20\% & 25.7M & 0.260 & 0.225 & 0.326 & 0.235 & 0.322  \\
        CLIP L14 score top 30\% & 38.4M & 0.273 & 0.230 & 0.338 & 0.251 & 0.328  \\
        CLIP L14 score top 40\% & 51.2M & 0.262 & 0.226 & 0.330 & 0.260 & 0.327  \\
        CLIP L14 score top 50\% & 64.1M & 0.254 & 0.218 & 0.322 & 0.262 & 0.315  \\
        CLIP L14 score top 75\% & 96.1M & 0.212 & 0.180 & 0.287 & 0.242 & 0.285  \\
        CLIP L14 score top 90\% & 115M & 0.188 & 0.164 & 0.258 & 0.225 & 0.266  \\
        
        Image-based clustering (ImageNet1k) & 29.2M & 0.268 & 0.213 & 0.319 & 0.256 & 0.312  \\
        Image-based clustering (ImageNet21k) & 45.1M & 0.238 & 0.198 & 0.304 & 0.252 & 0.312  \\
        Image-based sampling, $\alpha$=0 & 128M & 0.170 & 0.150 & 0.266 & 0.209 & 0.254   \\
        Image-based sampling, $\alpha$=0.2 & 128M & 0.249 & 0.193 & 0.292 & 0.221 & 0.284 \\
        Image-based sampling, $\alpha$=0.5 & 128M & 0.269 & 0.196 & 0.301 & 0.216 & 0.284  \\
        Image-based sampling, $\alpha$=1 & 128M & 0.207 & 0.145 & 0.264 & 0.166 & 0.239 \\
        Image-based sampling, $\alpha$=2 & 128M & 0.118 & 0.082 & 0.207 & 0.110 & 0.180 \\
        ImageNet distance (L14, top 30\%) and English & 19.8M & 0.212 & 0.158 & 0.272 & 0.178 & 0.259  \\
        ImageNet distance (L/14, top 20\%) & 25.8M & 0.193 & 0.138 & 0.276 & 0.176 & 0.252  \\
        ImageNet distance (L/14, top 30\%) & 38.5M & 0.212 & 0.159 & 0.283 & 0.201 & 0.269  \\
        ImageNet distance (L/14, top 40\%) & 51.3M & 0.212 & 0.165 & 0.273 & 0.212 & 0.270  \\
        
        Text-based clustering (ImageNet1k) & 4.3M & 0.099 & 0.090 & 0.173 & 0.109 & 0.166  \\
        Text-based clustering (ImageNet21k) & 31.7M & 0.255 & 0.215 & 0.328 & 0.249 & 0.307 \\
        Text-based sampling with average score, $\alpha$=0 & 128M & 0.136 & 0.110 & 0.213 & 0.140 & 0.209 \\
        Text-based sampling with average score, $\alpha$=0.5 & 128M & 0.222 & 0.178 & 0.273 & 0.206 & 0.269  \\
        Text-based sampling with average score, $\alpha$=1 & 128M & 0.245 & 0.204 & 0.302 & 0.251 & 0.293  \\
        Text-based sampling with average score, $\alpha$=1.2 & 128M & 0.231 & 0.200 & 0.298 & 0.240 & 0.289  \\
        Text-based sampling with max score, $\alpha$=0 & 128M & 0.140 & 0.116 & 0.242 & 0.138 & 0.225  \\
        Text-based sampling with max score, $\alpha$=0.5 & 128M & 0.229 & 0.190 & 0.290 & 0.205 & 0.283  \\
        Text-based sampling with max score, $\alpha$=1 & 128M & 0.247 & 0.209 & 0.300 & 0.241 & 0.295  \\
        Text-based sampling with max score, $\alpha$=1.2 & 128M & 0.235 & 0.200 & 0.298 & 0.239 & 0.290  \\
        
        Intersect IN1k image clustering and CLIP B32 score top 30\% & 14.2M & 0.305 & 0.243 & 0.342 & 0.250 & 0.328  \\
        Intersect IN1k image clustering and CLIP L14 score top 30\% & 14.0M & 0.297 & 0.239 & 0.346 & 0.231 & 0.328  \\
        Intersect IN21k image clustering and CLIP B32 score top 30\% & 21.1M & 0.298 & 0.244 & 0.347 & 0.256 & 0.336  \\
        Intersect IN21k image clustering and CLIP L14 score top 30\% & 20.8M & 0.290 & 0.241 & 0.339 & 0.244 & 0.328  \\
        \bottomrule
    \end{tabular}
}
    \label{tab:full-medium}
\end{table*}
\newpage
\begin{table*}
 \rowcolors{3}{light-light-gray}{white}
    \small
    \centering
    \caption{Baseline results for the filtering track, {\small \texttt{large}} scale.}
\resizebox{\textwidth}{!}{
    \begin{tabular}{lcccccc}\toprule
         \multirow{2}{*}{Filtering}   & Training & \multirow{2}{*}{ImageNet} & ImageNet & \multirow{2}{*}{VTAB} & \multirow{2}{*}{Retrieval} & Average over\\
        & dataset size &  & dist. shifts &  & & 38 datasets\\\midrule
        No filtering & 1.28B & 0.459 & 0.378 & 0.426 & 0.419 & 0.437 \\
        Random subset (75\%) & 960M & 0.456 & 0.379 & 0.435 & 0.415 & 0.442 \\
        Random subset (50\%) & 640M & 0.453 & 0.377 & 0.427 & 0.413 & 0.433 \\
        Random subset (25\%) & 320M & 0.447	& 0.373 & 0.424 & 0.407 & 0.434 \\
        Random subset (10\%) & 128M & 0.426 & 0.350 & 0.417 & 0.396 & 0.442 \\
        Random subset (1\%) & 12.8M & 0.135 & 0.118 & 0.219 & 0.135 & 0.218 \\
        
        Caption length & 874M & 0.474 & 0.392 & 0.438 & 0.443 & 0.445 \\
        Image size & 777M & 0.466 & 0.375 & 0.421 & 0.438 & 0.429 \\
        English (fasttext) & 630M & 0.500 & 0.414 & 0.449 & 0.460 & 0.462 \\
        English (fasttext), caption length, and image size & 298M & 0.516 & 0.423 & 0.446 & 0.480 & 0.458 \\
        English (cld3) & 256M & 0.486 & 0.405 & 0.462 & 0.472 & 0.458 \\
        CLIP B32 score top 10\% & 128M & 0.543 & 0.440 & 0.471 & 0.435 & 0.483 \\
        CLIP B32 score top 20\% & 257M & 0.578 & 0.465 & 0.516 & 0.463 & 0.515 \\
        CLIP B32 score top 30\% & 384M & 0.578 & 0.466 & 0.525 & 0.475 & 0.527 \\
        CLIP B32 score top 40\% & 512M & 0.560 & 0.454 & 0.512 & 0.478 & 0.511 \\
        CLIP B32 score top 50\% & 640M & 0.546 & 0.450 & 0.504 & 0.484 & 0.505 \\
        CLIP B32 threshold at 0.3 + English filter & 94.3M & 0.553 & 0.447 & 0.511 & 0.482 & 0.502 \\
        CLIP B32 threshold at 0.28 + English filter & 130M & 0.553 & 0.453 & 0.510 & 0.495 & 0.501 \\
        CLIP B32 threshold at 0.3 & 258M & 0.579 & 0.464 & 0.501 & 0.465 & 0.505 \\
        CLIP L14 score top 10\% & 128M & 0.528 & 0.444 & 0.482 & 0.413 & 0.486 \\
        CLIP L14 score top 20\% & 257M & 0.570 & 0.466 & 0.524 & 0.455 & 0.521 \\
        CLIP L14 score top 30\% & 384M & 0.578 & 0.474 & 0.538 & 0.466 & 0.529 \\
        CLIP L14 score top 40\% & 512M & 0.564 & 0.462 & 0.533 & 0.468 & 0.529 \\
        CLIP L14 score top 50\% & 641M & 0.548 & 0.455 & 0.539 & 0.469 & 0.528 \\
        Image-based clustering (ImageNet1k) & 294M & 0.572 & 0.454 & 0.483 & 0.481 & 0.481 \\
        Image-based clustering (ImageNet21k) & 450M & 0.527 & 0.433 & 0.468 & 0.463 & 0.471 \\
        Text-based clustering (ImageNet1k) & 42.7M & 0.419 & 0.355 & 0.340 & 0.309 & 0.361 \\
        Text-based clustering (ImageNet21k) & 317M & 0.561 & 0.465 & 0.465 & 0.479 & 0.476 \\
        Intersect IN1k image clustering and CLIP B32 score top 30\% & 143M & 0.632 & 0.498 & 0.525 & 0.504 & 0.528 \\
        Intersect IN1k image clustering and CLIP L14 score top 30\% & 140M & 0.631 & 0.508 & 0.546 & 0.498 & 0.537 \\
        Intersect IN21k image clustering and CLIP B32 score top 30\% & 211M & 0.605 & 0.481 & 0.531 & 0.494 & 0.519  \\
        Intersect IN21k image clustering and CLIP L14 score top 30\% & 208M & 0.506 & 0.416 & 0.466 & 0.424 & 0.471  \\
 \bottomrule 
    \end{tabular}
    }
    \label{tab:full-large}
\end{table*}

\begin{table*}
 \rowcolors{3}{light-light-gray}{white}
  \rowcolors{3}{light-light-gray}{white}
    \small
    \centering
    \caption{Baseline results for the filtering track, {\small \texttt{xlarge}} scale.}
\resizebox{\textwidth}{!}{
    \begin{tabular}{lcccccc}\toprule
         \multirow{2}{*}{Filtering}   & Training & \multirow{2}{*}{ImageNet} & ImageNet & \multirow{2}{*}{VTAB} & \multirow{2}{*}{Retrieval} & Average over \\
        & dataset size &  & dist. shifts &  & & 38 datasets\\\midrule
        No filtering & 12.8B & 0.723 & 0.612 & 0.611 & 0.569 & 0.621  \\
        CLIP B32 score top 30\% & 3.84B & 0.764 & 0.640 & 0.628 & 0.599 & 0.638  \\
        CLIP B32 threshold at 0.28 + English filter & 1.3B & 0.755 & 0.637 & 0.624 & 0.620 & 0.636 \\
        CLIP L14 score top 20\% & 2.56B & 0.761 & 0.649 & 0.630 & 0.575 & 0.636 \\
        CLIP L14 score top 25\% & 3.2B & 0.768 & 0.656 & 0.621 & 0.585 & 0.637  \\
        CLIP L14 score top 30\% & 3.84B & 0.764 & 0.655 & 0.643 & 0.588 & 0.650  \\
        Intersect IN1k image clustering and CLIP L14 score top 30\% & 1.38B & 0.792 & 0.679 & 0.652 & 0.608 & 0.663 \\
        \bottomrule 
    \end{tabular}
    }
    \label{tab:full-xlarge}
\end{table*}

\clearpage
\input{datasheet}

%% file: datasheet.tex
\section{Datasheet}

\begin{enumerate}[label=Q\arabic*]
\subsection{Motivation}

\item \textbf{For what purpose was the dataset created?} Was there a specific task in mind? Was there a specific gap that needed to be filled? Please provide a description.

\begin{itemize}
\item The purpose of \datanet and the associated \pool dataset is to enable study of what makes a strong image-text dataset, which supports a broad range of applications. Prior work mainly focuses on data curation in the context of supervised datasets and smaller scales. For a fuller treatment see Section~\ref{sec:relatedwork}. In our initial release of \datanet we focus on 38 downstream image classification and image retrieval tasks. For details see Section \ref{sec:evaluation} and Appendix \ref{sec:app-eval}.
\end{itemize}

\item \textbf{Who created the dataset (e.g., which team, research group) and on behalf of which entity (e.g., company, institution, organization)?}

\begin{itemize}
\item \datanet and \pool were created by a group of researchers with the following affiliations, listed in alphabetical order: Allen Institute for Artificial Intelligence (AI2), Apple, Columbia University, Google Research, Graz University of Technology, Hebrew University, Juelich Supercomputing Center, LAION, Research Center Juelich, StabilityAI, Tel Aviv University, University of Illinois Urbana-Champaign, University of Texas at Austin, University of Washington.
\end{itemize}

\item \textbf{Who funded the creation of the dataset?} If there is an associated grant, please provide the name of the grantor and the grant name and number.

\begin{itemize}
\item Compute for this research was generously provided by StabilityAI. For more specific acknowledgments, see the acknowledgment section at the end of the main paper.
\end{itemize}

\item \textbf{Any other comments?}

\begin{itemize}
\item We hope that \pool will help to facilitate data-centric questions in ML and AI towards the next generation of web-scale datasets, that 1) yield higher accuracy models and 2) models that are safer and more equitable.
\end{itemize}

\subsection{Composition}

\item \textbf{What do the instances that comprise the dataset represent (e.g., documents, photos, people, countries)?} \textit{Are there multiple types of instances (e.g., movies, users, and ratings; people and interactions between them; nodes and edges)? Please provide a description.}

\begin{itemize}
\item Each instance is a pair of url and corresponding image alt-text. The url points to an image that a user can then try to download. Each sample is also tagged with metadata, discussed in Q25.
\end{itemize}

\item \textbf{How many instances are there in total (of each type, if appropriate)?}

\begin{itemize}
\item There are 12.8B instances in \pool. For breakdowns and statistics see Appendix \ref{app:pool-stats}.
\end{itemize}

\item \textbf{Does the dataset contain all possible instances or is it a sample (not necessarily random) of instances from a larger set?} \textit{If the dataset is a sample, then what is the larger set? Is the sample representative of the larger set (e.g., geographic coverage)? If so, please describe how this representativeness was validated/verified. If it is not representative of the larger set, please describe why not (e.g., to cover a more diverse range of instances, because instances were withheld or unavailable).}

\begin{itemize}
\item We find $\sim$88B possible samples in common crawl. These samples are globally shuffled to ensure i.i.d. sampling for all sampling based parts of the downstream pipeline. Of these samples we attempt to download $\sim$40B samples. Due to various download issues, such as dead links and throttling, we are able to successfully download $\sim$16.8B samples. After NSFW filtering and evaluation set deduplication we end up with $\sim$13.1B viable samples, from which we randomly sample 12.8B for \pool. For a complete treatment and visualization of our data processing funnel, see Appendix \ref{app:metadata}. For each sample we also release metadata shown in Table \ref{tab:app_metadata}.
\end{itemize}

\item \textbf{What data does each instance consist of?} \textit{“Raw” data (e.g., unprocessed text or images) or features? In either case, please provide a description.}

\begin{itemize}
\item Each sample contains an image url for download and an associated alt-text caption. Additionally, each sample contains metadata fields shown in Table \ref{tab:app_metadata} (e.g., image aspect ratio and CLIP features).
\end{itemize}

\item \textbf{Is there a label or target associated with each instance?} \textit{If so, please provide a description.}

\begin{itemize}
\item We do not provide any category labels; however, the text associated with each image can be considered a soft, noisy label for each sample. Such labels are common in modern image-text training paradigms (e.g., image-text representation alignment, image captioning objectives, text-conditional image generation objectives, etc.).
\end{itemize}

\item \textbf{Is any information missing from individual instances?} \textit{If so, please provide a description, explaining why this information is missing (e.g., because it was unavailable). This does not include intentionally removed information, but might include, e.g., redacted text.}

\begin{itemize}
\item No, each sample is an image-text pair.
\end{itemize}

\item \textbf{Are relationships between individual instances made explicit (e.g., users' movie ratings, social network links)?} \textit{If so, please describe how these relationships are made explicit.}

\begin{itemize}
\item No, the dataset is released as it is with no explicit attempt to establish relationships between instances.
\end{itemize}

\item \textbf{Are there recommended data splits (e.g., training, development/validation, testing)?} \textit{If so, please provide a description of these splits, explaining the rationale behind them.}

\begin{itemize}
\item No. The test tasks are existing image classification tasks. We run a deduplication model to try to prevent test set contamination in \pool.
\end{itemize}

\item \textbf{Are there any errors, sources of noise, or redundancies in the dataset?} \textit{If so, please provide a description.}

\begin{itemize}
\item \pool is sourced from Common Crawl, which can be thought of as a snapshot of the internet. Hence, there can be considerable noise (e.g., alt-text being unrelated to its associated image), duplicate data, etc.
\end{itemize}

\item \textbf{Is the dataset self-contained, or does it link to or otherwise rely on external resources (e.g., websites, tweets, other datasets)?} \textit{If it links to or relies on external resources, a) are there guarantees that they will exist, and remain constant, over time; b) are there official archival versions of the complete dataset (i.e., including the external resources as they existed at the time the dataset was created); c) are there any restrictions (e.g., licenses, fees) associated with any of the external resources that might apply to a future user? Please provide descriptions of all external resources and any restrictions associated with them, as well as links or other access points, as appropriate.}

\begin{itemize}
\item The data is not self-contained and rather links other external resources on the internet. Links point to resources distributed across the internet. There is no guarantee that the resources will exist in perpetuity or that that the resources will not change. To mitigate against data poisoning in future \pool downloads, we release SHA256 hashes of images. Due to the size of the dataset, it is not possible to provide it in an archival form.
\end{itemize}

\item \textbf{Does the dataset contain data that might be considered confidential (e.g., data that is protected by legal privilege or by doctor–patient confidentiality, data that includes the content of individuals’ non-public communications)?} \textit{If so, please provide a description.}

\begin{itemize}
\item The dataset is comprised of data that was readily available on the public internet at the time of our download. However, it is possible that the dataset contains confidential information (e.g., private data that is hosted publicly for nefarious reasons or out of ignorance of said data being confidential).
\end{itemize}

\item \textbf{Does the dataset contain data that, if viewed directly, might be offensive, insulting, threatening, or might otherwise cause anxiety?} \textit{If so, please describe why.}

\begin{itemize}
\item Considering the plurality of people and their backgrounds across the world, it is highly likely that there is content in \pool that may upset people. Common Crawl scrapes the internet, which has pornographic, hateful, racist, sexist, and otherwise abhorrent and toxic material. While we attempt to do thorough NSFW filtering, these methods are not 100\% accurate. At the 12.8B scale at which we operate, it is highly likely that there is still toxic content in the dataset. We consider the dataset as a research artifact and hope future work will look critically at \pool in the hopes of developing even better safety filters.
\end{itemize}

\item \textbf{Does the dataset relate to people?} \textit{If not, you may skip the remaining questions in this section.}

\begin{itemize}
\item People may appear in the dataset; however, in an effort to preserve privacy, our downloading tooling automatically blurs all detected faces in \pool images.
\end{itemize}

\item \textbf{Does the dataset identify any subpopulations (e.g., by age, gender)?}

\begin{itemize}
\item While \pool does not explicitly identify subpopulations in its metadata, it is plausible to extract such information for some images using the corresponding textual caption.
\end{itemize}

\item \textbf{Is it possible to identify individuals (i.e., one or more natural persons), either directly or indirectly (i.e., in combination with other data) from the dataset?} \textit{If so, please describe how.}

\begin{itemize}
\item We conjecture that even with our face blurring procedure, it may still be possible to identify individuals. Face blurring relies of a face detection model, which could fail (See Appendix \ref{app:face} for experimental validation of the employed detector). It is also possible to identify certain celebrities or athletes, who may wear distinctive clothing that is associated with them. It is also likely that names are contained in textual captions, though it is not guaranteed that these names correspond to people in images due to the inherent noisiness of internet captions.
\end{itemize}

\item \textbf{Does the dataset contain data that might be considered sensitive in any way (e.g., data that reveals racial or ethnic origins, sexual orientations, religious beliefs, political opinions or union memberships, or locations; financial or health data; biometric or genetic data; forms of government identification, such as social security numbers; criminal history)?} \textit{If so, please provide a description.}

\begin{itemize}
\item Yes. \pool is created using images and corresponding alt-text that are available on the public internet. Given the 12.8B scale of \pool, it is highly likely that there is sensitive data in the dataset. To mitigate against making sensitive content more accessible, we 1) run NSFW image filtering and 2) NSFW text filtering when generating \pool, discarding all samples that are flagged. Additionally we 3) provide automatic face blurring in our \pool download scripts to blur all detected faces.
\end{itemize}

\item \textbf{Any other comments?}

\begin{itemize}
\item \pool is a research artifact, and we hope it will be useful for those studying how to make internet-scale datasets safer.
\end{itemize}

\subsection{Collection Process}
\label{datasheet:collection}

\item \textbf{How was the data associated with each instance acquired?} \textit{Was the data directly observable (e.g., raw text, movie ratings), reported by subjects (e.g., survey responses), or indirectly inferred/derived from other data (e.g., part-of-speech tags, model-based guesses for age or language)? If data was reported by subjects or indirectly inferred/derived from other data, was the data validated/verified? If so, please describe how.}

\begin{itemize}
\item Data is directly downloaded from the public internet.
\end{itemize}

\item \textbf{What mechanisms or procedures were used to collect the data (e.g., hardware apparatus or sensor, manual human curation, software program, software API)?} \textit{How were these mechanisms or procedures validated?}

\begin{itemize}
\item We iterate on the LAION-5B data collection process, making an effort to emphasize safety. We ran python based processing scripts to parse Common Crawl dumps, download images, filter our NSFW content, deduplicate samples against downstream tests sets, blur faces, and compute CLIP features. We ran processes on 100s of AWS CPU nodes for Common Crawl parsing and data download. Other steps were run on one of StabilityAI's GPU cluster. For software links see Q37. For software validation related to NSFW content filtering and face blurring see Appendices \ref{app:nsfw} and \ref{app:face} respectively. In brief, for NSFW image filtering, we validate against commercial APIs and on the NSFW test set introduced in LAION-5B. For face detection (used for face blurring), we evaluate against commercial APIs. We find strong performance for both modules.
\end{itemize}

\item \textbf{If the dataset is a sample from a larger set, what was the sampling strategy (e.g., deterministic, probabilistic with specific sampling probabilities)?}

\begin{itemize}
\item See Q7.
\end{itemize}

\item \textbf{Who was involved in the data collection process (e.g., students, crowdworkers, contractors) and how were they compensated (e.g., how much were crowdworkers paid)?}

\begin{itemize}
\item The researching authors were involved in the data collection as an open source effort. No researchers were compensated specifically for their involvement in this project.
\end{itemize}

\item \textbf{Over what timeframe was the data collected? Does this timeframe match the creation timeframe of the data associated with the instances (e.g., recent crawl of old news articles)?} \textit{If not, please describe the timeframe in which the data associated with the instances was created.}

\begin{itemize}
\item Data was downloaded between December 2022 and March 2023. The urls are collected from Common Crawl dumps between 2014 and 2022. Common Crawl dumps may include urls from the early days of the internet. Hence, the download/collection timeframe does not match the creation timeframe. Additionally, future users of \pool and its subsets will have to download data themselves using our tooling.
\end{itemize}

\item \textbf{Were any ethical review processes conducted (e.g., by an institutional review board)?} \textit{If so, please provide a description of these review processes, including the outcomes, as well as a link or other access point to any supporting documentation.}

\begin{itemize}
\item Our dataset collection process iterates on the LAION-5B process, which found IRB review was not necessary as they ``do not intervene with the people depicted in the data as well as the data being public.''~\cite{laion5b}. Additionally, the NeurIPS ethics review found no serious ethical issues with LAION-5B. We take even more stringent safety measures than the original LAION-5B dataset, in that we filter out data that is flagged as NSFW by our detection pipeline and blur detected faces in \pool, automatically in our released download tooling. All this being said, a formal ethics review has not been conducted to date.
\end{itemize}

\item \textbf{Does the dataset relate to people?} \textit{If not, you may skip the remaining questions in this section.}

\begin{itemize}
\item Yes. People may appear in the dataset. Detected faces are blurred when downloading \pool with our tooling.
\end{itemize}

\item \textbf{Did you collect the data from the individuals in question directly, or obtain it via third parties or other sources (e.g., websites)?}

\begin{itemize}
\item We collect data from websites across the internet.
\end{itemize}

\item \textbf{Were the individuals in question notified about the data collection?} \textit{If so, please describe (or show with screenshots or other information) how notice was provided, and provide a link or other access point to, or otherwise reproduce, the exact language of the notification itself.}

\begin{itemize}
\item Individuals were not notified about the data collection.
\end{itemize}

\item \textbf{Did the individuals in question consent to the collection and use of their data?} \textit{If so, please describe (or show with screenshots or other information) how consent was requested and provided, and provide a link or other access point to, or otherwise reproduce, the exact language to which the individuals consented.}

\begin{itemize}
\item Following our usage of Common Crawl and \url{https://github.com/rom1504/img2dataset} for download images, we respect \texttt{robots.txt} files, which specify parts of websites that a crawler may access. It is, however, possible that images of people, medical images, etc. were uploaded to the internet without a person's consent. To mitigate against such safety concerns we make an effort to do rigorous NSFW filtering and blur all detected faces automatically in our download tooling.
\end{itemize}

\item \textbf{If consent was obtained, were the consenting individuals provided with a mechanism to revoke their consent in the future or for certain uses?} \textit{If so, please provide a description, as well as a link or other access point to the mechanism (if appropriate).}

\begin{itemize}
\item In conjunction with LAION, we use \url{https://laion.ai/dataset-requests/} to monitor user takedown requests. We will also make an effort to provide a user with the url at which their sensitive content is hosted---if they do not have this information already---, so they can take further action as they see fit (e.g., contacting the host to request that the content is taken down from the internet).
\end{itemize}

\item \textbf{Has an analysis of the potential impact of the dataset and its use on data subjects (e.g., a data protection impact analysis) been conducted?} \textit{If so, please provide a description of this analysis, including the outcomes, as well as a link or other access point to any supporting documentation.}

\begin{itemize}
\item We conduct a fairness evaluation on models trained on \pool and its derivative. See Appendix \ref{app:fairness} for details. \citet{Birhane2021MultimodalDM} conduct an extensive study in the context of LAION-400M, which is an image-text dataset also sourced from Common Crawl, finding a plethora of dangerous and unsafe content. Our dataset differs from LAION-400M in that we conduct NSFW preprocessing and face blurring for detected faces. \pool only contains samples that pass our NSFW safety checks and our download tooling automatically blurs detected faces. However, since \pool is created from the internet, it is still likely that it contains some harmful data.
\end{itemize}

\item \textbf{Any other comments?}

\begin{itemize}
\item We hope that future work will use \pool to study how to construct safer, web-scale datasets.
\end{itemize}

\subsection{Preprocessing, Cleaning, and/or Labeling}

\item \textbf{Was any preprocessing/cleaning/labeling of the data done (e.g., discretization or bucketing, tokenization, part-of-speech tagging, SIFT feature extraction, removal of instances, processing of missing values)?} \textit{If so, please provide a description. If not, you may skip the remainder of the questions in this section.}

\begin{itemize}
\item Yes. See Q7.
For more details see Appendix \ref{app:metadata}.
\end{itemize}

\item \textbf{Was the “raw” data saved in addition to the preprocessed/cleaned/labeled data (e.g., to support unanticipated future uses)?} \textit{If so, please provide a link or other access point to the “raw” data.}

\begin{itemize}
\item Raw data is not available or distributed due to safety considerations. We distribute only urls that are in the dataset on HuggingFace---and not urls of images our preprocessing flagged as NSFW.
\end{itemize}

\item \textbf{Is the software used to preprocess/clean/label the instances available?} \textit{If so, please provide a link or other access point.}

\begin{itemize}
\item We use the following, open-source software to aid in data processing:
\begin{itemize}
\item Apache Spark: \url{https://spark.apache.org}
\item Ray: \url{https://www.ray.io}
\item img2dataset: \url{https://github.com/rom1504/img2dataset}
\item OpenAI CLIP: \url{https://github.com/openai/CLIP}
\item Near dedulicate detector: \url{https://github.com/lyakaap/ISC21-Descriptor-Track-1st}
\item Face detector: \url{https://github.com/deepinsight/insightface}
\item Detoxify, for detecting toxic language: \url{https://github.com/unitaryai/detoxify}
\item A modified version of the following NSFW image detector: 
\url{https://github.com/LAION-AI/CLIP-based-NSFW-Detector}. Specifically, we use the dataset used to train this model to train our own 4-layer MLP classifier.
\end{itemize}
\end{itemize}

\item \textbf{Any other comments?}

\begin{itemize}
\item \pool and \datanet would not be possible without tools developed by the open-source community.
\end{itemize}

\subsection{Uses}

\item \textbf{Has the dataset been used for any tasks already?} \textit{If so, please provide a description.}

\begin{itemize}
\item The full dataset (and subsets) have been used to train several CLIP models at various scales and compute budgets as presented in our main paper. We evaluate these models zero-shot on 38 downstream image classification and retrieval tasks. See Section \ref{sec:evaluation} and Appendix \ref{sec:app-eval} for more details.
\end{itemize}

\item \textbf{Is there a repository that links to any or all papers or systems that use the dataset?} \textit{If so, please provide a link or other access point.}

\begin{itemize}
\item No. However, there is a leaderboard associated with \datanet. Interested parties can investigate the submissions and further study publications that make use of our data. See: \url{https://www.datacomp.ai/leaderboard.html}.
\end{itemize}

\item \textbf{What (other) tasks could the dataset be used for?}

\begin{itemize}
\item The dataset could also be used for training image captioning models and language-conditional image generation models. Note: generative image models trained on \pool are not expected to generate recognizable human faces as our download tooling automatically blurs detected faces. \pool could be used for sociological studies, for example, examining societal biases or to better understand what is on the public internet.
\end{itemize}

\item \textbf{Is there anything about the composition of the dataset or the way it was collected and preprocessed/cleaned/labeled that might impact future uses?} \textit{For example, is there anything that a future user might need to know to avoid uses that could result in unfair treatment of individuals or groups (e.g., stereotyping, quality of service issues) or other undesirable harms (e.g., financial harms, legal risks) If so, please provide a description. Is there anything a future user could do to mitigate these undesirable harms?}

\begin{itemize}
\item \pool and its derivatives are not intended for production ready products, including but not limited to those related to race, gender identity or expression, ethnicity, sexual orientation, age, socioeconomic status, disability, religion, national origin or creed. \pool is not suitable for any software that makes decisions involving people. \pool is collected from the internet and hence reflects many of the biases, unfairness, and stereotypes currently existing in our societies. \pool is intended as a research artifact to study multimodal dataset curation and the effect of data curation strategies on downstream models.
\end{itemize}

\item \textbf{Are there tasks for which the dataset should not be used?} \textit{If so, please provide a description.}
\begin{itemize}
\item \pool in its current form or the subsets presented in this paper should not be used in software that makes decisions related to people. The known biases (Appendix \ref{app:fairness}) make deploying software, especially widely decimated production-level products, built on \pool incredibly irresponsible. \pool is designed as a research artifact for academic exploration. We also do not condone the use of \pool in surveillance or military applications.
\end{itemize}

\item \textbf{Any other comments?}

\begin{itemize}
\item Our goal with \pool and \datanet was to put a benchmark in place so the community can start measuring dataset progress along many different axes (e.g., model performance on diverse tasks). We believe this is crucial to develop more performant and safer datasets.
\end{itemize}

\subsection{Distribution}

\item \textbf{Will the dataset be distributed to third parties outside of the entity (e.g., company, institution, organization) on behalf of which the dataset was created?} \textit{If so, please provide a description.}

\begin{itemize}
\item Yes. We use HuggingFace datasets for public release.
\end{itemize}

\item \textbf{How will the dataset be distributed (e.g., tarball on website, API, GitHub)?} \textit{Does the dataset have a digital object identifier (DOI)?}

\begin{itemize}
\item The dataset will be distributed via HuggingFace datasets at \href{https://huggingface.co/datasets/mlfoundations/datacomp_pools/tree/main}{\url{https://huggingface.co/datasets/mlfoundations/datacomp_pools/tree/main}}
\end{itemize}

\item \textbf{When will the dataset be distributed?}

\begin{itemize}
\item \datanet will be available starting May 2023.
\end{itemize}

\item \textbf{Will the dataset be distributed under a copyright or other intellectual property (IP) license, and/or under applicable terms of use (ToU)?} \textit{If so, please describe this license and/or ToU, and provide a link or other access point to, or otherwise reproduce, any relevant licensing terms or ToU, as well as any fees associated with these restrictions.}

\begin{itemize}
\item We distribute the url-text sample and metadata under a standard CC-BY-4.0 licence.
\end{itemize}

\item \textbf{Have any third parties imposed IP-based or other restrictions on the data associated with the instances?} \textit{If so, please describe these restrictions, and provide a link or other access point to, or otherwise reproduce, any relevant licensing terms, as well as any fees associated with these restrictions.}

\begin{itemize}
\item We do not copyright samples in the dataset.
\end{itemize}

\item \textbf{Do any export controls or other regulatory restrictions apply to the dataset or to individual instances?} \textit{If so, please describe these restrictions, and provide a link or other access point to, or otherwise reproduce, any supporting documentation.}

\begin{itemize}
\item The dataset is provided as an index of url-text pairs.
\end{itemize}

\item \textbf{Any other comments?}

\begin{itemize}
\item We provide several subsets of \pool (between 12.8M samples and the full dataset of 12.8B samples). Hence, it is possible to download and experiment with subset of the data.
\end{itemize}

\subsection{Maintenance}

\item \textbf{Who will be supporting/hosting/maintaining the dataset?}

\begin{itemize}
\item HuggingFace currently hosts the url-text pairs and metadata. The \datanet team will be responsible for maintaining the dataset.
\end{itemize}

\item \textbf{How can the owner/curator/manager of the dataset be contacted (e.g., email address)?}

\begin{itemize}
\item We can be contacted at \url{contact@datacomp.ai}.
\end{itemize}

\item \textbf{Is there an erratum?} \textit{If so, please provide a link or other access point.}

\begin{itemize}
\item Currently there are no errata. If issues are discovered, we will communicate with the public via our website \url{https://datacomp.ai}.
\end{itemize}

\item \textbf{Will the dataset be updated (e.g., to correct labeling errors, add new instances, delete instances)?} \textit{If so, please describe how often, by whom, and how updates will be communicated to users (e.g., mailing list, GitHub)?}

\begin{itemize}
\item At the present time there is no intention to update \pool for scientific reasons. However, we will respond to user takedown requests (see Q56). \pool is inherently noisy and the purpose of releasing it is to encourage researchers in the community to study dataset cleaning in the context of image-text samples.
\end{itemize}

\item \textbf{If the dataset relates to people, are there applicable limits on the retention of the data associated with the instances (e.g., were individuals in question told that their data would be retained for a fixed period of time and then deleted)?} \textit{If so, please describe these limits and explain how they will be enforced.}

\begin{itemize}
\item We will use the following website, \url{ https://laion.ai/dataset-requests}, for user takedown requests, where ``Sample ID'' is the sample uid.
\end{itemize}

\item \textbf{Will older versions of the dataset continue to be supported/hosted/maintained?} \textit{If so, please describe how. If not, please describe how its obsolescence will be communicated to users.}

\begin{itemize}
\item This is the first version of \datanet and the associated \pool dataset. We do not intend to maintain deprecated version of \pool. We will communicate deprication notices through our website: \url{https://datacomp.ai}.
\end{itemize}

\item \textbf{If others want to extend/augment/build on/contribute to the dataset, is there a mechanism for them to do so?} \textit{If so, please provide a description. Will these contributions be validated/verified? If so, please describe how. If not, why not? Is there a process for communicating/distributing these contributions to other users? If so, please provide a description.}

\begin{itemize}
\item All alterations to the dataset will be handled on a case-by-case basis.
\end{itemize}

\item \textbf{Any other comments?}

\begin{itemize}
\item We encourage community members to contact us at \url{contact@datacomp.ai} with inquiries related to dataset maintainence.
\end{itemize}

\end{enumerate}